%% file: neurips_2026.tex
\documentclass{article}

\usepackage[preprint]{neurips_2026}

\usepackage[utf8]{inputenc} 
\usepackage[T1]{fontenc}    
\usepackage{hyperref}       
\usepackage{url}            
\usepackage{booktabs}       
\usepackage{amsfonts}       
\usepackage{nicefrac}       
\usepackage{microtype}      
\usepackage{xcolor}         
\usepackage{graphicx}       
\usepackage{colortbl}       
\usepackage{array}          
\usepackage{amsmath}
\usepackage{enumitem}       

\newcommand{\frameworkName}{\textit{CLR-voyance}}
\newcommand{\datasetName}{\textit{CLR-POMDP}}

\title{\frameworkName: Reinforcing Open-Ended Reasoning for Inpatient Clinical Decision Support with Outcome-Aware Rubrics}

\author{%
  Aishik Nagar$^{1}$ \quad Arun-Kumar Kaliya-Perumal$^{2}$ \quad
  Yu-Hsuan Han$^{3}$ \quad Andrew Sheng-Han Huang$^{4}$ \\
  {\bfseries Kristen Kee$^{5}$ \quad Yushi Cao$^{1}$ \quad
  Yiming Chen$^{1}$ \quad Hongchao Jiang$^{1}$} \\
  \vspace{2pt}\\
  $^{1}$ASUS Intelligent Cloud Services (AICS) 
  $^{2}$Rehabilitation Research Institute of Singapore, \\
        Nanyang Technological University
  $^{3}$Department of Family Medicine, \\
        Taipei Veterans General Hospital
  $^{4}$School of Medicine, \\ National Yang Ming Chiao Tung University 
  $^{5}$Yong Loo Lin School of Medicine, \\
        National University of Singapore
}

\graphicspath{{results/figures/}{results/}}

\begin{document}

\maketitle

\begin{abstract}
Inpatient clinical reasoning is a sequential decision under partial observability: the clinician sees the admission so far and must choose the next action whose downstream consequences are not yet visible. 
Existing clinical-LLM evaluations and reinforcement-learning (RL) rewards collapse this structure into closed-form retrieval, full clinical journey leakage, or unanchored LLM-as-judge scoring.
Here, we introduce \textit{\frameworkName{}}, an end-to-end framework that reformulates inpatient reasoning as a Partially Observable Markov Decision Process (POMDP) and supervises it with rewards that are simultaneously \emph{outcome-grounded} and \emph{clinician-validated}. 
We instantiate the formulation as \textit{\datasetName{}}, which partitions successful patient journeys into a policy-visible \emph{past} and an oracle-only \emph{future}.
Using the past information, an oracle LLM generates a case-specific query-answer pair, and the \textbf{first adaptive rubric for clinical reasoning} which is verifiable in the future of the patient journey.
These rubrics are used for both RL post-training and evaluation of models for inpatient clinical reasoning. 
\frameworkName{} post-trains Qwen3-8B and MedGemma-4B with GRPO followed by model merging, yielding state-of-the-art inpatient clinical reasoning while retaining generalist capabilities.
\frameworkName{}-8B achieves \textbf{84.91\% on \datasetName{}}, ahead of frontier medical reasoning models like GPT-5  (77.83\%) and MedGemma-27B (66.66\%) and has comparable or better performance on existing medical benchmarks.
To ensure that our setting is clinically meaningful, we conduct a large-scale clinician alignment study, where physicians curate per-case rubrics, grade candidate responses against them, and provide blinded pairwise preferences of model reasoning. 
This study provides insights on clinical LLM-as-a-judge and clinical preference-model selection, which can inform the community at large.
\frameworkName{} has been deployed for \textbf{6+ months at a partner public hospital}, drafting thousands of reasoning-heavy inpatient notes and 1.03T+ tokens processed to date.
\end{abstract}


\section{Introduction}

At the heart of inpatient medicine is a sequential decision under partial observability \citep{mehandru2025er}. Standing at the bedside, a clinician sees the admission so far (triage, vitals, prior labs, imaging, the medications already given, the consults already placed) and must choose the next action whose downstream consequences they have not yet observed: which drug to start or hold, when to transfer. The information that would resolve the question is, by construction, in the future. The clinician acts under uncertainty, and the patient's trajectory afterwards is what tells us whether they acted well. The closer a large language model (LLM) gets to that bedside, the more its evaluations and reward signals must align with this process \citep{bedi2025medhelm}.

\begin{figure}[t]
  \centering
  \includegraphics[width=\linewidth]{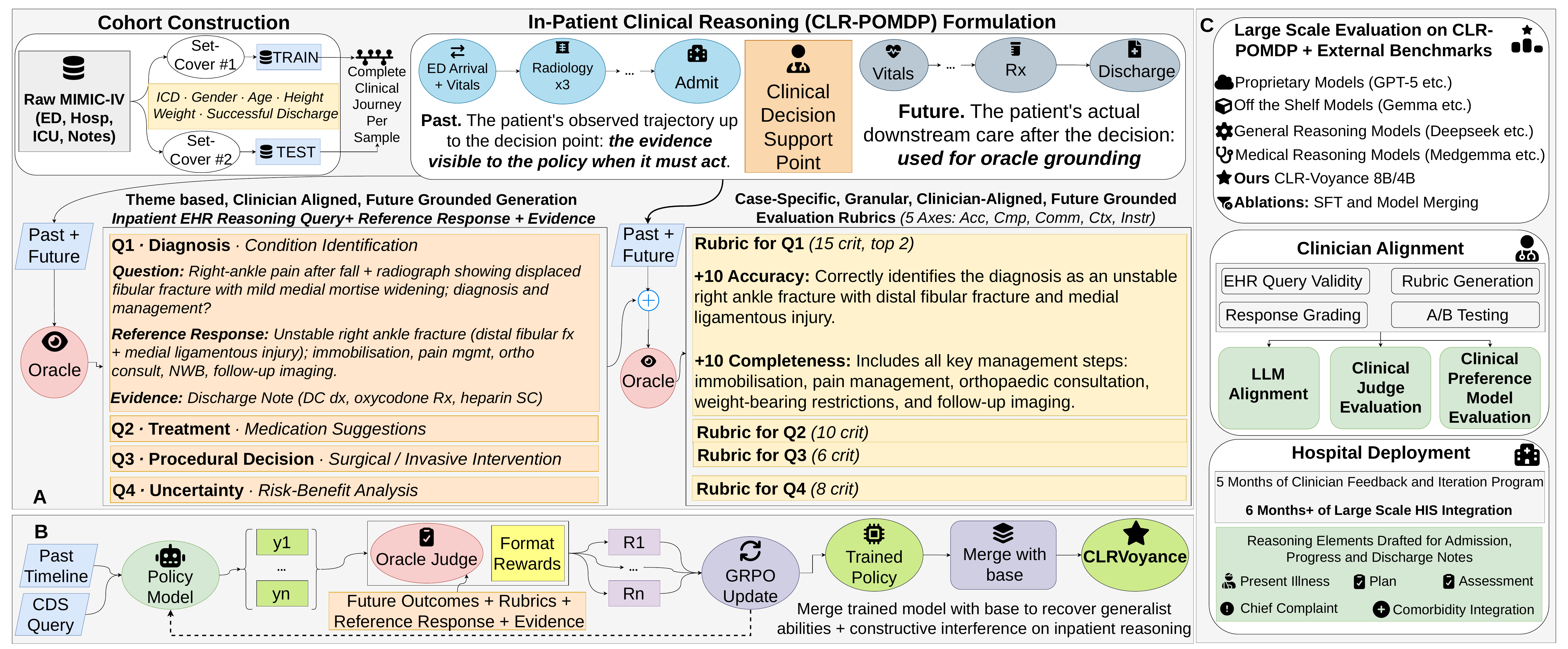}
  \caption{Overview of the \frameworkName{} framework: \textbf{(A)} \datasetName{} construction, \textbf{(B)} GRPO post-training and weight-space merging on the rubric reward, and \textbf{(C)} evaluation, two-cohort clinician validation, and hospital deployment.}
  \label{fig:overview}
\end{figure}

The decisions inpatient clinicians take every day are, in fact, structurally inexpressible without a past/future split. For example, actions like \emph{Dosage adjustments} (``does this patient's vancomycin need a renal-function-based dose change?'') and \emph{transfer-readiness} decisions share a common structure: the past makes the question well-posed (the medication is already on board, the patient is already in the ICU), and the future supplies the verifiable answer (the actual dose change, the actual transfer event). Benchmarks that take a single time-collapsed document (a discharge summary, a code list) as input cannot, by construction, supervise these query types without leaking the answer into the question.

Yet, existing clinical-LLM works usually do not align with this process. Exam-style benchmarks evaluate reasoning in closed-form formats that frontier models have largely saturated, and their scores remain only loosely predictive of real-world clinical performance \citep{zuo2025medxpertqa,kim2025limitations,bedi2025medhelm,wang2025accuracy}. Longitudinal-EHR benchmarks have begun to push past the exam format, but they typically expose the entire admission and recover ground truth from the document that already records the answer \citep{kweon2024ehrnoteqa,gao2023dr,yang2026ehrstruct,fleming2024medalign,liao2025ehr}. 
Clinical reward signals present similar problems. Binary verifiers inspired by math and code work on MCQ style ``verifiable medical problems'' \citep{chen2024huatuogpt} but cannot grade open-ended management, while multi-objective rule-based clinical-RL rewards \citep{gu2025clinical,tziakouri2025reinforcement,zhang2025editgrpo,xu2025medground,liao2025ehr} ground in static references rather than in what a real care team did. General-domain rubric-as-reward work \citep{arora2025healthbench,gunjal2025rubrics,chen2025rm,xu2026alternating,mu2024rule} replaces those rules with rubrics that are static, conversation-specific, or per-instance for general text, but never conditioned on a patient's downstream trajectory. Finally, most clinical-LLM studies stop at benchmark numbers \citep{chen2024huatuogpt,gu2025clinical,zhang2025editgrpo,liao2025ehr,yang2026ehrstruct,kweon2024ehrnoteqa}. While some studies add structured clinician evaluation \citep{tu2024towards,vedadi2025towards,arora2025healthbench,bedi2025medhelm,wang2025accuracy}, the annotations are typically reported once as a grader-vs-clinician correlation and then set aside. Only a small handful reach real production hospital deployment, the closest analogues being a research pilot on eight patients \citep{masayoshi2025ehr} and a hospital-wide RAG retrieval / summarisation assistant \citep{griot2025implementation}, neither of which trains on inpatient clinical reasoning. 
\textbf{\textit{Thus, no existing study jointly supports outcome-grounded inpatient reasoning at scale, with clinician-alignment at every step of the process, and real-world large scale evidence verified outside its training distribution}}. 

We introduce \textbf{\frameworkName{}}\footnote{Appendix~\ref{app:contrast-matrix} contrasts \frameworkName{} against 60+ prior works across eleven factors as a complete reference.}, an end-to-end framework that closes these gaps as outlined below:

\begin{itemize}
    \item \textbf{A real-EHR formulation for partially observable inpatient reasoning.}
    We introduce \datasetName{}, a POMDP-style formulation built from MIMIC-IV admissions in which models observe only the patient event history before a sampled decision point, while evaluation is grounded in the successful future care actions taken for the patient. The construction preserves raw longitudinal EHR event streams at production-realistic context lengths ($\geq$60K tokens).

    \item \textbf{Outcome-grounded judging with per-case adaptive rubrics.}
    We propose the \textit{Grounded Judge}: an oracle LLM with access to the full admission trajectory which generates a clinically meaningful query, an event-cited reference answer, and a per-case multi-axis rubric conditioned on that patient's actual future. A separate grading call converts this rubric into dense rewards, enabling open-ended clinical reasoning evaluation beyond static or outcome-agnostic rubrics.
    This is, to our knowledge, the first per-case adaptive rubric for inpatient clinical reasoning, and more broadly the first rubric-as-reward signal for open-ended decision tasks anchored in true causal outcomes.

    \item \textbf{Post-training small models for frontier-level inpatient reasoning.}
    We train Qwen3-8B and MedGemma-4B with GRPO on rubric-derived rewards, with weight-space merging to recover generalist capabilities. The resulting \frameworkName{}-8B reaches 84.91\% on \datasetName{}, outperforming GPT-5, MedGemma-27B, and open clinical-reasoning baselines.

    \item \textbf{Clinician validation and large scale hospital integration.}
    Four board-certified physicians validate the pipeline across spine/orthopaedic and general-medicine/obesity cohorts. Their annotations support rubric and judge alignment, a clinical LLM-as-a-judge evaluation, and a clinical preference-model evaluation. We perform large scale hospital integration on a physician defined reasoning scope. \frameworkName{} has been \textbf{deployed for over six months at a partner public hospital}, drafting the integrative reasoning elements alongside summarisation primitives for the descriptive ones, with over 1.03 trillion tokens processed to date and \textbf{thousands of patient notes generated daily}. 
\end{itemize}


\input{related_work}

\input{methodology}

\section{Results}
\label{sec:results}

\begin{figure}[t]
  \centering
  \includegraphics[width=\linewidth]{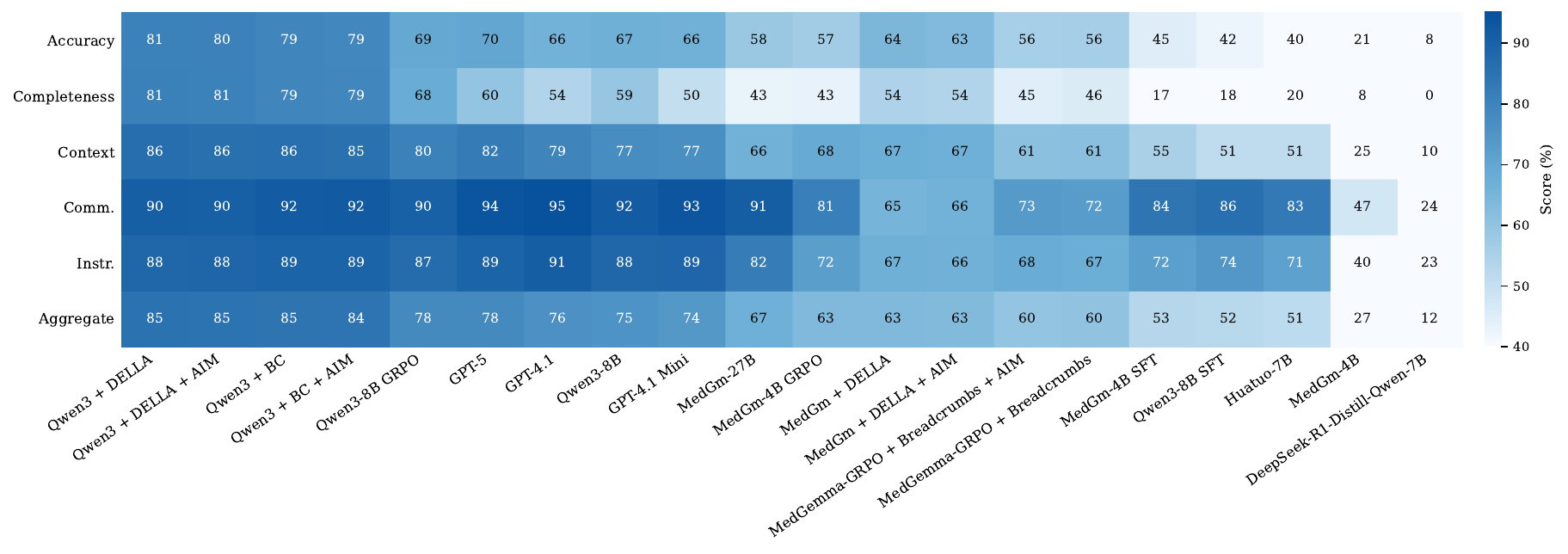}
  \caption{Per-axis \datasetName{} scores (\%). Models ordered by descending Aggregate.}
  \label{fig:per-axis-heatmap}
\end{figure}

We evaluate \frameworkName{}-8B on the held-out \datasetName{} test split, comprising 7{,}927 admissions with four action-space queries each. Aggregate rubric scores for the full panel are summarised in Figure~\ref{fig:per-axis-heatmap}. \frameworkName{}-8B (Qwen3-GRPO with DELLA-Linear merging) attains an aggregate score of \textbf{84.91\%}. 
For reference, GPT-5 \citep{singh2025openai} reaches 77.83\%, Qwen3-8B GRPO~ 77.95\%, GPT-4.1 75.74\% \citep{openai2025gpt41}, the Qwen3-8B base 75.28\% \citep{yang2025qwen3}, GPT-4.1-mini 73.76\%, MedGemma-27B 66.66\% \citep{sellergren2025medgemma}, and HuatuoGPT-o1-7B~\citep{chen2024huatuogpt} 51.24\%. The four \frameworkName{} merge variants (DELLA-Linear, DELLA-Linear+AIM, Breadcrumbs, Breadcrumbs+AIM) score within 0.7 percentage points (pp) of one another (84.25--84.91\%), and each is at least 6.4~pp above the strongest non-merge baseline. A paired Wilcoxon signed-rank test\footnote{The paired Wilcoxon signed-rank is a non-parametric paired-samples test whose null hypothesis is that the median of per-admission score differences is zero; the reported $p$ is the probability of seeing a difference distribution at least this extreme under that null, so $p<10^{-300}$ is an overwhelming rejection of the null in favour of \frameworkName{}-8B exceeding GPT-5.} on matched per-admission scores yields $p<10^{-300}$ for \frameworkName{}-8B exceeding GPT-5.

\paragraph{Per-axis decomposition.}
We next decompose the aggregate gap by rubric axis. The improvement over GPT-5 is concentrated on the three clinical-content axes: \frameworkName{}-8B exceeds GPT-5 by 21.36~pp on Completeness (80.93 vs.~59.57), 11.08~pp on Accuracy (80.73 vs.~69.65), and 4.05~pp on ContextAwareness (86.16 vs.~82.11). On the two conversational axes, GPT-5 retains a small advantage of 3.38~pp on CommunicationQuality (93.83 vs.~90.45) and 0.98~pp on InstructionFollowing (89.10 vs.~88.12). We interpret this asymmetry as a direct consequence of the per-case rubric design (\S\ref{sec:method-judge}). \emph{Accuracy}, \emph{Completeness}, and \emph{ContextAwareness} are graded against the patient's downstream events, where outcome-grounded supervision is most informative. \emph{CommunicationQuality} and \emph{InstructionFollowing} are graded on stylistic criteria on which modern frontier models are already strong, a saturation pattern that recent clinical-LLM meta-analyses also report~\citep{wang2025accuracy,kim2025limitations}. \citet{arora2025healthbench} report the same axis-level pattern at the cross-model level for HealthBench. Thus, in the inpatient clinical reasoning domain, \frameworkName{} closes most of the completeness gap and a substantial fraction of the accuracy gap that HealthBench identifies.

\paragraph{Action-space decomposition.}
The same pattern holds across the four reasoning action-space buckets. Compared to GPT-5, \frameworkName{}-8B is higher by 8.0~pp on Procedural (75.0 vs.~67.0), 7.6~pp on Uncertainty (78.2 vs.~70.5), 6.5~pp on Treatment (72.4 vs.~65.9), and 2.7~pp on Diagnosis (66.9 vs.~64.2). The smallest gain on Diagnosis is consistent with frontier models having broad exposure to labelled diagnostic tasks during pre-training~\citep{wang2025accuracy}. The largest gains fall on Procedural and Uncertainty, the two buckets where the correct answer is most directly conditioned on the patient's actual subsequent trajectory~\citep{mehandru2025er}: the next procedure depends on what the clinical team actually did next, and the risk-benefit call depends on the actual subsequent escalation or de-escalation. The same axis-level pattern repeats against the medical-LLM baselines with larger absolute gaps, consistent with their training corpora optimising for general medical-text fluency rather than for trajectory-grounded reasoning over a partially observed admission.


\paragraph{Per-instance analysis.}
Across the 7{,}927 paired (admission $\times$ query) instances against GPT-5, the two models tie on $47$--$57\%$ of cases per action-space bucket. On the diverging cases \frameworkName{} is preferred 3 to 5 times more often than GPT-5, with the ratio reaching $4.0\times$ on Procedural Decision Making (726 wins to 181 losses) and $4.5\times$ on Responding under Uncertainty (668 to 148). \frameworkName{} satisfies $73.6\%$ of positive rubric criteria against $68.2\%$ for GPT-5 and triggers the rubric's negative-point safety flags at $1.30\%$ vs.~$1.79\%$, a 27\% relative reduction matching the role of rule-based safety rewards in RLHF~\citep{mu2024rule}. The Completeness lift coincides with longer outputs ($\sim$8.3K vs.~$\sim$3.0K characters), which we interpret as the rubric reward pulling the policy toward exhaustive enumeration of clinical steps rather than incidental verbosity~\citep{gunjal2025rubrics}. Four representative cases with full rubrics and per-criterion breakdowns are in Appendix~\ref{app:cases}.

\paragraph{Effect of weight-space merging.}
On Qwen3-8B, GRPO alone reaches $77.95\%$ aggregate (marginally above GPT-5), and DELLA-Linear merging~\citep{deep2024dellamerging} with the base then adds a further $6.96$~pp to take \frameworkName{}-8B to $84.91\%$, broadly distributed across the three clinical-content axes. Per-axis numbers, the $\pm 0.7$~pp band across the four merge variants (DELLA-Linear, DELLA-Linear+AIM~\citep{nobari2025activationinformedmerginglargelanguage}, Breadcrumbs~\citep{davari2024breadcrumbs}, Breadcrumbs+AIM),. The analysis on the 4B family are in Appendix~\ref{app:training-ablation} due to space constraints.

\paragraph{GRPO versus SFT on the same data.}
SFT and GRPO are run as independent ablations from the same base. On Qwen3-8B, which ships with a reasoning prior, SFT on the oracle reference $(\rho^{\star}, y^{\star})$ regresses the aggregate from $75.32\%$ (base) to $52.36\%$ by contracting the policy onto a single short oracle answer, while GRPO reaches $77.99\%$ by rewarding rubric-aligned enumeration. On MedGemma-4B, which lacks the $\mathtt{<think>}$/answer scaffolding, SFT instead lifts the very low base from $27.01\%$ to $52.89\%$ on the conversational axes, but GRPO from the same base reaches $63.40\%$ by adding the content-axis lift the static SFT target cannot provide. Per-configuration failure-mode analysis and the full per-axis breakdown are in Appendix~\ref{app:training-ablation}.

\paragraph{The 4B family: \frameworkName{}-4B impact on MedGemma-4B.}
The merge dynamics differ on the 4B family. The MedGemma-4B base reaches $27.01\%$ aggregate and GRPO from the same base reaches $63.40\%$. The DELLA-Linear merge then leaves the aggregate near-flat at $63.35\%$ but \emph{re-balances} the axes. Accuracy and Completeness gain $+7.48$ and $+11.32$~pp, while CommunicationQuality and InstructionFollowing drop by $16.22$ and $5.60$~pp. We attribute this to the smaller 4B base having less generalist conversational headroom to recover via weight-space interpolation, so the merge dilutes rather than augments the GRPO-installed conversational behaviour~\citep{yuan2026behavior}. Qualitative observation in hospital production nonetheless shows improved instruction-following and structured-generation behaviour for the merged 4B. Relative to the MedGemma-4B base, our final 4B model is $+36.34$~pp on aggregate, with per-axis lifts of $+43.54$, $+46.38$, and $+42.50$ on Accuracy, Completeness, and ContextAwareness.

\paragraph{Reasoning-distilled baselines.}
For reference, we additionally evaluate two open-source post-trained checkpoints derived from alternative GRPO based reasoning recipes and the same base-model size and family as \frameworkName{}. 
DeepSeek-R1-Distill-Qwen-7B reaches an aggregate of $11.99\%$, an order of magnitude lower than every other model in our evaluation. 
On inspection, we find that the model emits long reasoning traces that fail every positive Accuracy and Completeness criterion the rubric scores against. 
DeepSeek-R1-0528-Qwen3-8B produces syntactically corrupted outputs (truncated text, repeated tokens, or empty responses) on the majority of \datasetName{} cases. 
Taken together, this suggests that general-purpose reasoning post-training, which is often overindexed on verifiable tasks such as mathematics and code~\citep{shao2024deepseekmath,gunjal2025rubrics}, does not transfer directly to inpatient EHR reasoning for clinical decision support.

\input{external_evals/external_evals}

\input{clinician_validation}

\input{hospital_deployment}

\input{discussion_qualitative}

\medskip

\bibliographystyle{plainnat}
\bibliography{custom}


\appendix



\input{appendix_contrast}

\input{appendix_repro}

\input{appendix_clinician}

\input{appendix_qualitative}

\input{appendix_deploy}

\input{appendix_cases}


\newpage
\input{checklist.tex}

\end{document}

%% file: related_work.tex
\section{Related Work}
\paragraph{Clinical benchmark}
The longitudinal-EHR benchmark line spans discharge-summary QA, retrospective progress-note tasks, structured snapshots, instruction-following over the full record, event-stream code prediction, and supervised reasoning over per-task slices \citep{kweon2024ehrnoteqa,gao2023dr,yang2026ehrstruct,fleming2024medalign,ben2024cpllm,wornow2023ehrshot,liao2025ehr}. The closest formulation to our setting, ER-Reason \citep{mehandru2025er}, splits ER cases past/future across five decision points but exposes the full stay on two, and reports zero-shot frontier-model scores rather than training a model on its own task setting. Dialogue and clinical-pathway benchmarks \citep{nori2025sequential,tu2024towards,vedadi2025towards,li2025aligning} take partial observability seriously but operate on patient-actor or online-dialogue data, not raw inpatient EHR. Across the longitudinal-EHR line, prior work has been overwhelmingly evaluation-only: a partially-observable inpatient-reasoning regime that doubles as a training distribution, and on which models can actually be post-trained at scale, has not been built.

\paragraph{Clinical reward-shaping}
HealthBench uses adaptive rubrics rubric based  evaluations but targets free-form health \emph{conversations} with little inpatient-EHR coverage. The ``verifiability bottleneck'' for clinical RL \citep{chen2024huatuogpt,liu2025beyond,huang2025reinforcement,xu2026alternating,wang2025text2grad} therefore remains: in clinical care almost every plausible action looks reasonable, and the one ground truth available for a single patient (the future care actions taken on that patient) has not, until now, been used as the reward signal.

\paragraph{Practical evidence}
Among the clinician-evaluation subset, none targets inpatient-EHR-based reasoning, and most studies \citep{zhou2025automating,wu2025first,shan2025comparing,zoller2024human} are reported once as a grader-vs-clinician correlation and then set aside rather than re-used as a training resource. The current regime of clinical-LLM deployment, which has been integrated in hospitals at all, has been retrieval and summarisation \citep{griot2025implementation,fleming2024medalign,nagar2025umedsum}: faithful condensation of the patient timeline into the \emph{descriptive} elements of an inpatient note (Description, Subjective, Objective). The largest reported deployment to date is an open-ended chatbot \citep{griot2025implementation} in which clinical-decision-support is the smallest of four use-case buckets at $11\%$. No clinical-LLM system has yet been directly integrated into the day-to-day clinical activities that demand decision support under partial observability.

%% file: methodology.tex

\section{Methodology}
\label{sec:method}

We organise the methodology around three contributions: (i) the \datasetName{} formulation (\S\ref{sec:method-data}), (ii) the Grounded Judge that produces dense outcome-aware reward signals (\S\ref{sec:method-judge}), and (iii) the post-training and evaluation stack (\S\ref{sec:method-train}). Figure~\ref{fig:overview} summarises the entire pipeline.

\subsection{From EHRs to a Clinical POMDP: the \datasetName{} formulation}
\label{sec:method-data}

\paragraph{Source and POMDP framing.}
We treat each admission as a POMDP $\mathcal{M}=(\mathcal{S},\mathcal{A},\mathcal{O},\mathcal{T},\mathcal{Z},r)$ over latent patient states $\mathcal{S}$, clinical actions $\mathcal{A}$, EHR observations $\mathcal{O}$, transition kernel $\mathcal{T}$, observation-emission kernel $\mathcal{Z}$, and reward $r$ ($r_{\text{rub}}$ of \S\ref{sec:method-judge}). The clinician observes an event stream $o_{1:t}\in\mathcal{O}^t$ from a latent patient state $s_t\in\mathcal{S}$ and chooses the next action $a_t\in\mathcal{A}$ whose downstream consequences are not yet visible. \datasetName{} is derived from MIMIC-IV \citep{johnson2023mimic,johnson2024mimicivphysionet,johnson2023mimicivedphysionet,johnson2023mimicivnotephysionet,goldberger2000physiobank}; we extract a unified, chronologically ordered timeline spanning the four MIMIC-IV modules (ED, hosp, ICU, notes) and present it to the policy verbatim in JSON-shaped event-level form (no pre-summarisation, no template), at a 65{,}536-token context window. 

\paragraph{Cohort and temporal split.}
A greedy set-cover cohort over (ICD code, sex, binned age, height, weight) selects training admissions, with a disjoint $20\%$ random sample of the remainder forming the test split. Each admission's chronologically sorted event sequence $e_{1:n}$ is split at $k \sim \mathcal{N}(n/2,\, n/6)$ clamped to $[1, n{-}1]$, drawing many splits per admission so the dataset stratifies across the stay (triage through discharge) without a hand-coded stage taxonomy. Two constraints support outcome-grounded supervision: (a) ICD-source events are removed from the past (the diagnostic label is what a reasoner should \emph{recover}, not read off a row), and (b) admissions with re-admission or death within two years are excluded, so the future is a high-fidelity care trajectory rather than a record of downstream harm.

\paragraph{Action spaces and training datum.}
Each split instantiates four reasoning action spaces that span the inpatient decision support under partial observability (Diagnosis, Treatment, Procedural, Uncertainty), one query per space. The resulting datum is $d = (\mathrm{past},\, \mathrm{meta},\, c,\, q,\, y^{\star},\, \rho^{\star},\, \mathrm{src},\, R)$, where $c$ is the action-space category, $q$ the query, $y^{\star}$ and $\rho^{\star}$ the reference answer and reasoning, $\mathrm{src}$ the future events evidencing $y^{\star}$, and $R$ the rubric (\S\ref{sec:method-judge}). Train and test splits are built admission-wise.

\subsection{The Grounded Judge: outcome-aware queries, references, and per-case rubrics}
\label{sec:method-judge}

The Grounded Judge is an oracle role played by an LLM that has full access to the withheld future and uses it across two stages to author a query, a reference, and a per-case rubric. \emph{Stage 1} (prompted with $(\mathrm{past}, \mathrm{future}, c)$) emits $(q, y^{\star}, \rho^{\star}, \mathrm{src})$ under two hard constraints: $q$ must be answerable from the past alone and $y^{\star}$ must be evidenced by at least one future event in $\mathrm{src}$ (source). \emph{Stage 2} synthesises the rubric $R$ conditioned on past, future, query, and reference. Each criterion is a self-contained pass/fail statement tagged with one of five clinician-centred axes (Accuracy, Completeness, ContextAwareness, CommunicationQuality, InstructionFollowing) and weighted in $[-10,+10]$, with positive points rewarding helpful behaviour and negative points penalising unsafe behaviour. These axes are designed to capture two complementary aspects of what matters to clinicians: \emph{Accuracy}, \emph{Completeness}, and \emph{ContextAwareness} grade \emph{what} the candidate proposes, anchored to the patient's downstream events and to the actions a real care team actually took next; \emph{CommunicationQuality} and \emph{InstructionFollowing} grade \emph{how} that information is presented, reflecting both how clinicians consume model output and how they prefer the model's decision support to be integrated into hospital workflows. 
Verbatim oracle prompts and the JSON schema are in Appendix~\ref{app:prompts}.

\paragraph{Rubric-based grading.}
A separate Grounded Judge call grades the policy's candidate $\hat y$ against $R$, seeing past, reference, source list, query, candidate, and the rubric with compact criterion identifiers, and returning per-criterion booleans. The scalar reward is
\[
  r_{\text{rub}}(\hat y \mid d) \;=\; \operatorname{clip}_{[0,1]}\!\Bigl(\tfrac{\sum_{i:\,j_i=\text{true}} p_i}{\sum_{i:\, p_i > 0} p_i}\Bigr),
\]
where $p_i$ is the signed weight of criterion $i$ and $j_i$ the grader's verdict. Normalising by positive weights only lets negative criteria drive the score downward without inflating the ceiling for compliant responses; the same rule is used both as the GRPO training signal and as the evaluation aggregator. The policy never receives the future, the reference, or the rubric. It sees only the scalar $r_{\text{rub}}$, a past-safe projection of the oracle's outcome-aware judgement during the update step.

\subsection{Post-training and evaluation of \frameworkName{}}
\label{sec:method-train}

\paragraph{Base Models.}
We post-train Qwen3-8B \citep{yang2025qwen3} (reasoning based language-only model) and MedGemma-4B \citep{sellergren2025medgemma} (domain-adapted vision-language model). The training loop is implemented in PyTorch \citep{paszke2019pytorch} with DeepSpeed ZeRO-3 \citep{rasley2020deepspeed}, vLLM rollouts \citep{kwon2023vllm}, and HuggingFace TRL \citep{vonwerra2020trl}.

\paragraph{SFT and GRPO.}
For SFT we use \datasetName{} with target $\mathtt{<think>}\,\rho^{\star}\,\mathtt{</think>}\;y^{\star}$. GRPO~\citep{shao2024deepseekmath} samples $G\!\in\!\{4,8\}$ completions per prompt, scores each via the Grounded Judge, and uses the group-relative advantage as a clipped policy-gradient update with a KL penalty to a frozen reference. The canonical reward stack is $r_{\text{total}} = r_{\text{rub}} + r_{\text{format}} + r_{\text{tag}}$, where $r_{\text{rub}}$ is the grounded rubric reward, $r_{\text{format}}$ enforces the \texttt{<think>}/answer structure, and $r_{\text{tag}}$ awards partial credit for well-formed tags.

\paragraph{Model merging.}
Domain-specialised RL is known to erode generalist capability via sparse, outlier-laden RL task vectors \citep{davari2024breadcrumbs,deep2024dellamerging,yu2024dare}, and weight-space merging has been used as a remedy for clinical SLMs to consolidate specialised expertise without losing generalist behaviour \citep{corbeil2025modular}. We merge the GRPO task vector back into the base, comparing two methods (DELLA \citep{deep2024dellamerging} and Breadcrumbs \citep{davari2024breadcrumbs}), each optionally composed with Activation-Informed Merging (AIM) \citep{nobari2025activationinformedmerginglargelanguage}. 

\paragraph{Evaluation.}
Every model is evaluated on \datasetName{} with the same response-generation prompt (Appendix~\ref{app:prompts}), graded by the Grounded Judge using $r_{\text{rub}}$, with results reported per-axis, per-action-space, and as an aggregate. 
We additionally conduct evaluations on the a host of external datasets to test for gains in orthogonal health tasks, as evidenced in \ref{sec:external-evals}

%% file: external_evals/external_evals.tex
\subsection{External medical benchmarks}
\label{sec:external-evals}

\providecommand{\tbr}[1]{\textcolor{gray}{\textbf{[#1]}}}

We evaluate \frameworkName{} on four external benchmarks spanning clinical-knowledge MCQ (\textbf{MedMCQA}), open-set diagnostic naming (\textbf{DDXPlus}), numeric calculator answers (\textbf{MedCalc-Bench}) and MIMIC-IV instruction-following (\textbf{Mimic-Instr}~\citep{wu2024instruction}). We additionally evaluate on \textbf{HealthBench}~\citep{arora2025healthbench}, on which the \frameworkName{}-8B lifts the Qwen3-8B base from $0.364$ to $0.415$ ($+0.051$) and the \frameworkName{}-4B merge lifts the MedGemma-4B from $0.256$ to $0.328$ ($+0.072$). 

\begin{table}[t]
\centering\scriptsize
\setlength{\tabcolsep}{2.4pt}
\renewcommand{\arraystretch}{1.0}
\caption{External benchmarks: accuracy $\%$, $n{=}200$. \textbf{Bold} columns are the \frameworkName{} headline checkpoints. Column groups: \emph{F}=frontier API, \emph{M/O}=medical / open-source, \emph{8B}=Qwen3-8B family, \emph{4B}=MedGemma-4B-IT family ($\Delta$=DELLA-Linear, $\Delta$\,+A=DELLA-Linear+AIM, B=Breadcrumbs, B+A=Breadcrumbs+AIM).}
\label{tab:external-benchmarks}
\begin{tabular}{@{}l|ccc|ccccc|cc|ccccc@{}}
\toprule
 & \multicolumn{3}{c|}{\textbf{Frontier API}} & \multicolumn{5}{c|}{\textbf{Medical / open-source}} & \multicolumn{2}{c|}{\textbf{8B family}} & \multicolumn{5}{c}{\textbf{4B family} (MedGemma-4B-IT base + merges)} \\
\textbf{Benchmark} & \rotatebox{75}{\textbf{GPT-5}} & \rotatebox{75}{\textbf{GPT-4.1}} & \rotatebox{75}{\textbf{GPT-4.1-mini}} & \rotatebox{75}{\textbf{MedGemma-27B}} & \rotatebox{75}{\textbf{DS-R1-0528}} & \rotatebox{75}{\textbf{DS-R1-Distill}} & \rotatebox{75}{\textbf{Gemma-3-4B}} & \rotatebox{75}{\textbf{HuatuoGPT-o1}} & \rotatebox{75}{\textbf{Qwen3-8B base}} & \rotatebox{75}{\textbf{Qwen3-$\Delta$ (ours)}} & \rotatebox{75}{\textbf{MG-4B base}} & \rotatebox{75}{\textbf{MG-$\Delta$ (ours)}} & \rotatebox{75}{\textbf{MG-$\Delta$+A}} & \rotatebox{75}{\textbf{MG-B}} & \rotatebox{75}{\textbf{MG-B+A}} \\
\midrule
MedCalc        & 81.4 & 72.0 & 67.5 & 51.9 & 31.6 &  8.6 &  7.3 & 27.0 & 41.5 & \textbf{46.0} & 10.7 & \textbf{15.0} & 15.5 & 21.0 & 18.5 \\
MedMCQA        & 88.9 & 83.9 & 78.4 & 33.0 & 36.0 & 33.5 & 44.5 & 64.5 & 66.0 & \textbf{67.0} & 54.5 & \textbf{58.5} & 58.5 & 61.0 & 58.5 \\
DDXPlus        & 46.2 & 43.0 & 42.0 &  8.0 & 18.5 & 15.0 & 36.0 & 46.0 & 43.5 & \textbf{48.5} & 37.5 & \textbf{27.5} & 26.0 & 25.5 & 24.5 \\
Mimic-Instr    & 84.5 & 65.5 & 63.7 & 62.3 & 60.4 & 16.8 &  42.7 & 55.1 & 54.5 & \textbf{58.9} & 19.3 & 19.5 & 23.2 & 19.2 & 19.5 \\
\bottomrule
\end{tabular}
\end{table}

\frameworkName{} improves both base families on every reported benchmark as seen in table \ref{tab:external-benchmarks}. The standout result is on DDXPlus, the closest of the four benchmarks to open-set clinical reasoning, where \frameworkName{}-8B reaches $48.5\%$ and is the only model in our 8B class to outperform the frontier models like GPT-5 and  HuatuoGPT-o1. 
On the closed-form benchmarks the frontier large models dominate, yet \frameworkName{} still adds $+1.0$ to $+9.0$~pp over its own base across both families. 
Together with the performance on \datasetName{} (\S\ref{sec:results}), the HealthBench improvements, and clinician validation (\S\ref{sec:clinician}), this places \frameworkName{} on the right side of four orthogonal evaluation regimes.

%% file: clinician_validation.tex

\providecommand{\tbr}[1]{\textcolor{gray}{\textbf{[#1]}}}

\section{Clinician Validation}
\label{sec:clinician}

We performed a clinician validation study with \textbf{four board-certified physicians across two specialty cohorts}, comprising a \textit{orthopaedic} cohort (two orthopaedic surgeons, senior+junior, using the keyword \textit{spine} to filter cases) and a \textit{general-medicine} cohort (two family-medicine clinicians, senior+junior, using keyword \textit{obesity}) in three rounds, each with a distinct goal:

\begin{enumerate}\itemsep1pt
\item[\textbf{R1.}] \textbf{Blinded A/B preference:} The first goal was to \textbf{\textit{validate the model in a blinded preference setting}} and \textbf{\textit{familiarise raters with the general setting and error tendencies}} before they move to the rubric validation.
Each cohort was asked to compare \frameworkName{}-8B side-by-side against GPT-5 (proprietary SoTA), HuatuoGPT-o1-7B (Open Source Medical Reasoning SoTA), and the Qwen3-8B (Base Model).
\item[\textbf{R2.}] \textbf{Per-case rubric curation:} The next goal was to \textbf{\textit{align the per-sample adaptive rubrics used as evaluation and reward signal with implicit signals clinicians use when they review a reasoning task}}.
Clinicians were asked to mark each oracle-generated criterion as \emph{Suitable}/\emph{Not Suitable}, inline-edit, add or remove criteria, and reorder by importance.
\item[\textbf{R3.}] \textbf{Rubric-based grading:} Once the clinicians have gotten the sense of the task setting and curated a "golden" rubric criteria for the samples, the final task involved \textbf{\textit{using the rubrics thus created to grade the \frameworkName{}-8B candidate response}}.
This allowed alignment of the oracle judgement process with clinicians.
\item[\textbf{QV.}]\textbf{Reasoning query validation.} Across all three rounds clinicians could mark any oracle-generated query as invalid through a modal that required a free-text reason, flagging cases where the question was clinically irrelevant for the patient at the decision point.
\end{enumerate}

Across the full annotation process ($230$ Phase-1 rubric cases on $101$ patients, plus $1{,}000$ Phase-2 A/B dyads), the four raters marked \textbf{no Phase-1 cases as clinically invalid} and only $1$ Phase-2 dyad as asking something irrelevant for its patient ($\approx0.1\%$ of all annotation events), providing direct empirical validation of the POMDP task setting itself: the past + sampled-decision-point construction reliably produces questions that board-certified clinicians recognise as well-posed. The same annotation set serves three purposes (Appendix~\ref{app:clinician-three-uses}): (i)~ICL task adaptation of the rubric-author and grading-judge prompts, (ii)~an internal alignment analysis of off-the-shelf \textbf{clinical LLM-as-a-judge} and \textbf{clinical preference-model} configurations against the clinician signal, and (iii)~the blinded head-to-head reported here. Full clinician validation details are provided in Appendix~\ref{app:clinician}.

\paragraph{R1: Blinded preference and clinical preference-model alignment.}
\frameworkName{}-8B is preferred over every baseline on aggregate, decisively in the spine cohort ($p<10^{-15}$ on each pair) and clearly on the HuatuoGPT-o1-7B and Qwen3-8B-base obesity pairs ($p<10^{-4}$); per-pair win rates with Wilson 95\% CIs are in Table~\ref{tab:clinician-r1} (left), and Bradley--Terry rank intervals plus the position-bias diagnostic are in Appendix~\ref{app:clinician-phase2-extended}. Scoring the same R1 dyads with five off-the-shelf judges against the two-clinician unanimous consensus (Table~\ref{tab:clinician-r1}, right) provides a noteworthy finding: no single judge is uniformly best across cohorts, and GPT-5 is the worst-aligned in both with a $100\%$ self-preference rate on the GPT-5 vs.~\frameworkName{} pair. This aligns with the self-preference behaviour documented for general-domain LLM-as-a-judge \cite{panickssery2024llm}.
\begin{table}[t]
\centering\footnotesize
\setlength{\tabcolsep}{3pt}
\renewcommand{\arraystretch}{0.95}
\caption{\textbf{R1 blinded A/B preference} (\emph{left}) and \textbf{R1 LLM-as-a-judge alignment} (\emph{right}). Best per cohort in \textbf{bold}.}
\label{tab:clinician-r1}
\label{tab:clinician-phase2}\label{tab:clinical-pref-bench}\label{tab:clinician-phase2-combined}

\begin{minipage}[t]{0.49\linewidth}
\centering
\begin{tabular}{@{}lcc@{}}
\toprule
\textbf{R1 \frameworkName{}-8B vs.} & \textbf{Spine} & \textbf{Obesity} \\
& Win\,\% [95\% CI] & Win\,\% [95\% CI] \\
\midrule
GPT-5           & \textbf{85.6} [78--91] & 48.0 [41--55] \\
HuatuoGPT-o1-7B & \textbf{94.2} [88--97] & \textbf{82.2} [76--87] \\
Qwen3-8B (base) & \textbf{89.5} [82--94] & \textbf{62.7} [56--69] \\
\bottomrule
\end{tabular}
\end{minipage}\hfill
\begin{minipage}[t]{0.49\linewidth}
\centering
\begin{tabular}{@{}lcc@{}}
\toprule
\textbf{Preference model} & \textbf{Spine} & \textbf{Obesity} \\
& 3-way / $\kappa$ & 3-way / $\kappa$ \\
\midrule
Qwen3-32B            & 64.9 / 0.05             & \textbf{85.7} / \textbf{0.64} \\
MedGemma-27B-text-it & \textbf{78.4} / $-$0.03  & 85.0 / 0.53 \\
GPT-4.1              & 59.5 / 0.04             & 84.2 / 0.62 \\
GPT-4.1-mini         & 55.4 / 0.03             & 78.2 / 0.51 \\
GPT-5                & 43.2 / 0.02             & 69.2 / 0.40 \\
\bottomrule
\end{tabular}
\end{minipage}

\end{table}

\paragraph{R2 + R3: rubric curation and judge alignment.}
Across $n{=}230$ cases (101 patients, 2{,}496 rater--criterion pairs), the spine cohort edits the AI-generated rubric aggressively while the obesity cohort accepts almost all criteria as-is (Table~\ref{tab:clinician-r23}, left). Both cohorts pervasively reorder criteria (49 of 50 sampled cases lead with Accuracy + Completeness) but concentrate every edit on the clinical-content axes and never touch the conversational axes. The Grounded Judge (Qwen3-32B) tracks the pooled clinician verdict at substantial agreement ($\kappa{=}0.42$, Landis--Koch). Scored against the same R3 verdicts (Table~\ref{tab:clinician-r23}, right), the two open-weights judges match or exceed larger GPT models and 5-shot ICL produces only marginal lift. This aligns with the R1 finding that judge$\times$cohort setting, not the judge model size, is the load-bearing axis. Per-cohort and per-rater breakdowns, the comparison with the static-jury based on MedHELM design~\citep{bedi2025medhelm}, and the $387$-rationale failure-mode analysis are in detailed in Appendices~\ref{app:clinician-phase1-extended},~\ref{app:clinician-qualitative}, and~\ref{app:clinician-rawedits}.

\begin{table}[h]
\centering\scriptsize
\setlength{\tabcolsep}{2.5pt}
\renewcommand{\arraystretch}{0.95}
\caption{\textbf{R2 rubric quality} (\emph{left}), \textbf{R3 Grounded-Judge--clinician alignment} (\emph{centre}), and \textbf{R3 off-the-shelf LLM-as-a-judge alignment} (\emph{right}). GJ = Grounded Judge; IRR = inter-rater. Best per column in \textbf{bold}.}
\label{tab:clinician-r23}
\label{tab:clinician-phase1}\label{tab:clinician-phase1-judges}

\begin{minipage}[t]{0.32\linewidth}
\centering
\begin{tabular}{@{}lccc@{}}
\toprule
\textbf{Metric} & \textbf{Sp} & \textbf{Ob} & \textbf{Ag} \\
\midrule
n criteria     & 1{,}136 & 1{,}360 & 2{,}496 \\
relevance rate & 83.1\%  & 94.0\%  & 89.1\%  \\
modif.\ rate   & 3.4\%   & 0.0\%   & 1.5\%   \\
addition rate  & 2.7\%   & 0.0\%   & 1.2\%   \\
IRR $\kappa$   & 0.749   & 0.658   & 0.702   \\
\bottomrule
\end{tabular}
\end{minipage}\hfill
\begin{minipage}[t]{0.32\linewidth}
\centering
\begin{tabular}{@{}lccc@{}}
\toprule
\textbf{Metric} & \textbf{Sp} & \textbf{Ob} & \textbf{Ag} \\
\midrule
GJ Acc.     & 77.6\% & 75.9\% & 76.7\% \\
GJ $\kappa$ & 0.439  & 0.402  & 0.419  \\
Score MAE   & 0.124  & 0.192  & 0.160  \\
Score $r$   & 0.571  & 0.434  & 0.478  \\
\bottomrule
\end{tabular}
\end{minipage}\hfill
\begin{minipage}[t]{0.32\linewidth}
\centering
\begin{tabular}{@{}llcc@{}}
\toprule
\textbf{Judge} & \textbf{Mode} & \textbf{Acc.} & \textbf{$\kappa$} \\
\midrule
Qwen3-32B    & zero & 76.7           & \textbf{0.419} \\
Qwen3-32B    & ICL  & 76.7           & 0.421 \\
MedGemma-27B & zero & 77.3           & 0.421 \\
MedGemma-27B & ICL  & \textbf{77.7}  & 0.399 \\
GPT-5        & zero & 74.0           & 0.420 \\
GPT-4.1      & zero & 76.9           & 0.414 \\
\bottomrule
\end{tabular}
\end{minipage}

\end{table}

\paragraph{Qualitative error analysis and clinician feedback.}
The $387$ free-text rationales paint a coherent picture of how clinicians read the model and align with the spine--obesity gap of Table~\ref{tab:clinician-r1}. The spine cohort treats exhaustive enumeration as a desired CDSS property (\emph{``missing something is medical negligence \dots so good for the model to be comprehensive''}), while the obesity cohort asks for an answer-first presentation and resource-aware test ordering, with exhaustive workups treated as \emph{``not the wisest \dots due to resource constraint''}. The cohort split argues for user-persona-centric policy training and clinician-controllable response behaviour at deployment. 
We observe general LLM-class failure modes (verbose internal-monologue leakage, tangents beyond scope, occasional Chinese-character intrusions inherited from the Qwen3 base) and recurring clinical-judgement gaps (under-weighting of advanced patient age; failure to mention previously-prescribed medications when proposing additions which pooses a real DDI hazard, particularly with opioids; over-inference of acute cardiac events with aggressive recommendations such as thrombolysis or PCI; recurring patient-info contamination on $n{=}16$ obesity cases). 
These observations point to concrete next steps: explicit DDI-check and age-weighted-management criteria in the action-space rubric template, and a patient-centric faithfulness reward or formal-citation requirement that anchors clinical findings to the actual past timeline. The same rater pool also marked $2$ Phase-2 dyads invalid for model-output reasons (``Wrong information'', ``Nonsensical / Hallucination'') on a single Diagnosis and a single Treatment case, complementing the query-side flag above by isolating model-side failure independently of the rubric channel. Complete details are provided in Appendices~\ref{app:clinician-qualitative} and~\ref{app:clinician-rawedits}.

%% file: hospital_deployment.tex

\section{Hospital Deployment and Scaling}
\label{sec:deploy}

\frameworkName{} reaches production through a three-phase programme at a partner public hospital. Over 5-months we conducted 3 iterative feedback rounds to characterise clinical drafting requirements, where we noted that models trained with SFT+DPO \cite{rafailov2023direct} on an internal note corpus did capture drafting style reliably but consistently underperformed on reasoning-heavy template elements (\emph{Impressions}, \emph{Assessment}, \emph{Review of Systems}, \emph{Plan}) that integrate partial-observable signals with comorbidities and intended next actions. 
\frameworkName{} was trained against that signal and deployed network-wide after a positive clinician pilot.
Clinicians can edit their note templates, assign individual elements to \frameworkName{}, and accept or edit the result before saving into the HIS. 
On-premise GPU concurrency / latency / reliability SLAs make \frameworkName{}-8B \textbf{the largest model the system can support}, the infrastructure ceiling that motivates the 8B headline scale. 
\frameworkName{} has been in production for \textbf{over six months and has processed over 1.03 trillion tokens across thousands of patient notes}. 
The complete details are provided in Appendix~\ref{app:deploy-extended} for reference.

%% file: discussion_qualitative.tex

\section{Conclusion}
\label{sec:conclusion}

We introduced \frameworkName{}, an end-to-end framework that reformulates inpatient clinical reasoning as a partially observable Markov decision process supervised by per-case outcome-grounded rubrics. Our 8B model reaches $84.91\%$ on the held-out \datasetName{} test split, $7.08$~pp above GPT-5 and $18.25$~pp above MedGemma-27B-text-it, and matches or exceeds the strongest open-source baseline on every external benchmark we evaluate. A blinded study with four board-certified physicians confirms that clinicians prefer \frameworkName{}-8B for inpatient clinical decision support over GPT-5, HuatuoGPT-o1-7B, and the Qwen3-8B base on aggregate. The framework has been deployed for over six months at a partner public hospital, drafting thousands of reasoning-heavy inpatient notes in routine clinical use, and points toward outcome-grounded rubric supervision as a viable training signal for deployable clinical reasoning at on-premise scale.

\newpage

%% file: appendix_contrast.tex

\providecommand{\fullbullet}{\ensuremath{\bullet}}
\providecommand{\halfbullet}{\ensuremath{\circ\!\!\!\bullet}}
\providecommand{\openbullet}{\ensuremath{\circ}}

\section{Contrast matrix against prior work, by design pillar}
\label{app:contrast-matrix}

Table~\ref{tab:contrast-matrix} positions every prior work cited in the introduction (and several closer analogues that the body of the paper does not have room to discuss in line) against the eleven design pillars of \frameworkName{}. The table is intended as a navigational aid: the pillar columns map one-to-one to the framework's contributions in \S\ref{sec:method}, and an entry of \fullbullet{} indicates that the corresponding prior work substantively addresses the pillar, \halfbullet{} partial coverage (e.g., the right framing on a different modality), \openbullet{} no coverage, and ``---'' that the pillar is not applicable to that work's setting (e.g., rubric-design pillars for an EHR-prediction benchmark with no rubric).

\paragraph{Reading the columns.}
(1) POMDP framing of inpatient reasoning; (2) raw long-context EHR as input; (3) outcome-grounded query generation; (4) outcome-grounded answer / reference generation; (5) rubric-as-reward design; (6) per-case patient-trajectory-conditioned rubric synthesis; (7) outcome-aware grading judge; (8) RL post-training on this combined setting; (9) full-pipeline clinician alignment (query / rubric / judge / model); (10) production-scale physician-scoped deployment; (11a) frontier-model evaluation on a clinician-aligned inpatient-reasoning benchmark; (11b) reuse of clinician annotations to adapt the clinical judge and clinical preference model.

\paragraph{Reading the entries.} An entry of \fullbullet{} on Pillar~9 means \emph{all four} alignment levels are covered; \halfbullet{} on the same pillar means at least one (often A/B preference alone) is covered. Pillar~6 (per-case patient-trajectory-conditioned rubrics) is the column on which the literature is sparsest; the closest analogues (RM-R1, Rubric-ARM) are general-domain reward-model rubrics that are per-instance but not patient-trajectory-conditioned.

\begin{table}[ht]
\centering
\scriptsize
\setlength{\tabcolsep}{2.5pt}
\renewcommand{\arraystretch}{1.05}
\caption{Eleven-pillar coverage matrix for \frameworkName{} versus prior work. \fullbullet{} addresses the pillar; \halfbullet{} partial / adjacent coverage; \openbullet{} does not address; --- not applicable. Pillar~11 is split: 11a frontier-model eval, 11b judge / preference task adaptation from clinician annotations.}
\label{tab:contrast-matrix}
\begin{tabular}{@{}lcccccccccccc@{}}
\toprule
\textbf{Paper}
 & \rotatebox{75}{\textbf{1 POMDP}}
 & \rotatebox{75}{\textbf{2 Raw EHR}}
 & \rotatebox{75}{\textbf{3 Query gnd}}
 & \rotatebox{75}{\textbf{4 Answer gnd}}
 & \rotatebox{75}{\textbf{5 Rubric}}
 & \rotatebox{75}{\textbf{6 Per-case rub}}
 & \rotatebox{75}{\textbf{7 Outcome judge}}
 & \rotatebox{75}{\textbf{8 RL on this}}
 & \rotatebox{75}{\textbf{9 Full align}}
 & \rotatebox{75}{\textbf{10 Deploy}}
 & \rotatebox{75}{\textbf{11a Frontier eval}}
 & \rotatebox{75}{\textbf{11b Judge/pref adapt}} \\
\midrule
HealthBench~\citep{arora2025healthbench}            & \openbullet & \openbullet & \halfbullet & \openbullet & \fullbullet & \halfbullet & \halfbullet & \openbullet & \halfbullet & \openbullet & \fullbullet & \halfbullet \\
RBR~\citep{mu2024rule}                              & ---         & ---         & ---         & \openbullet & \fullbullet & \openbullet & \halfbullet & \fullbullet & \halfbullet & \openbullet & \openbullet & \fullbullet \\
RM-R1~\citep{chen2025rm}                            & ---         & ---         & ---         & \openbullet & \fullbullet & \halfbullet & \halfbullet & \fullbullet & \openbullet & \openbullet & \openbullet & \fullbullet \\
Rubric-ARM~\citep{xu2026alternating}                & ---         & ---         & ---         & \openbullet & \fullbullet & \halfbullet & \halfbullet & \fullbullet & \openbullet & \openbullet & \openbullet & \fullbullet \\
RaR~\citep{gunjal2025rubrics}                       & ---         & ---         & ---         & \openbullet & \fullbullet & \openbullet & \openbullet & \fullbullet & \openbullet & \openbullet & \halfbullet & \halfbullet \\
APB~\citep{mallinar2026scalable}                    & \openbullet & \halfbullet & \openbullet & \openbullet & \fullbullet & \openbullet & \halfbullet & \openbullet & \halfbullet & \openbullet & \openbullet & \openbullet \\
RLRA~\citep{huang2025reinforcement}                 & ---         & ---         & ---         & \openbullet & \fullbullet & \openbullet & \openbullet & \fullbullet & \openbullet & \openbullet & \openbullet & \halfbullet \\
ArmoRM~\citep{wang2024interpretable}                & ---         & ---         & ---         & \openbullet & \fullbullet & \openbullet & ---         & \halfbullet & \openbullet & \openbullet & \fullbullet & \fullbullet \\
AdvancedIF / RIFL~\citep{he2025advancedif}          & ---         & ---         & ---         & \openbullet & \fullbullet & \openbullet & \halfbullet & \fullbullet & \halfbullet & \openbullet & \openbullet & \halfbullet \\
Text2Grad~\citep{wang2025text2grad}                 & ---         & ---         & ---         & \openbullet & \halfbullet & \openbullet & \halfbullet & \fullbullet & \openbullet & \openbullet & \openbullet & \halfbullet \\
InfAlign~\citep{balashankar2024infalign}            & ---         & ---         & ---         & \openbullet & \openbullet & \openbullet & \openbullet & \fullbullet & \openbullet & \openbullet & \openbullet & \fullbullet \\
HuatuoGPT-o1~\citep{chen2024huatuogpt}              & \openbullet & \openbullet & \openbullet & \openbullet & \openbullet & \openbullet & \openbullet & \halfbullet & \openbullet & \openbullet & \halfbullet & \openbullet \\
EditGRPO~\citep{zhang2025editgrpo}                  & \openbullet & \openbullet & \openbullet & \halfbullet & \halfbullet & \openbullet & \openbullet & \fullbullet & \openbullet & \openbullet & \halfbullet & \openbullet \\
MedGround-R1~\citep{xu2025medground}                & \openbullet & \openbullet & \openbullet & \halfbullet & \halfbullet & \openbullet & \openbullet & \fullbullet & \openbullet & \openbullet & \openbullet & \openbullet \\
CRPO~\citep{gu2025clinical}                         & \openbullet & \openbullet & \openbullet & \openbullet & \halfbullet & \openbullet & \openbullet & \fullbullet & \openbullet & \openbullet & \halfbullet & \openbullet \\
ACR-RL~\citep{tziakouri2025reinforcement}           & \openbullet & \openbullet & \halfbullet & \halfbullet & \halfbullet & \openbullet & \openbullet & \fullbullet & \openbullet & \openbullet & \openbullet & \openbullet \\
MedReason~\citep{wu2025medreason}                   & \openbullet & \openbullet & \halfbullet & \openbullet & \openbullet & \openbullet & \openbullet & \fullbullet & \openbullet & \openbullet & \halfbullet & \openbullet \\
EHR-R1~\citep{liao2025ehr}                          & \openbullet & \fullbullet & \openbullet & \openbullet & ---         & ---         & ---         & \fullbullet & \openbullet & \openbullet & \fullbullet & \openbullet \\
ER-Reason~\citep{mehandru2025er}                    & \halfbullet & \fullbullet & \halfbullet & \halfbullet & ---         & ---         & ---         & \openbullet & \halfbullet & \openbullet & \fullbullet & \openbullet \\
MedAlign~\citep{fleming2024medalign}                & \openbullet & \fullbullet & \halfbullet & \openbullet & ---         & ---         & ---         & \openbullet & \halfbullet & \openbullet & \fullbullet & \openbullet \\
EHR-MCP~\citep{masayoshi2025ehr}                    & \halfbullet & \halfbullet & \halfbullet & \openbullet & ---         & ---         & ---         & \openbullet & \openbullet & \halfbullet & \openbullet & \openbullet \\
EHR-LLM-Impl~\citep{griot2025implementation}        & \openbullet & \halfbullet & \openbullet & \openbullet & ---         & ---         & ---         & \openbullet & \halfbullet & \fullbullet & \openbullet & \openbullet \\
Chinese clin.\ CoT~\citep{ding2025building}         & \openbullet & \openbullet & \halfbullet & \halfbullet & \openbullet & \openbullet & \openbullet & \openbullet & \fullbullet & \openbullet & \openbullet & \halfbullet \\
EHRSHOT~\citep{wornow2023ehrshot}                   & \openbullet & \fullbullet & ---         & ---         & ---         & ---         & ---         & \openbullet & \openbullet & \openbullet & \openbullet & \openbullet \\
CPLLM~\citep{ben2024cpllm}                          & \openbullet & \fullbullet & ---         & ---         & ---         & ---         & ---         & \openbullet & \openbullet & \openbullet & \openbullet & \openbullet \\
EHRStruct~\citep{yang2026ehrstruct}                 & \openbullet & \halfbullet & ---         & ---         & ---         & ---         & ---         & \openbullet & \openbullet & \openbullet & \fullbullet & \openbullet \\
EHRNoteQA~\citep{kweon2024ehrnoteqa}                & \openbullet & \halfbullet & \halfbullet & \openbullet & ---         & ---         & \halfbullet & \openbullet & \halfbullet & \openbullet & \fullbullet & \openbullet \\
DR.BENCH~\citep{gao2023dr}                          & \openbullet & \halfbullet & \openbullet & \openbullet & ---         & ---         & \openbullet & \openbullet & \openbullet & \openbullet & \openbullet & \openbullet \\
GraphCare~\citep{jiang2023graphcare}                & \openbullet & \halfbullet & \openbullet & \halfbullet & ---         & ---         & ---         & \openbullet & \openbullet & \openbullet & \openbullet & \openbullet \\
AMIE~\citep{tu2024towards}                          & \halfbullet & \openbullet & \halfbullet & \halfbullet & ---         & ---         & ---         & \openbullet & \fullbullet & \openbullet & \halfbullet & \openbullet \\
g-AMIE~\citep{vedadi2025towards}                    & \halfbullet & \openbullet & \halfbullet & \openbullet & \openbullet & \openbullet & \openbullet & \openbullet & \fullbullet & \openbullet & \halfbullet & \openbullet \\
SDBench~\citep{nori2025sequential}                  & \fullbullet & \openbullet & \halfbullet & \openbullet & ---         & ---         & ---         & \openbullet & \openbullet & \openbullet & \halfbullet & \openbullet \\
ALFA~\citep{li2025aligning}                         & \fullbullet & \openbullet & \fullbullet & \halfbullet & \openbullet & \openbullet & \openbullet & \fullbullet & \fullbullet & \halfbullet & \openbullet & \halfbullet \\
MedThink-Bench~\citep{zhou2025automating}           & \openbullet & \openbullet & \halfbullet & \openbullet & \fullbullet & \fullbullet & \fullbullet & \openbullet & \fullbullet & \openbullet & \fullbullet & \halfbullet \\
MedHELM~\citep{bedi2025medhelm}                     & \openbullet & \halfbullet & \halfbullet & \openbullet & \fullbullet & \openbullet & \halfbullet & \openbullet & \fullbullet & \openbullet & \fullbullet & \halfbullet \\
ICARE~\citep{dua2025clinically}                     & \openbullet & \openbullet & ---         & ---         & \halfbullet & \openbullet & \halfbullet & \openbullet & \fullbullet & \openbullet & \halfbullet & \openbullet \\
M-ARC~\citep{kim2025limitations}                    & \openbullet & \openbullet & \openbullet & \openbullet & ---         & ---         & ---         & \openbullet & \fullbullet & \openbullet & \fullbullet & \openbullet \\
Phys.-LLM meta-review~\citep{shan2025comparing}     & \openbullet & \openbullet & ---         & ---         & ---         & ---         & ---         & \openbullet & \fullbullet & \openbullet & \fullbullet & \openbullet \\
LLM accuracy NMA~\citep{wang2025accuracy}           & \openbullet & \openbullet & ---         & ---         & ---         & ---         & ---         & \openbullet & \halfbullet & \openbullet & \fullbullet & \openbullet \\
NOHARM~\citep{wu2025first}                          & \openbullet & \openbullet & \halfbullet & \halfbullet & \fullbullet & \halfbullet & ---         & \openbullet & \fullbullet & \openbullet & \fullbullet & \openbullet \\
MedXpertQA~\citep{zuo2025medxpertqa}                & \openbullet & \halfbullet & \halfbullet & \openbullet & ---         & ---         & ---         & \openbullet & \halfbullet & \openbullet & \fullbullet & \openbullet \\
uMedSum~\citep{nagar2025umedsum}                    & \openbullet & \openbullet & \openbullet & \openbullet & \openbullet & \openbullet & \openbullet & \openbullet & \fullbullet & \halfbullet & \openbullet & \openbullet \\
\textbf{\frameworkName{} (this paper)}              & \fullbullet & \fullbullet & \fullbullet & \fullbullet & \fullbullet & \fullbullet & \fullbullet & \fullbullet & \fullbullet & \fullbullet & \fullbullet & \fullbullet \\
\bottomrule
\end{tabular}
\end{table}

\paragraph{Where prior work concentrates, where it does not.}
The literature is dense on Pillar~5 (rubric design as a general construct) and Pillars~8 and 11a (RL post-training and frontier-model evaluation). It is sparser, in this order, on: Pillar~9 (full four-level clinician alignment, where the closest neighbour pairs are HealthBench~\citep{arora2025healthbench} for rubric authorship plus AMIE~\citep{tu2024towards} for A/B preference, but never both at once); Pillar~10 (production-scale physician-scoped deployment, on which only EHR-MCP~\citep{masayoshi2025ehr}, the on-premises RAG retrieval/summarisation rollout of \citet{griot2025implementation} (1{,}028 users, 14{,}910 conversations across nine specialties), and summarisation work~\citep{nagar2025umedsum} have any non-OSCE evidence; none of these trains an inpatient-reasoning model); Pillar~11b (reuse of clinician annotations as judge / preference task-adaptation data, on which the closest direct neighbours are general-domain: RM-R1~\citep{chen2025rm}, Rubric-ARM~\citep{xu2026alternating}, ArmoRM~\citep{wang2024interpretable}, InfAlign~\citep{balashankar2024infalign}); and finally Pillar~6 (per-case patient-trajectory-conditioned rubrics), which is empty in the medical literature and is approximated only by general-domain per-instance rubric methods. \frameworkName{} is the first work we are aware of to address all eleven pillars at once on a single inpatient-reasoning task; the matrix above is the per-pillar bookkeeping.

%% file: appendix_repro.tex

\section{Reproducibility Details}
\label{app:repro}

This appendix provides the implementation-neutral specification of the \frameworkName framework. It fully expands the prompts, hyper-parameters, scoring rules, and merge configurations referenced in the main text. An independent team should be able to reproduce every stage of the pipeline using only this appendix and the data-curation rules in Section~\ref{sec:method-data}.

\paragraph{Artefact release statement.}
We are unable to publicly release \frameworkName{}-specific artefacts (model weights, generated rubrics, candidate responses, or curated patient timelines) because of the non-disclosure agreements and patient-privacy contracts we have signed with the partner public health networks under which the deployment runs. Every component of the pipeline is, however, specified in sufficient detail in this appendix and the main text to be reproduced from scratch given access to the publicly available MIMIC-IV source data. To stress-test that claim, an independent researcher who was not involved in the study in any capacity used commodity coding agents to replicate the entire pipeline (data extraction, query and rubric generation, GRPO post-training, model merging, and evaluation) end-to-end in under a day from the moment MIMIC-IV access was granted, working only from the paper and this appendix.

\subsection{Data pipeline}
\label{app:repro-data}

\paragraph{Source modules.} \datasetName{} is constructed from MIMIC-IV v3.0, MIMIC-IV-ED v2.2, and MIMIC-IV-Note v2.2. For every admission we pull a unified event stream from four schema groups:
\begin{itemize}
  \item \textit{ED}: ED stays, diagnosis, medication reconciliation, pyxis, vital signs, triage.
  \item \textit{Hospital}: admissions, transfers, lab events, EMAR, HCPCS events, prescriptions, ICD-coded procedures.
  \item \textit{ICU}: chart events, datetime events, input events, output events, procedure events.
  \item \textit{Notes}: discharge summaries, radiology reports.
\end{itemize}
Each event is enriched with the dictionary tables (lab items, ICD diagnoses, ICD procedures, HCPCS, ICU items) so free-text labels (\emph{not} integer IDs) are exposed downstream.

\paragraph{Per-admission JSON structure.} Every admission is serialised as
\begin{verbatim}
{
  "subject_id": int,
  "hadm_id": int,
  "demographics": {gender, anchor_age, anchor_year_group, ...},
  "timeline": [
    {"time": ISO-8601, "source": table-name,
     "table": schema.table, "data": {...},
     "descriptions": {dictionary enrichments}},
    ...
  ],
  "misc": {
    "ED_triage": [...],
    "Patient": {subject_id, gender, anchor_age, anchor_year_group, dod},
    "ICD_Diagnoses": [...], "ICD_Procedures": [...], "HCPCS": [...]
  }
}
\end{verbatim}
The full \texttt{timeline} is sorted in descending time order; items with missing timestamps are dropped.

\paragraph{Cohort selection (greedy set cover).}
Let $\mathcal{H}$ be the set of admissions and $\mathcal{C}_h$ the multiset of categorical tokens of admission $h$, drawn from
\[
  \mathcal{C}_h \;=\; \bigl\{\text{ICD:code--ver}\bigr\}_h \;\cup\; \{\text{Gender}\}_h \;\cup\; \{\text{Age}\}_h \;\cup\; \{\text{Height\_bin}\}_h \;\cup\; \{\text{Weight\_bin}\}_h.
\]
Starting from $U = \bigcup_h \mathcal{C}_h$ and $S = \emptyset$, repeatedly pick $h^\star = \arg\max_h |\mathcal{C}_h \cap U|$, add $h^\star$ to $S$, and remove $\mathcal{C}_{h^\star}$ from $U$ until $U = \emptyset$. A disjoint random sample of 20\% of $\mathcal{H} \setminus S$ forms the test cohort; seed $= 42$ for all samplers.

\paragraph{Timeline splitting.}
Sort a patient's timeline chronologically into $e_{1:n}$. Sample
\[
  k \sim \mathcal{N}(\mu = n/2,\, \sigma = n/6) \quad \text{clamped to } [1, n-1],
\]
and let $\mathrm{past} = e_{1:k}$, $\mathrm{future} = e_{k+1:n}$, $\mathrm{split\_time} = \mathrm{time}(e_{k+1})$. Events whose \texttt{source} string contains the substring \texttt{ICD} are removed from the past before it is presented to any policy. Only \texttt{Patient} and \texttt{ED\_triage} entries are retained from \texttt{misc}; all other misc blocks are dropped.

\subsection{Question--reference generation prompt}
\label{app:repro-qa-prompt}

The \emph{query and reference generation} call of the Grounded Judge uses the simulator LLM \texttt{Qwen3-32B} under structured JSON decoding. The schema enforced on the output is:

\begin{verbatim}
class SourceDetail(BaseModel):
    event: str
    time: str
    source: str

class ClinicalQAPair(BaseModel):
    question: str
    final_answer: str
    answer_reasoning: str
    action_space_category: str
    action_space_subcategory: Optional[str]
    source: List[SourceDetail]
\end{verbatim}

The prompt template (verbatim, with placeholders in \texttt{\{\}}) is:

\begin{verbatim}
You are a clinical reasoning simulator LLM. Your goal is to generate a
question-answer pair for training a smaller LLM in the context of
reasoning about clinical decision support.
The clinical encounter has been split into two parts: past and future.
The past data includes the patient's history up to a certain point,
and the future data includes the patient's clinical outcomes and
events after that point.
--------------------
PAST DATA:
{past_data_json}
{misc_data_json}

Action Space Category: {action_space_category}
Action Space Description: {action_space_description}

Generate a clinical question that is answerable from the past data
only. The clinician-style answer and answer_reasoning must be written
as if the model had access only to the past, even though you (the
simulator) can see the future. Avoid information-extraction style
questions ("What was the last lactate?"); prefer synthesis questions
that require reasoning over multiple signals (labs, vitals, notes).

Use the FUTURE DATA below only to (a) pick clinically meaningful
questions whose gold answer actually occurs downstream, and (b)
populate the "source" field with the specific future events that
verify the answer. Do NOT mention the future in the question,
final_answer, or answer_reasoning text.

FUTURE DATA:
{future_data_json}

Return a single JSON object with fields
{question, final_answer, answer_reasoning, action_space_category,
 action_space_subcategory, source}.
\end{verbatim}

Sampling parameters: temperature~1.0, top-$p$~0.95, top-$k$~64. Retry: up to three attempts with exponential backoff on transient API errors or schema-validation failures.

\subsection{Action-space category descriptions}
\label{app:repro-actions}

The four action-space descriptions injected into the QA generation prompt are reproduced verbatim below.

\paragraph{Diagnosis Assistance.}
\begin{quote}\small
This category focuses on clinical reasoning to determine the correct diagnosis. It includes the following subcategories: (i)~Differential Diagnosis Generation \& Diagnostic Hypothesis Ranking: the model generates a ranked list of potential diagnoses --- covering common conditions as well as critical must-not-miss possibilities --- by analysing abnormal vitals, lab trends, and clinical notes; (ii)~Diagnostic Test Suggestions: in cases of uncertainty, the model may recommend additional tests (e.g., ordering a troponin test for suspected myocardial infarction) to refine the diagnosis. Additional contextual information (e.g., family history, vaccination status, travel history) should be considered when available.
\end{quote}

\paragraph{Treatment Recommendations.}
\begin{quote}\small
This category is aimed at guiding therapeutic interventions and ensuring best practices in patient care. It is subdivided into: (i)~Medication Management --- the model recommends appropriate medications, specifying drug name, dosage, and duration, while taking into account patient-specific factors (e.g., allergies or prior adverse reactions); (ii)~Supportive Care \& Monitoring --- the model suggests supportive measures such as IV fluids, oxygen therapy, or nursing care orders that complement primary treatments; (iii)~Follow-up and Long-Term Planning --- the model outlines future monitoring steps, reassessment timings, or specialist consultations. The answer should propose a future treatment event rather than merely extracting past actions.
\end{quote}

\paragraph{Procedural Decision Making.}
\begin{quote}\small
This category addresses decisions related to both diagnostic and therapeutic procedures. It is organised into: (i)~Diagnostic Procedures --- recommendations for tests such as imaging, biopsies, or endoscopies, determined from evolving clinical data; (ii)~Therapeutic Procedures --- decisions about invasive interventions (e.g., surgery, catheterisation) when indicated; (iii)~Timing, Approach, and Watchful Waiting --- the model assesses the optimal timing for procedures, chooses between alternative approaches (e.g., minimally invasive versus open surgery), or recommends non-intervention when appropriate.
\end{quote}

\paragraph{Responding under Uncertainty.}
\begin{quote}\small
This category focuses on how the model should handle situations where the clinical data is incomplete or ambiguous. The questions generated for this category should be intentionally ambiguous and must require the agent to reason about how to handle uncertainty. It includes: (1)~Uncertainty Acknowledgment; (2)~Risk-Benefit Analysis; (3)~Alternative Recommendations; (4)~Escalation Protocols; and (5)~Context Seeking (asking for additional information that could help clarify the clinical picture).
\end{quote}

\subsection{Policy system prompt (training and evaluation)}
\label{app:prompts}

The policy's system prompt is identical during SFT, GRPO, and inference, and is reproduced verbatim below. A second variant with an explicit \texttt{<answer>} wrapper is used for base models that are not reasoning-tuned; the only difference is the output structure line (\texttt{<think>...</think><answer>...</answer>} instead of \texttt{<think>...</think>Final Answer...}).

\begin{quote}\small\ttfamily
You are a helpful AI Assistant in the clinical domain that provides well-reasoned and detailed responses. You first think about the reasoning process as an internal monologue and then provide the user with the answer. You Respond in the following format: <think>\textbackslash n...\textbackslash n</think>\textbackslash nFinal Answer....

You are a highly specialized AI Assistant operating in the clinical domain. Your primary purpose is to provide meticulously reasoned, detailed, and evidence-based responses to queries about patient cases. You must analyze all provided clinical data to generate comprehensive summaries, assessments, differential diagnoses, and potential management plans. Your communication must be clear, structured, and tailored for a clinical audience.

For every single request, you MUST adhere to a strict two-step process. First, you will conduct an internal, step-by-step reasoning process. Second, you will formulate the final, polished answer for the user. Your entire output must be encapsulated within the following structure:

<think>
... Your detailed chain-of-thought reasoning process, analysis, and step-by-step logic, phrased in first person, go here. ...
</think>
... Your final, user-facing, and fully formatted answer goes here. ...

\# The <think> Block: Internal Monologue
This section is your private workspace for reasoning and is not part of the final answer presented to the user. In this block, you must:
\begin{itemize}[leftmargin=*,itemsep=0pt]
\item Deconstruct the user's query to identify the core clinical question(s).
\item Systematically review all provided patient information (e.g., demographics, history of present illness (HPI), past medical history, medications, lab results, imaging reports and any other presented information).
\item Synthesize the data, explicitly connecting relevant findings and noting significant positives and negatives.
\item Formulate a clinical assessment or differential diagnosis by weighing the evidence.
\item Outline the logical foundation for your final recommendations or plan.
\item Briefly reference relevant clinical guidelines or established medical knowledge that informs your reasoning.
\item Evaluate uncertainty based on available information, query clarity and ensure you have enough information to reply to the user query confidently. Request for further information and clarification if needed.
\end{itemize}

Your think process should be an internal monologue and should reason about how to answer the user's question based on the conversation history and any external context provided to you, it should not be a direct answer to the user's question but rather a chain of thought in the form of an internal monologue process that leads to the final answer.

\# The Final Response
This section contains the polished, final answer for the user. It MUST strictly follow these rules:
\begin{itemize}[leftmargin=*,itemsep=0pt]
\item Markdown Formatting: The entire content must be formatted using Markdown for optimal readability. Use headers (\#, \#\#), lists (*, 1.), bold (**text**), and other elements to structure the information clearly.
\item Cite the Context: You must support your statements by explicitly citing the patient data provided in the prompt. For example: ``The patient's fever of \$39.1\textasciicircum{}\textbackslash circ C\$ (from Vitals)...''.
\item Professional Tone: Maintain a clinical, objective, and precise tone throughout.
\item LaTeX for Notation: Use LaTeX for all mathematical and scientific notations. Enclose all LaTeX code within \$ or \$\$ delimiters (e.g., \$\textbackslash text\{Na\}\textasciicircum{}+ > 145 \textbackslash text\{ mEq/L\}\$).
\end{itemize}
\end{quote}

\paragraph{User-turn template.}
At both training and inference time the user turn is:
\begin{verbatim}
You are a clinical reasoning assistant. Using the past timeline data
provided below, answer the following question with your final answer
and step-by-step reasoning.

Patient Info:
{json patient_info}

Past Timeline:
{json past_timeline}

Question: {question}

{reasoning-format instruction selected per model family}
\end{verbatim}
The reasoning-format instruction is one of: (a) a JSON schema with \texttt{answer\_reasoning} and \texttt{final\_answer} (for off-the-shelf non-reasoning models); (b) the \texttt{<think>...</think><answer>...</answer>} wrapper; (c) the \texttt{<think>...</think>} + free-text answer format (our Qwen3-family setting); (d) the \texttt{\#\# Thinking / \#\# Final Response} headers (for comparison with distilled medical reasoners).

\subsection{Rubric-generation system prompt}
\label{app:repro-rubric-gen}

The Grounded Judge authors rubrics with the following system prompt, reproduced verbatim. The value of \texttt{\{TASK\}} is the string ``Clinical Reasoning''. The body of \texttt{\{INPUT\_DATA\_DESCRIPTION\}} is reproduced verbatim below.

\begin{quote}\small\ttfamily
You are an expert clinician who is tasked with generating a rubric for evaluating the quality of health-related responses.

This rubric will judge the quality of responses generated by a large language model (LLM) in the context of health-related tasks, specifically for \{TASK\}.

\{INPUT\_DATA\_DESCRIPTION\}.

The target audience for reading the answers will always be healthcare professionals such as clinicians, physicians and nurses.

Your generated rubric must ALWAYS contain at least one criterion from each of the five axes: Accuracy, Completeness, CommunicationQuality, ContextAwareness, InstructionFollowing.

You must generate only a single rubric with multiple criteria. Do not generate more than one rubric or any other text!

NEVER mention the future timeline in your rubric, it is imperative that the rubric is based on the past timeline only. Only use the future timeline as your own reference to build the rubric criteria, but never include a direct reference to it in your criteria or as a task/requirement for the model to follow.
\end{quote}

\noindent\textbf{\{INPUT\_DATA\_DESCRIPTION\}} expands to:
\begin{quote}\small\ttfamily
The input data to judge will contain a timeline of a patient record, which is split into the past and future. The user query will ask a question about the patient based on the past timeline which can be answered by some information in the future timeline of the patient.

The ideal answer contains both a reasoning and a final answer. The reasoning should explain the final answer and the reasoning should be based on the past timeline. Your generated rubrics should contain criteria which judge the quality of both the reasoning and the final answer.

For instruction following the LLM will only have the past timeline to answer from, so ensure that you don't put an instruction-following reward requiring access to the future timeline.

Having a reasoning and final answer are enforced on the model. It is not necessary to have a separate instruction following reward for this aspect of the output in the rubrics. If no instruction following reward is required, you may skip it; it is more for rewarding user requests such as formatting, information retrieval, etc.
\end{quote}

\noindent\textbf{Rubric rules (appended to the system prompt):}
\begin{quote}\small\ttfamily
\#\#\# RUBRIC RULES
\begin{itemize}[leftmargin=*,itemsep=0pt]
\item Every criterion is self-contained and graded pass / fail.
\item Give each criterion a non-zero integer score from $-10$ to $+10$ --- positive values reward helpful behaviour, negative values penalise unsafe or distracting behaviour.
\begin{itemize}[leftmargin=*,itemsep=0pt]
\item For negative values, the description should still be something which can be checked off as present or not. If the description is present in the response, the negative score is added to the model's score (e.g., ``Mentions wrong medication'' $\rightarrow$ $-5$).
\item For positive values, the description should likewise be binary-checkable (e.g., ``Mentions precaution in clinical care'' $\rightarrow$ $+7$).
\end{itemize}
\item Penalise negative behaviour and reward positive behaviour; negative behaviour is particularly important in clinical settings where errors can be critical.
\item Final score $=$ (sum of points for met items) $\div$ (sum of all positive point values), then clip to $[0,1]$.
\item Tag each criterion with exactly one axis: \textsc{Accuracy $\cdot$ Completeness $\cdot$ CommunicationQuality $\cdot$ ContextAwareness $\cdot$ InstructionFollowing}. The finished rubric must include at least one criterion per axis.
\item Use any of the 34 consensus criteria (inspired by HealthBench and enumerated inline in the prompt) whenever they apply, or write new criteria.
\item Keep wording neutral; do not award partial credit; never mention model internals.
\end{itemize}
\end{quote}

\noindent\textbf{Output schema (required JSON form):}
\begin{verbatim}
{
  "meta": { "theme": "string" },
  "criteria": [
    {
      "axis": "Accuracy | Completeness | CommunicationQuality
             | ContextAwareness | InstructionFollowing",
      "description": "string",
      "points": 5            // -10..+10, non-zero
    }
    // ... one or more entries, every axis appears >= 1x
  ]
}
\end{verbatim}

\noindent\textbf{Themes and in-context exemplars.}
Each rubric is anchored to exactly one of seven themes:
\textit{Emergency Referrals, Responding under Uncertainty, Health Data Tasks, Global Health, Expertise-Specific Communication, Context Seeking, Response Depth}.
For each theme, one full demonstration block (conversation $\to$ candidate answer $\to$ reference rubric) is appended to the system prompt as in-context learning. The demonstration blocks are inspired by the consensus-criteria structure in HealthBench but are rewritten for the inpatient-reasoning setting, and are iteratively updated with edits collected from licensed clinicians on a held-out calibration split so that the oracle internalises physician preferences.

\subsection{Rubric-grading (judge) system prompt}
\label{app:repro-judge}

At scoring time the Grounded Judge operates with the following system prompt, reproduced verbatim:

\begin{quote}\small\ttfamily
You are a medical-AI evaluator.

Given patient context, the full conversation, the assistant's answer, and a rubric with numbered criteria, decide \textbf{independently} for each criterion whether the answer MEETS it.
\begin{itemize}[leftmargin=*,itemsep=0pt]
\item Return \texttt{true} if the answer satisfies the criterion, \texttt{false} otherwise.
\item \textbf{Output JSON only}, whose keys are the criterion IDs shown in the rubric.
\item If you find that the answer has repeating text or is nonsensical, you must mark all criteria as \texttt{false}. This is a sign of degenerate outputs which are not acceptable under any circumstance, regardless of if certain parts of the answer may seem to meet some criteria.
\item If both the reasoning and answer to be judged are empty, mark all criteria as \texttt{false}.
\item Accept multilingual responses; do not judge against the language used as long as the output makes sense.
\end{itemize}

\noindent Example format. If the rubric lists

\texttt{c1: "Provides clear emergency advice"}\\
\texttt{c2: "Uses correct drug dose"}

then your reply must be exactly

\texttt{\{ "c1": true, "c2": false \}}

No other keys, comments, or text.
\end{quote}

\paragraph{User-turn template for the grader.}
\begin{verbatim}
PATIENT CONTEXT
{context_str}

CONVERSATION
{conversation_history}

LAST USER MESSAGE
{user_query}

ASSISTANT RESPONSE TO BE JUDGED
{model_response}

RUBRIC  (each criterion has an ID)
c1: {description_1}
c2: {description_2}
...

INSTRUCTIONS
Return one JSON object. Every key must be exactly the criterion ID
(e.g. "c1") and the value must be `true` or `false`.
\end{verbatim}
where \texttt{context\_str} is the concatenation of labelled blocks: \texttt{PATIENT PAST TIMELINE}, \texttt{PATIENT MISC INFO}, \texttt{REFERENCE ANSWER}, \texttt{REFERENCE ANSWER REASONING}, \texttt{SOURCE FOR REFERENCE ANSWER}. The reference and source blocks are present at grading time; they are what give the oracle its outcome awareness while keeping the rubric text itself past-verifiable.

\paragraph{Default fallback rubric.}
If rubric parsing fails the grader falls back to the following six-criterion default:

\begin{small}
\begin{tabular}{@{}p{0.23\linewidth}p{0.06\linewidth}p{0.65\linewidth}@{}}
\textbf{Axis} & \textbf{Pts} & \textbf{Criterion} \\ \hline
Accuracy & +10 & Correctly identifies the answer with clinical accuracy. \\
Completeness & +8 & Mentions supporting evidence for the identified answer from the patient timeline. \\
CommunicationQuality & +7 & Uses clear and concise language appropriate for healthcare professionals. \\
ContextAwareness & +9 & References relevant clinical context such as the identified source from the EHR document. \\
Completeness & $-6$ & References unnecessary or imprecise information in its reasoning or answer. \\
InstructionFollowing & +5 & Provides a reasoning section and final answer as requested. \\
\end{tabular}
\end{small}

\subsection{Reward functions used during GRPO}
\label{app:repro-rewards}

All rewards are computed on-policy per completion; the group mean and variance across the $G$ rollouts form the baseline.

\paragraph{Grounded rubric reward $r_{\text{rub}}$.}
Computed via the judge pipeline above:
\[
r_{\text{rub}}(\hat y \mid d) = \operatorname{clip}_{[0,1]}\Bigl(\tfrac{\sum_{i: j_i = \text{true}} p_i}{\sum_{i: p_i > 0} p_i}\Bigr).
\]
Degenerate rollouts (empty reasoning/answer or trivial repetition) are mapped to $r_{\text{rub}} = 0$.

\paragraph{Format rewards.}
Two mutually-exclusive variants depending on whether the policy emits an \texttt{<answer>} block.

\emph{With explicit answer wrapper.} $r_{\text{format}} = 1$ iff the completion matches the regex
\begin{verbatim}
^<think>.*?</think>.*?<answer>.*?</answer>$
\end{verbatim}
evaluated with \texttt{DOTALL}. Otherwise $r_{\text{format}} = 0$.

\emph{Without answer wrapper} (Qwen3-family). $r_{\text{format}} = 1$ iff the completion matches
\begin{verbatim}
^\s*<think>\s*.*?\s*</think>\s*(\S[\s\S]*)$
\end{verbatim}
and the trailing content does not itself start with another \texttt{<think>}.

\paragraph{Tag rewards.}
Partial-credit structural rewards on the \texttt{<think>} scaffolding.

\emph{With answer wrapper.} $r_{\text{tag}} = 0.25\cdot[n(\texttt{<think>})\!=\!1] + 0.25\cdot[n(\texttt{</think>})\!=\!1] + 0.25\cdot[n(\texttt{<answer>})\!=\!1] + 0.25\cdot[n(\texttt{</answer>})\!=\!1]$.

\emph{Without answer wrapper.} $r_{\text{tag}} = 0.4\cdot[n(\texttt{<think>})\!=\!1] + 0.4\cdot[n(\texttt{</think>})\!=\!1] + 0.2 \cdot [\text{non-empty content after \texttt{</think>} that does not start with \texttt{<think>}}]$.

\paragraph{Reasoning-steps reward (optional).}
$r_{\text{steps}} = \min(1,\, k/3)$ where $k$ counts regex matches of step markers (``Step N:'', numbered lists, bullet points, transition words) inside the \texttt{<think>} block. We found this useful for the 1.5B scale but not for the 8B scale.

\paragraph{Repetition penalty (optional).}
$r_{\text{rep}} = -\lambda \cdot \bigl(1 - |\text{unique-}n\text{-grams}|\,/\,|\text{total-}n\text{-grams}|\bigr)$ with $n=3$ and $\lambda=1.0$.

\paragraph{Canonical reward stack.}
For Qwen3-8B and Qwen2.5-7B we use $r_{\text{total}} = r_{\text{rub}} + r_{\text{format}} + r_{\text{tag}}$ (unit weights). For the 1.5B anchor we additionally enable $r_{\text{steps}}$ and $r_{\text{rep}}$ for early stabilisation.

\subsection{Training hyper-parameters}
\label{app:repro-hp}

\begin{small}
\begin{tabular}{@{}p{0.28\linewidth}p{0.16\linewidth}p{0.16\linewidth}p{0.16\linewidth}p{0.16\linewidth}@{}}
\textbf{Hyper-parameter} & \textbf{Qwen2.5-1.5B} & \textbf{Qwen2.5-7B} & \textbf{Qwen3-8B} & \textbf{MedGemma-4B} \\ \hline
\multicolumn{5}{@{}l}{\textit{SFT (ablation only; not in the production recipe)}} \\
Learning rate & $1\!\times\!10^{-5}$ & $1\!\times\!10^{-5}$ & $1\!\times\!10^{-5}$ & $1\!\times\!10^{-5}$ \\
Per-device batch & 8 & 2 & 2 & 2 \\
Grad.\ accumulation & 2 & 4 & 4 & 4 \\
Epochs & 3 & 1--3 & 1 & 3 \\[3pt]
\multicolumn{5}{@{}l}{\textit{GRPO (production recipe)}} \\
Init.\ checkpoint & base & base & base & base \\
Learning rate & $2\!\times\!10^{-5}$ & $1\!\times\!10^{-6}$ & $1\!\times\!10^{-6}$ & $1\!\times\!10^{-6}$ \\
$G$ (generations) & 8 & 4 & 8 & 4 \\
Per-device batch & 8 & 1 & 4 & 1 \\
Grad.\ accumulation & 4 & 2 & 4 & 2 \\
Max prompt length & 65{,}536 & 65{,}536 & 65{,}536 & 65{,}536 \\
Max completion len & 2{,}048 & 2{,}048 & 4{,}096 & 2{,}048 \\
Epochs & 5 & 1 & 1 & 1 \\
LR schedule & cosine & cosine & cosine & cosine \\
Warm-up ratio & 0.1 & 0.1 & 0.1 & 0.1 \\
Precision & bf16 & bf16 & bf16 & bf16 \\
Attention & FlashAttn-2 & FlashAttn-2 & FlashAttn-2 & FlashAttn-2 \\
Distributed & ZeRO-3 & ZeRO-3 & ZeRO-3 & ZeRO-3 \\
\end{tabular}
\end{small}

\paragraph{Production initialisation.}
Every production GRPO run is initialised directly from the base checkpoint listed at the top of its column; SFT is reported only as a stand-alone ablation in Appendix~\ref{app:training-ablation} and is not chained into the GRPO step.

\subsection{Merge configurations (weight-space)}
\label{app:repro-merge}

All merges are between the base model $\theta_{\mathrm{base}}$ and its GRPO-trained counterpart $\theta_{\mathrm{grpo}}$. For \frameworkName{}-8B the base is \texttt{Qwen/Qwen3-8B} and the GRPO checkpoint is the final policy of Section~\ref{app:repro-hp}. The filter is applied to the task vector $\Delta = \theta_{\mathrm{grpo}} - \theta_{\mathrm{base}}$ before adding it back at unit weight.

\paragraph{DELLA-Linear (final recipe used for deployment).}
Row-wise adaptive pruning of $\Delta$: within each row of each weight matrix, the largest-magnitude entries are retained at density $\rho$; the rest are zeroed. We use $\rho = 0.15$, \texttt{epsilon} $= 0.02$, and merge weight $1.0$; dtype \texttt{bfloat16}. This is the configuration that produced our headline clinical-reasoning and generalist-preservation numbers.

\paragraph{Breadcrumbs (ablation).}
Two-sided magnitude pruning of $\Delta$: the top $\gamma$ quantile (destructive outlier spikes) and the bottom $\beta$ quantile (RL noise) are pruned; the remaining middle band is retained at density $\rho$. We use $\rho = 0.15$, $\gamma = 0.02$, weight $1.0$.

\paragraph{DARE-Linear (ablation).}
Random dropout of $\Delta$ at rate $1 - \rho$ with rescaling by $1/\rho$, followed by linear addition. We use $\rho = 0.5$, weight $1.0$.

\paragraph{Activation-Informed Merging (negative-result ablation).}
A continual-learning post-filter applied after any of the above: a calibration batch is passed through the base model, per-weight activation magnitudes are recorded, and weights whose activation magnitude exceeds a threshold are locked to their base values before the filtered $\Delta$ is added. Across all three filters we found AIM neither improved clinical-reasoning scores nor measurably protected generalist behaviour once DELLA-Linear was in place; in a minority of settings it slightly regressed both. We therefore do not include AIM in the production recipe.

\subsection{Evaluation protocol (reproducibility view)}
\label{app:repro-eval}

\paragraph{\datasetName{} evaluation.}
For every (admission, action-space) item in the held-out split:
\begin{enumerate}[leftmargin=*,itemsep=0pt]
\item Construct the policy prompt (Section~\ref{app:prompts}).
\item Sample one generation from the model under test with temperature 0.2 and a maximum of 4{,}096 new tokens.
\item Parse the generation into \texttt{answer\_reasoning} and \texttt{final\_answer} according to the model's reasoning-trace family.
\item Call the Grounded Judge with the per-item rubric (Section~\ref{app:repro-rubric-gen}) and compute $r_{\text{rub}}$.
\end{enumerate}
Report: overall mean $r_{\text{rub}}$; per-axis aggregates (Accuracy, Completeness, Communication Quality, Context Awareness, Instruction Following); per-action-space aggregates (four categories); and a \emph{critical-accuracy} score computed over criteria whose axis is \textsc{Accuracy} and whose absolute weight $\geq 8$. Rubrics and verdicts are persisted so that re-aggregation against alternative scoring rules is possible post-hoc.

\paragraph{OpenAI HealthBench.}
The same checkpoints are evaluated on the public HealthBench \textsc{full}, \textsc{hard}, and \textsc{consensus} subsets using the official grader template and the same positive-point-normalised scoring rule. No fine-tuning on HealthBench is performed; this is a pure transfer test.

\paragraph{Clinician A/B preference.}
A blinded A/B study with licensed physicians on a held-out panel of inpatient cases compares pairwise responses from \frameworkName, GPT-5, and HuatuoGPT-o1. Reviewers see only the patient past and the two anonymised responses; judgements are collected on clinical usefulness, safety, and communication clarity.

\subsection{Determinism and seeding}
All data-preparation samplers (cohort selection, timeline splitting, train/test split) use seed 42. RL rollouts are non-deterministic by design; we report bootstrapped confidence intervals over the test set rather than single-seed point estimates. Every stage of the pipeline writes a self-contained artefact (traces, QA pairs, rubrics, rollouts, verdicts) so that re-running a downstream stage against a frozen upstream is straightforward.

\subsection{Training-stack ablation: SFT vs.\ GRPO and merge variants}
\label{app:training-ablation}

This appendix expands the two-sentence ablation reference in \S\ref{sec:results}. We report the full per-stage and per-merge-variant per-axis numbers on \datasetName{}, and then walk through the failure modes that distinguish each stage by inspecting representative outputs and the per-criterion verdict patterns from the Grounded Judge.

\paragraph{Per-stage table.}
Table~\ref{tab:training-stage-ladder} reports the per-axis scores for the four configurations we evaluate on each base family: \textit{Base}, \textit{SFT} on the oracle reference $(\rho^{\star}, y^{\star})$ as a stand-alone ablation, \textit{GRPO} on the rubric reward $r_{\text{rub}}$ initialised from the same base, and the two strongest weight-space merges of the GRPO task vector back into that base (DELLA-Linear and Breadcrumbs). SFT and GRPO are both initialised from \textit{Base} and are not chained; the merge rows are initialised from the corresponding GRPO row. All numbers are micro-aggregated over per-criterion points across the held-out 7{,}927-admission test split (Section~\ref{app:repro-eval}) under the same Grounded Judge and sampling parameters.

\begin{table}[h]
\centering\footnotesize
\setlength{\tabcolsep}{4pt}
\renewcommand{\arraystretch}{0.95}
\caption{Per-axis \datasetName{} scores (\%) for each training configuration on the Qwen3-8B and MedGemma-4B families. \textbf{Bold} marks the best score per family per axis. SFT and GRPO are independent ablations off \textit{Base}; the merge rows are initialised from the corresponding GRPO row.}
\label{tab:training-stage-ladder}
\begin{tabular}{@{}llrrrrrr@{}}
\toprule
\textbf{Family} & \textbf{Configuration} & \textbf{Acc.} & \textbf{Comp.} & \textbf{Ctx.} & \textbf{Comm.} & \textbf{Instr.} & \textbf{Aggr.} \\
\midrule
Qwen3-8B    & Base                 & 66.57 & 58.72 & 76.97 & \textbf{91.62} & 88.24 & 75.32 \\
Qwen3-8B    & SFT (ablation)       & 42.10 & 17.83 & 51.08 & 85.98 & 74.17 & 52.36 \\
Qwen3-8B    & GRPO                 & 69.21 & 67.92 & 80.29 & 90.15 & 86.68 & 77.99 \\
Qwen3-8B    & GRPO + Breadcrumbs   & 79.22 & 79.29 & 85.87 & \textbf{91.67} & \textbf{89.05} & 84.53 \\
Qwen3-8B    & GRPO + DELLA-Linear  & \textbf{80.77} & \textbf{80.98} & \textbf{86.19} & 90.50 & 88.14 & \textbf{84.95} \\
\midrule
MedGemma-4B & Base                 & 20.57 &  8.11 & 24.79 & 47.19 & 40.06 & 27.01 \\
MedGemma-4B & SFT (ablation)       & 44.78 & 16.74 & 55.19 & \textbf{84.24} & 71.73 & 52.89 \\
MedGemma-4B & GRPO                 & 56.63 & 43.17 & \textbf{68.47} & 81.20 & \textbf{72.16} & \textbf{63.40} \\
MedGemma-4B & GRPO + Breadcrumbs   & 55.87 & 45.58 & 61.14 & 72.45 & 67.29 & 59.79 \\
MedGemma-4B & GRPO + DELLA-Linear  & \textbf{64.11} & \textbf{54.49} & 67.29 & 64.98 & 66.56 & 63.35 \\
\bottomrule
\end{tabular}
\end{table}

\paragraph{Why SFT regresses on Qwen3-8B.}
The most striking row in Table~\ref{tab:training-stage-ladder} is the Qwen3-8B SFT ablation, which drops the aggregate by $22.95$~pp relative to the base. Qwen3-8B already carries a reasoning prior, and the SFT target $\mathtt{<think>}\,\rho^{\star}\,\mathtt{</think>}\;y^{\star}$ contracts the policy in two harmful ways simultaneously. First, $y^{\star}$ is a single concise oracle answer that mentions one downstream care action; cloning it freezes Completeness onto a one-action ceiling that the rubric's per-criterion enumeration explicitly rewards beyond ($-40.89$~pp Completeness vs.\ base). Second, $\rho^{\star}$ is written as a tidy oracle monologue rather than as the model's natural exploratory $\mathtt{<think>}$ trace, and cloning it suppresses the model's pre-installed reasoning style on prompts that fall outside the SFT distribution ($-25.89$~pp ContextAwareness, $-14.07$~pp InstructionFollowing). On reading the SFT outputs we observe two characteristic failure modes. \emph{Mode A: short verbatim mimicry.} The policy emits a one-paragraph $\mathtt{<think>}$ followed by a one-line answer, hits the matched-criterion in the rubric, and misses every other positive criterion the case carries. \emph{Mode B: format collapse on out-of-distribution prompts.} On prompts that drift from the SFT format the policy emits a $\mathtt{<think>}$ block but no terminal answer, and the rubric scores all positive criteria as false. The conversational axes (CommunicationQuality, InstructionFollowing) survive SFT, while the content axes collapse.

\paragraph{Why SFT lifts MedGemma-4B but still trails GRPO.}
Run as the same stand-alone ablation, SFT on MedGemma-4B reaches $52.89\%$, $25.88$~pp above its very low base ($27.01\%$); the lift is concentrated on the conversational axes ($+37.05$ Communication, $+31.67$ Instruction) and is more modest on the content axes ($+24.21$ Accuracy, $+8.63$ Completeness). The reading is that the SFT target is essentially a faithful demonstration of the $\mathtt{<think>}$/answer scaffolding plus a clinical answer, and that scaffolding is what MedGemma-4B was missing relative to a reasoning-tuned base. GRPO from the same base reaches $63.40\%$ ($+36.39$~pp over base, $+10.51$~pp over the SFT ablation), and the lift sits on the content axes ($+36.06$ Accuracy, $+35.06$ Completeness, $+43.68$ ContextAwareness vs.\ base) where the rubric reward provides a direct gradient that the static SFT target does not. The content gains come at a small concession on the conversational axes ($-3.04$ Communication, $-0.43$ Instruction vs.\ SFT), which the merge step then has to weigh against the content lift it provides.

\paragraph{Merge-variant comparison on Qwen3-8B.}
Table~\ref{tab:merge-variants-qwen} reports the four merge variants we evaluate on the Qwen3-8B GRPO task vector. All four variants score within $0.7$~pp of one another at the aggregate level, and the per-axis ordering is also stable: every variant sits within $\pm 1$~pp of the panel mean on Communication and Instruction, within $\pm 1$~pp of one another on Context, and within $\pm 2$~pp on the two content axes. This narrow band tells us that on a Qwen3-8B--scale base with our GRPO recipe the merge step delivers most of its lift through the structural act of restoring base-model weights to a high-density subset of the task-vector entries; the choice between density-controlled (DELLA-Linear), magnitude-banded (Breadcrumbs), and stochastic-thinning (DARE-Linear; not shown in the table because it underperformed by $>3$~pp aggregate in initial runs and we did not scale it) filters matters less than the presence of a filter at all. Activation-Informed Merging (AIM) sits within noise of its base filter on every axis ($-0.21$ aggregate against DELLA-Linear, $-0.24$ aggregate against Breadcrumbs); we therefore drop it from the production recipe for both the cost-of-calibration and the negligible lift.

\begin{table}[h]
\centering\footnotesize
\setlength{\tabcolsep}{4pt}
\renewcommand{\arraystretch}{0.95}
\caption{Merge-variant per-axis comparison on Qwen3-8B GRPO (\datasetName{}, \%). All four variants are within $0.7$~pp of one another at the aggregate; AIM is within noise of its base filter.}
\label{tab:merge-variants-qwen}
\begin{tabular}{@{}lrrrrrr@{}}
\toprule
\textbf{Variant} & \textbf{Acc.} & \textbf{Comp.} & \textbf{Ctx.} & \textbf{Comm.} & \textbf{Instr.} & \textbf{Aggr.} \\
\midrule
GRPO + DELLA-Linear        & \textbf{80.77} & \textbf{80.98} & \textbf{86.19} & 90.50 & 88.14 & \textbf{84.95} \\
GRPO + DELLA-Linear + AIM  & 80.40 & 80.76 & 85.60 & 90.48 & 88.41 & 84.74 \\
GRPO + Breadcrumbs         & 79.22 & 79.29 & 85.87 & 91.67 & 89.05 & 84.53 \\
GRPO + Breadcrumbs + AIM   & 78.71 & 78.99 & 85.34 & \textbf{91.91} & \textbf{89.08} & 84.29 \\
\bottomrule
\end{tabular}
\end{table}

\paragraph{Merge-variant comparison on MedGemma-4B and the axis-rebalancing pattern.}
Table~\ref{tab:merge-variants-mg} reports the same comparison for MedGemma-4B GRPO. Here the merge step does not raise the aggregate (DELLA-Linear $63.35$, Breadcrumbs $59.79$, vs.\ pre-merge GRPO $63.40$); it \emph{re-balances} the per-axis allocation. DELLA-Linear lifts the two content axes by $7.48$ and $11.32$~pp and concedes $16.22$ and $5.60$~pp on the conversational axes. Breadcrumbs is more conservative on the conversational side ($-8.75$ Communication) but also less aggressive on content. Reading paired outputs on the same admissions, the conversational drop is mostly the merged 4B re-acquiring the base MedGemma-4B's tendency to drop into mid-answer markdown and to truncate before a closing summary; the content lift is the merged 4B preserving the GRPO-installed differential-and-evidence structure. We retain DELLA-Linear as the production 4B recipe because the per-axis content lifts are the ones that matter for the deployment surface (\S\ref{sec:deploy}), and because the conversational drop is partially recovered by the deployed system prompt. The qualitative outputs that justify this trade are the four representative cases in Appendix~\ref{app:cases}.

\begin{table}[h]
\centering\footnotesize
\setlength{\tabcolsep}{4pt}
\renewcommand{\arraystretch}{0.95}
\caption{Merge-variant per-axis comparison on MedGemma-4B GRPO (\datasetName{}, \%). The merge step on the 4B family is a per-axis rebalancing rather than an aggregate lift: DELLA-Linear gains content-axis points and concedes conversational-axis points relative to pre-merge GRPO.}
\label{tab:merge-variants-mg}
\begin{tabular}{@{}lrrrrrr@{}}
\toprule
\textbf{Variant} & \textbf{Acc.} & \textbf{Comp.} & \textbf{Ctx.} & \textbf{Comm.} & \textbf{Instr.} & \textbf{Aggr.} \\
\midrule
GRPO (no merge)             & 56.63 & 43.17 & \textbf{68.47} & \textbf{81.20} & \textbf{72.16} & \textbf{63.40} \\
GRPO + DELLA-Linear         & \textbf{64.11} & \textbf{54.49} & 67.29 & 64.98 & 66.56 & 63.35 \\
GRPO + DELLA-Linear + AIM   & 63.05 & 53.59 & 66.77 & 65.96 & 66.25 & 62.94 \\
GRPO + Breadcrumbs          & 55.87 & 45.58 & 61.14 & 72.45 & 67.29 & 59.79 \\
GRPO + Breadcrumbs + AIM    & 55.61 & 44.61 & 61.02 & 73.27 & 67.80 & 59.74 \\
\bottomrule
\end{tabular}
\end{table}

\paragraph{Per-criterion verdict patterns.}
To complement the per-axis aggregates we also inspected the Grounded Judge's per-criterion verdict pattern on a fixed 200-item slice. Three patterns recurred. (i) Compared head-to-head against the SFT ablation on Qwen3-8B, GRPO converts on average $4.7$ false-to-true verdicts per case on positive criteria of the Completeness axis, dominated by ``mentions a differential beyond the most likely diagnosis'' and ``cites the relevant past lab/vital with a value''. (ii) The GRPO $\to$ DELLA-Linear merge step adds an additional $1.8$ false-to-true conversions per case on positive Accuracy criteria, almost exclusively those that require correct named entities (drug names, anatomic sites, ICD-style descriptors) where the merge restores base-model lexical breadth. (iii) The same merge step does not move negative-points criteria in either direction on the spine cohort, but it lifts the negative-trigger rate by $0.4$~pp on the obesity cohort, consistent with the cohort-conditional comprehensiveness-vs-overhead split documented in Appendix~\ref{app:clinician-qualitative-themes}.

\paragraph{What this ablation does and does not show.}
This ablation supports the design choice of GRPO with a weight-space merge as the production recipe, and it supports DELLA-Linear as the chosen merge filter for both families. It does not show that merging is uniformly beneficial: on the 4B family it is a per-axis re-allocation rather than a clean win, and the choice to retain it is a deployment-surface decision rather than a benchmark-score decision. We also do not vary the GRPO group size $G$ or the merge density $\rho$ in this table; those variations are smaller in magnitude than the configuration-level effects shown here and do not change the ordering.

%% file: appendix_clinician.tex

\providecommand{\tbr}[1]{\textcolor{gray}{\textbf{[#1]}}}

\section{Clinician Validation: Extended Protocol and Results}
\label{app:clinician}

This appendix provides the implementation-neutral specification of the user study summarised in \S\ref{sec:clinician}. All formulas, annotation schemas, and analysis steps are reproduced here so that the protocol is reproducible without access to the underlying application.

\subsection{Rater pool and cohort construction}
\label{app:clinician-cohort}

\paragraph{Raters.}
Four board-certified physicians participated, organised into two specialty cohorts of two raters each. The \emph{spine/orthopaedic} cohort comprised two orthopaedic surgeons, one senior (attending) and one junior (resident/fellow). The \emph{general-medicine/obesity} cohort comprised two family-medicine clinicians, also one senior and one junior. All four raters were clinically active at the time of annotation and had no prior exposure to the model outputs under evaluation. Raters were instructed to skip any case they felt under-qualified to annotate. The study was conducted as part of an on-site research collaboration with the partner network.

\paragraph{Case cohorts.}
From the \datasetName held-out split, two case cohorts were constructed by keyword filtering over the patient timeline text (diagnosis descriptions, procedure descriptions, and clinical note content). The \emph{spine} cohort was filtered with the keyword set $\{\text{spine}, \text{orthopaedic}, \text{orthopedic}\}$. The \emph{obesity} cohort was filtered with the keyword set $\{\text{obesity}, \text{obese}, \text{BMI}, \text{bariatric}, \text{weight loss}\}$. Each cohort was then capped to a comparable annotation budget per cohort. The same cohort cases are used for both Phase~1 and Phase~2, so that every case produces paired rubric-alignment and pairwise-preference data joinable at the case level.

\paragraph{Coverage across action spaces.}
Because a single admission in \datasetName \ contributes one question per action space (\S\ref{sec:method-data}), each cohort's cases span \textit{Diagnosis Assistance}, \textit{Treatment Recommendations}, \textit{Procedural Decision Making}, and \textit{Responding under Uncertainty}, ensuring that the per-axis and per-action-space breakdowns reported in Tables~\ref{tab:clinician-phase1-peraxis}--\ref{tab:clinician-phase1-percohort} are populated for both cohorts.

\subsection{Annotation interface}
\label{app:clinician-ui}

Both phases are delivered through a purpose-built web application. Annotations are persisted to a central store so that drafts can be resumed, and every DOM-level edit event (toggles, drag-reorders, inline edits, rationale keystrokes) is accumulated as an \texttt{interaction\_count} effort proxy per submission.

\paragraph{Shared elements.}
Every page presents the patient's past timeline as event cards in chronological order, an on-screen ``Clinical Question'' panel placed at the end of the past, a \emph{Future Outcomes} panel clearly demarcated as \emph{``not visible to models''}, and the oracle-authored reference response with an amber disclaimer stating it is AI-generated with full access to the future and is a \emph{reference only, not a candidate to grade}. Any case can be marked invalid at any time via a modal that requires a free-text justification; invalid submissions are stored but excluded from aggregate quality metrics.

\begin{figure}[h]
\centering
\includegraphics[width=\linewidth]{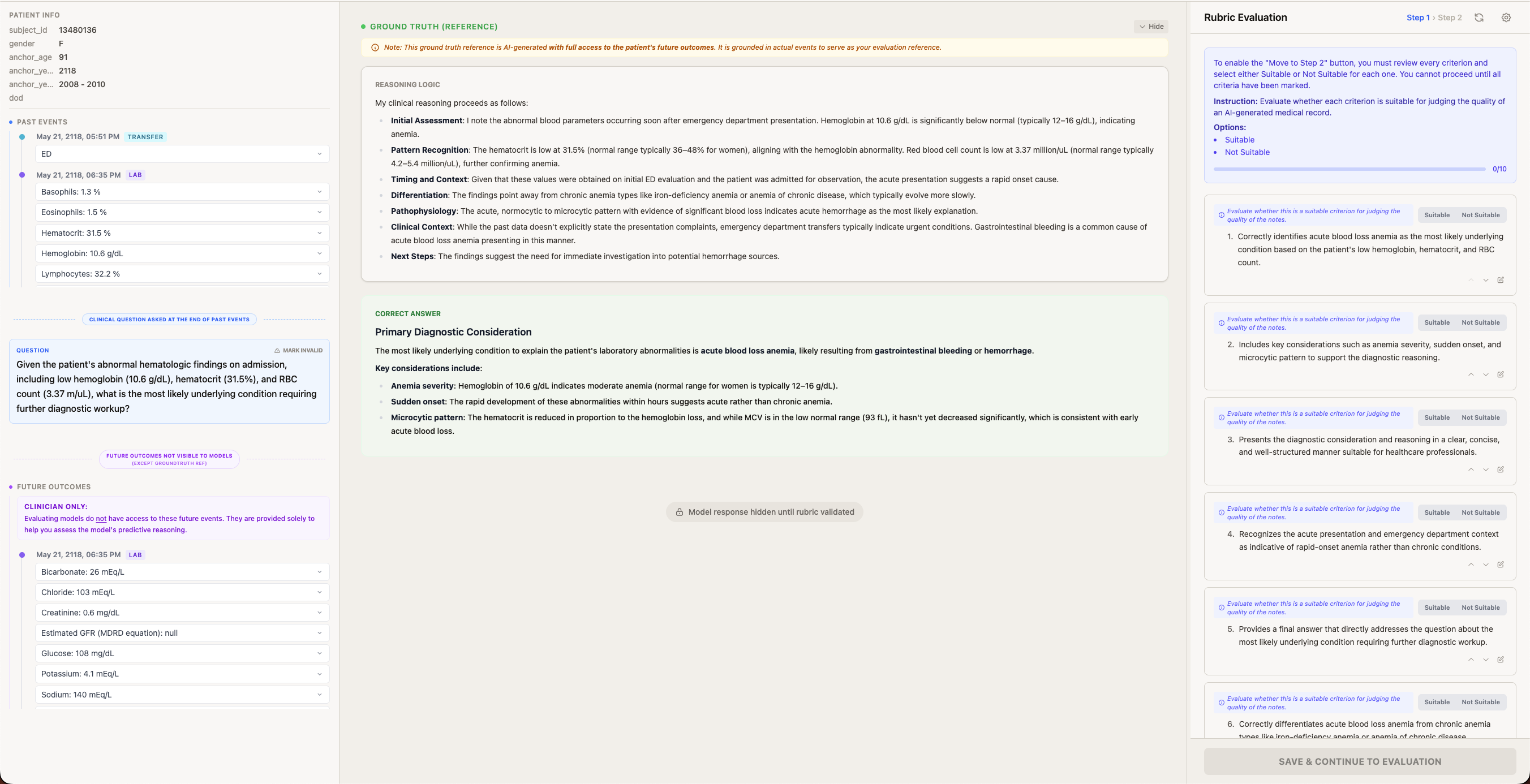}
\caption{Phase~1 Step~1 (rubric editing) annotation interface. Left: patient timeline and clinical question. Right: per-criterion cards with \textsc{Suitable}/\textsc{Not Suitable} toggles, in-place editing, drag-reorder, delete, and an ``add criterion'' affordance at the bottom. The model response is hidden in this step so that rubric edits cannot be motivated by the candidate output.}
\label{fig:ui-rubrics-step1}
\end{figure}

\begin{figure}[h]
\centering
\includegraphics[width=\linewidth]{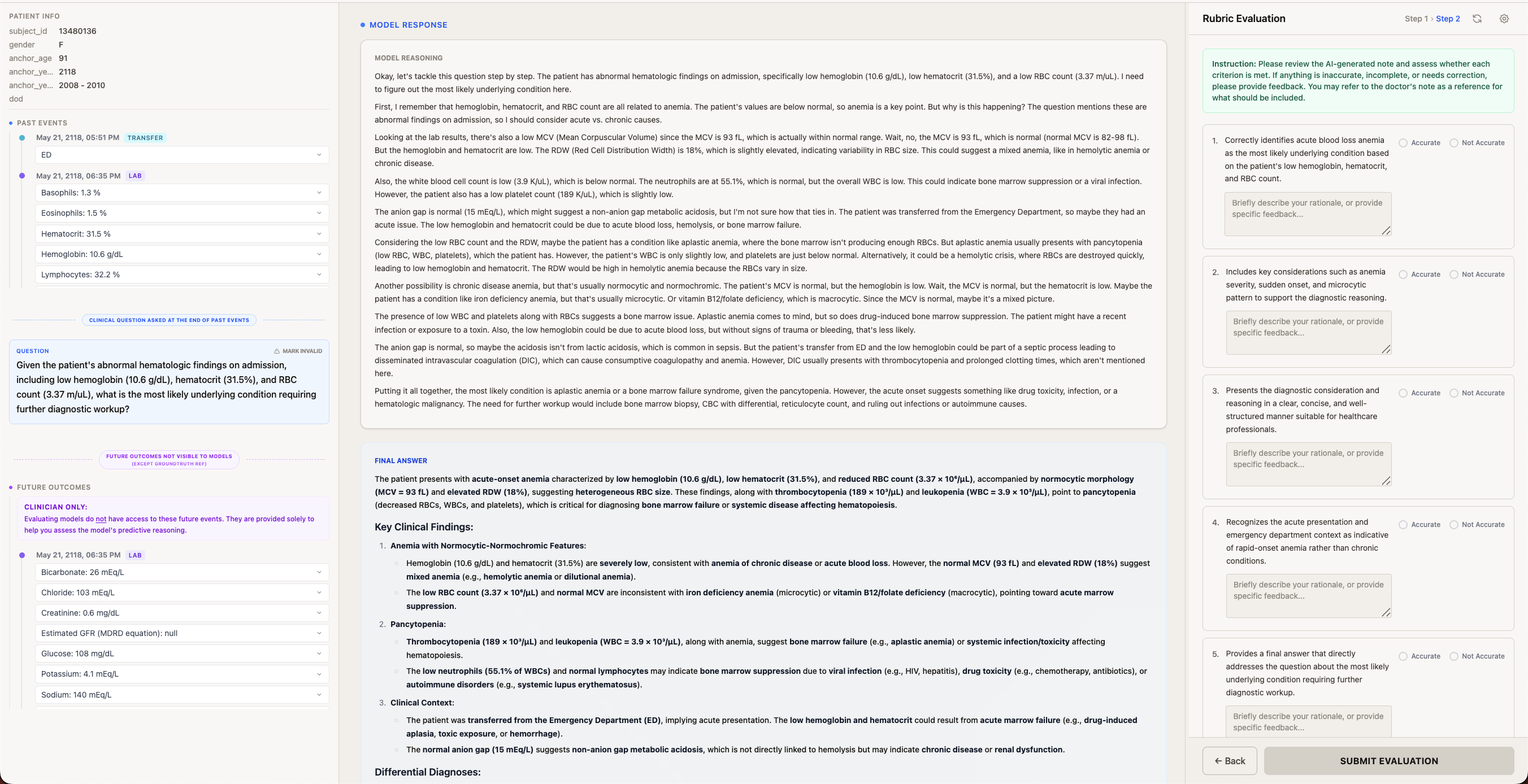}
\caption{Phase~1 Step~2 (rubric grading) annotation interface. The candidate model response is now revealed and the rubric is locked; each criterion gets an \textsc{Accurate}/\textsc{Not Accurate} radio plus a short free-text rationale. A back-arrow returns to Step~1 if the rater wants to refine the rubric in light of the response.}
\label{fig:ui-rubrics-step2}
\end{figure}

\paragraph{Phase~1 interface (rubrics).}
A three-pane layout: (i) patient context and timeline, (ii) reference response plus model response (the latter hidden during Step~1), (iii) a rubric console with per-criterion cards. Step~1 (Figure~\ref{fig:ui-rubrics-step1}) presents each criterion as an editable card with inline actions: toggle \textsc{Suitable} / \textsc{Not Suitable}, edit description, move up/down, delete (only for physician-added criteria), reset to oracle wording (only for modified criteria), and a bottom button to add a new criterion. A submit guard prevents advancing to Step~2 unless every criterion has an explicit \textsc{Suitable}/\textsc{Not Suitable} verdict. Step~2 (Figure~\ref{fig:ui-rubrics-step2}) reveals the candidate model response, locks the rubric, and presents each criterion with \textsc{Accurate}/\textsc{Not Accurate} radios plus a short free-text rationale; raters may traverse back to Step~1 to refine the rubric before submitting. Exactly one candidate response is graded per case in Phase~1; model identity is not shown.

\begin{figure}[h]
\centering
\includegraphics[width=\linewidth]{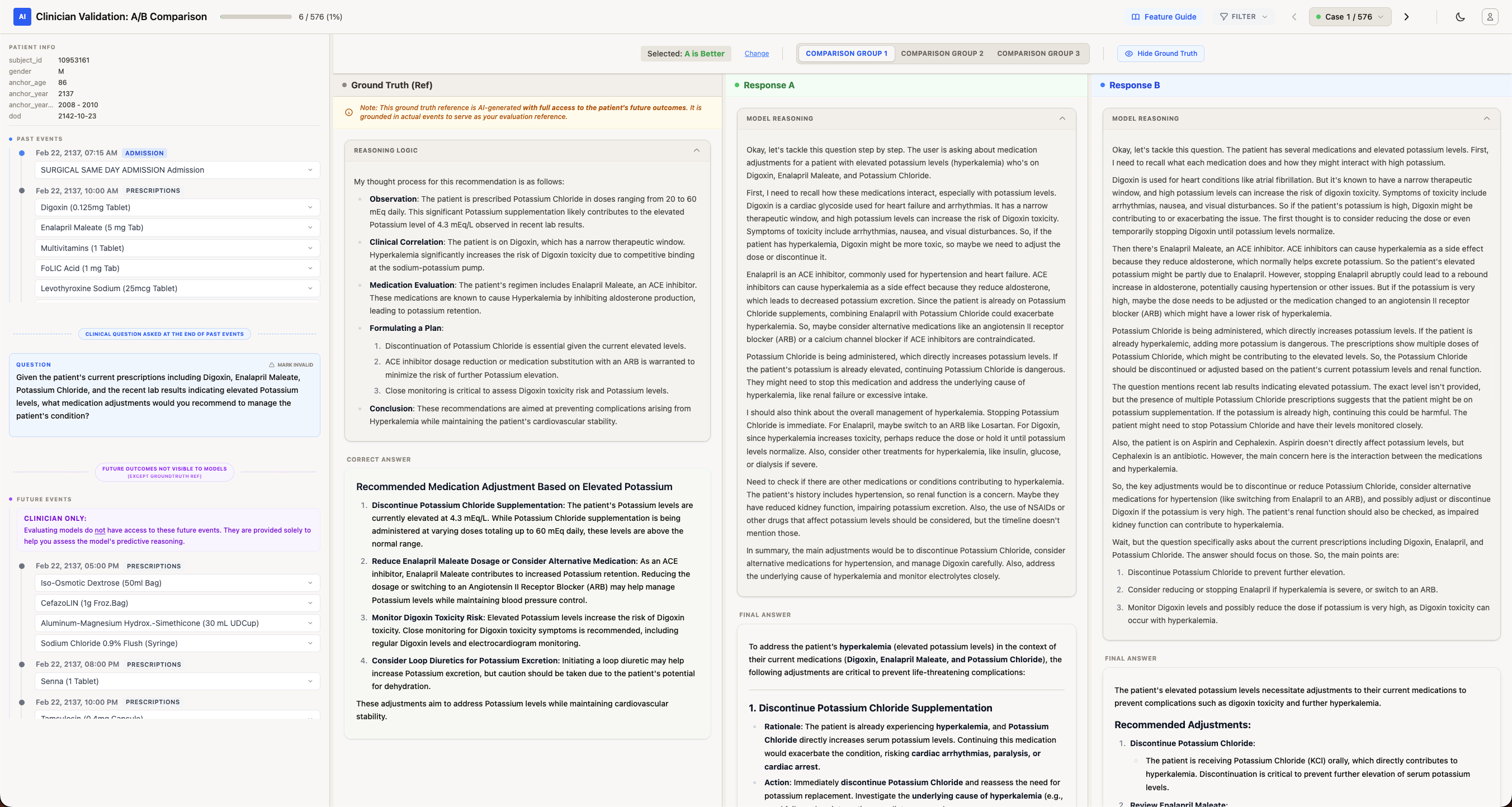}
\caption{Phase~2 (blinded A/B preference) annotation interface. Top: patient context and clinical question. Below: side-by-side anonymised candidate responses with the reference response visible above (clearly labelled as oracle output, not a candidate). The pair-selector pill row at the top of the response panel rotates between the three model pairs; left/right placement is randomised per (case, pair) and the mapping is persisted for the position-bias diagnostic.}
\label{fig:ui-ab}
\end{figure}

\paragraph{Phase~2 interface (A/B).}
A four-pane layout (Figure~\ref{fig:ui-ab}): (i) patient context and timeline, (ii) reference response, (iii) response A, (iv) response B. A sticky decision bar offers \{\emph{A is Better}, \emph{Tie}, \emph{B is Better}\} (keyboard shortcuts \texttt{1} and \texttt{2}). For each case, three model pairs are shown sequentially via a pill-style pair selector; the rater must complete all three before advancing. For each pair the client randomises left/right placement with probability $1/2$ and persists the mapping \{\texttt{actualModelA}, \texttt{actualModelB}, \texttt{displayedAsA}, \texttt{displayedAsB}\}; a post-hoc position-bias check uses this mapping. Model names are removed from the UI at both the component and the serialisation layer to prevent leakage.

\paragraph{Audit fields persisted per submission.}
For every case and every submission type the store captures the rater identifier, experiment identifier, patient identifier, sample identifier, submission type (\texttt{clinical\_reasoning} for rubric judgement or \texttt{ab\_clinical\_reasoning} for A/B), the type-specific payload (criterion verdicts or pairwise choice), \texttt{is\_invalid} and \texttt{invalid\_reason}, \texttt{results\_metadata.interaction\_count}, \texttt{is\_draft}, and server-side timestamps for each attempt. Drafts are upserted; a final submission promotes the draft in place.

\subsection{Verbatim rater instructions}
\label{app:clinician-instructions}

The following instructions are reproduced verbatim from the physician guide, lightly elided only to remove UI-navigation details that are obvious from Appendix~\ref{app:clinician-ui}.

\paragraph{Study framing.}
\begin{quote}\small\itshape
There are 2 key things we want clinician validation for our research and development:
(1) The Judgement Criteria --- This will be the first phase of the clinician validation process. Since our entire reward generation and judgement process is automated, we want to collect data on how aligned the judgement criteria is compared to actual clinicians evaluating the model response.
(2) The Model Performance --- Next, we want to supplement our automated testing process with a blind A/B test where we present our model response compared to other frontier model responses and let clinicians select the ``better'' response.
\end{quote}

\paragraph{Task setting.}
\begin{quote}\small\itshape
We have taken completed patient journeys and split them into a past and a future. The goal is to answer a complex clinical reasoning query at the point when the past timeline ends. Practically, this aims to emulate an in-patient setting, with the future timeline serving as a ground truth of actions taken to ensure a successful discharge.
\end{quote}

\paragraph{Phase~1 rater prompt.}
\begin{quote}\small\itshape
Think of this step as designing an exam for the model. You are the examiner and your goal is to do the following:
(1) Set the correct question to ask the student (model) --- evaluate whether the question is valid to ask given the past timeline. This means whether answering the question provides crucial insights into further actions, follow-ups or interventions for the patient.
(2) Set the correct evaluation rubrics --- if the question is valid, how would you evaluate a response for the question? Do the different criteria for evaluation align with how you would judge the response of a student responding to the question?
(3) Grade the response --- given a response by a model, use the rubrics you defined to grade the model.
\end{quote}

\paragraph{Phase~1 Step~1 operating rules (rubric edits).}
\begin{quote}\small\itshape
Each card represents one specific fact or clinical reasoning step that must be present in the answer. You should mark it as suitable if the criterion is suitable for judging the quality of the output, else mark it as unsuitable. Note that this is not just the accuracy of the response, but also the presentation, formatting, context awareness and any other expectation you may have when reviewing a response from the model.

Add missing Criteria in the evaluation rubrics.
When to use: You need to evaluate a factor that isn't in the current rubric but important for judging a response to the question.
Guidelines:
(a) Atomic: Add separate criteria for different concepts (e.g., reasoning steps, formatting).
(b) Yes/No: Criteria must be answerable by ``Accurate'' or ``Not Accurate'' based on the model's output.
(c) Prioritize: Drag and reorder your added criteria to rank them by importance.
\end{quote}

\paragraph{Phase~1 Step~2 grading rule (including the negative-criterion rule).}
\begin{quote}\small\itshape
In Step 2, review the AI's output:
Reasoning: Internal thought process. Should be clinically sound.
Final Answer: User-facing response. Should be safe and well-formatted.
Evaluate both aspects.
Do not grade the reference answer in green, only the model response in blue!
For each criterion, mark if the model was Accurate or Not Accurate. Use the Reasoning Box to explain why the model failed (e.g., ``Hallucinated lab value'').
\textbf{Note: For negative criteria (e.g., ``Model does NOT mention X''), mark Accurate if the statement is true (i.e., X is missing).}
\end{quote}

\paragraph{Phase~2 rater prompt.}
\begin{quote}\small\itshape
Review the Model A and Model B response.
Reasoning: Internal clinical thought process.
Answer: Final user-facing response.
Ensure the reasoning is sound and the answer is well-formatted. Consider Accuracy, Factuality, Faithfulness, Informativeness, Context Awareness, Formatting and Presentation of findings when selecting the best response. Also consider which reasoning is more sound, which final answer is generally better. Look for nuances like missing lab values, considerations for patient conditions or dangerous interactions.
Choose the better response. If both are equally good (or equally bad), select ``It's a Tie''.
Toggle between the 3 comparison groups to compare 3 different model pairs. Ensure that you compare all 3 before moving on to the next case.
\end{quote}

\subsection{Analysis-engine specification}
\label{app:clinician-metrics}

Physician annotations are ingested by an analysis engine that produces a single machine-generated report per run. The engine consumes (i) the per-case rubric with its original oracle wording and point weights, (ii) the per-rater Phase~1 submissions, (iii) the per-rater Phase~2 submissions, and (iv) the oracle's own stored verdict vector and scalar score for each case. It emits a markdown report and a machine-readable JSON artefact. All metrics below are computed on pooled annotations unless stated otherwise; senior and junior results are additionally reported separately.

\paragraph{Rubric-level quality metrics (Phase~1 Step~1).}
Let $\mathcal{K}$ be the full set of criteria after the rater has finished Step~1 on a case; let $\mathcal{K}^{\text{orig}}\subseteq\mathcal{K}$ be the oracle-authored subset, and let $\mathcal{K}^{\text{new}}=\mathcal{K}\setminus\mathcal{K}^{\text{orig}}$ be the rater-added criteria. Each criterion carries flags \texttt{is\_new}, \texttt{is\_modified}, \texttt{not\_relevant} and metadata \{\texttt{axis}, \texttt{description}, \texttt{points}, \texttt{order}\}.
\begin{align*}
\text{Query validity rate} &\;=\; 1 - \tfrac{|\{\text{cases marked invalid}\}|}{|\{\text{total cases}\}|} \\
\text{Criterion relevance rate} &\;=\; 1 - \tfrac{|\{k \in \mathcal{K}^{\text{orig}}: \texttt{not\_relevant}\}|}{|\mathcal{K}^{\text{orig}}|} \\
\text{Modification rate} &\;=\; \tfrac{|\{k \in \mathcal{K}^{\text{orig}}: \texttt{is\_modified}\}|}{|\mathcal{K}^{\text{orig}}|} \\
\text{Addition rate} &\;=\; \tfrac{|\mathcal{K}^{\text{new}}|}{|\mathcal{K}^{\text{orig}}|} \\
\text{Axis coverage}_a &\;=\; \mathbf{1}\bigl[\exists k \in \mathcal{K}: \texttt{axis}(k)=a\bigr], \quad a \in \text{axes}.
\end{align*}
Per-axis versions of relevance, modification, and addition are reported alongside the pooled totals (Appendix~\ref{app:clinician-phase1-extended}).

\paragraph{Inter-rater reliability (Phase~1 Step~2).}
Let $v^{s}_k \in \{0,1\}$ and $v^{j}_k \in \{0,1\}$ denote the binary \textsc{Accurate}/\textsc{Not Accurate} verdicts of the senior and junior surgeon on criterion $k$, restricted to criteria retained as \textsc{Suitable}. Over all (case, criterion) pairs we report
\begin{itemize}[leftmargin=*,itemsep=1pt]
\item \textbf{Percent agreement}: $\mathrm{PA} = \tfrac{1}{N}\sum_k \mathbf{1}[v^{s}_k = v^{j}_k]$.
\item \textbf{Cohen's $\kappa$}: $\kappa = (P_o - P_e)/(1 - P_e)$ with $P_o=\mathrm{PA}$ and $P_e$ the product of marginal rates for each value.
\item \textbf{Fleiss' $\kappa$}: multi-rater generalisation of the above with equal-sized judge panels per item.
\item \textbf{Krippendorff's $\alpha$}: the nominal-metric variant of Krippendorff's $\alpha$ over the same binary verdicts, robust to missing raters per item.
\end{itemize}

\paragraph{Judge--clinician alignment (Phase~1 Step~2, oracle vs.\ clinician).}
Let $v^{o}_k$ be the oracle's binary verdict on the clinician-aligned criterion $k$ and $v^{c}_k$ be the pooled clinician verdict (majority; ties broken conservatively as \textsc{Not Accurate}). Define the confusion matrix
\[
\text{TP}=\!\!\!\sum_{k: v^{c}_k=1, v^{o}_k=1}\!\!\!1, \quad
\text{FP}=\!\!\!\sum_{k: v^{c}_k=0, v^{o}_k=1}\!\!\!1, \quad
\text{FN}=\!\!\!\sum_{k: v^{c}_k=1, v^{o}_k=0}\!\!\!1, \quad
\text{TN}=\!\!\!\sum_{k: v^{c}_k=0, v^{o}_k=0}\!\!\!1.
\]
From this we compute accuracy, precision, recall, $F_1$, and balanced $F_1$ (mean of pass-$F_1$ and fail-$F_1$); and Cohen's $\kappa$ between $v^{o}$ and $v^{c}$. We additionally report the per-axis $F_1$ for each of the five axes.

\paragraph{Score-level alignment (Phase~1 Step~2).}
Per case $d$ we compute the \emph{normalised scalar score} both from the oracle and from the clinician using the scoring rule of \S\ref{sec:method-judge}:
\[
r_{\text{rub}}^{(\cdot)}(d) = \operatorname{clip}_{[0,1]}\!\Bigl(\tfrac{\sum_{k:\,v^{(\cdot)}_k=1}p_k}{\sum_{k:\,p_k>0}p_k}\Bigr),
\]
and then
\[
\text{MAE} = \tfrac{1}{N}\sum_{d}\bigl|r_{\text{rub}}^{o}(d) - r_{\text{rub}}^{c}(d)\bigr|,
\qquad
r_{\text{Pearson}}, \rho_{\text{Spearman}} \text{ over } \{r_{\text{rub}}^{o}(d)\}_{d}, \{r_{\text{rub}}^{c}(d)\}_{d}.
\]

\paragraph{A/B preference metrics (Phase~2).}
For each pair $(m_1, m_2)$ let $w_{m_1}, w_{m_2}, t$ be the pooled counts of ``$m_1$ better'', ``$m_2$ better'', and ``tie'' across raters and cases. We report:
\begin{align*}
\text{Win rate}(m_1 \text{ vs.\ } m_2) &= \tfrac{w_{m_1}}{w_{m_1}+w_{m_2}+t},
\quad \text{Tie rate} = \tfrac{t}{w_{m_1}+w_{m_2}+t},\\
\text{Position-bias rate} &= \tfrac{|\{\text{cases where left pane preferred}\}|}{|\{\text{non-tie cases}\}|},\\
\text{Senior-vs-junior concordance} &= \tfrac{|\{\text{cases where the two raters agree}\}|}{|\{\text{cases both rated}\}|}.
\end{align*}

\paragraph{Uncertainty quantification.}
All pooled rates are reported with a two-sided bootstrap 95\% confidence interval computed with 1{,}000 resamples of cases (with replacement). For pairwise preference and judge--clinician accuracy, we additionally report a one-sided binomial test of the null ``arm $m_1$ is not preferred over arm $m_2$'' and ``oracle and clinician are independent'' respectively, with Bonferroni correction across the three A/B arms.

\paragraph{Analysis report contents.}
The auto-generated report contains, at minimum: (i) per-case diagnostic rows with raw annotations, (ii) pooled Phase~1 tables, (iii) the full confusion matrix $(v^{o}, v^{c})$ with per-axis breakdowns, (iv) the pooled A/B win/tie/loss matrix with CIs, (v) senior-vs-junior splits for every metric, (vi) disagreement case studies with model outputs, clinician rationales, and the specific criteria that disagreed. The JSON artefact contains the same data in a machine-consumable form.

\subsection{Three reuses of the clinician-annotated set}
\label{app:clinician-three-uses}

The Phase-1 + Phase-2 annotation pass is structurally one artefact (per-criterion verdicts $\times$ pairwise preferences across two cohorts of two clinicians each), reused for three structurally distinct downstream purposes that we deliberately keep separate.

\paragraph{Use~1: ICL task adaptation of the rubric-author and per-criterion judge LLMs.}
Clinician-edited rubrics from Phase~1 Step~1 (criteria added, deleted, reworded, reweighted) are inserted as in-context exemplars into the rubric-authoring oracle prompt; the clinician verdict vectors $v^{c}$ from Phase~1 Step~2 are inserted as in-context exemplars into the grading-judge prompt. We do \emph{not} fine-tune the judge weights (no SFT, no DPO): with on the order of $\sim$100 cases per cohort, the gradient step is too noisy to risk freezing one cohort's stylistic preferences into the judge weights, while the in-context exemplars remain re-applicable across model generations and across cohorts. The headline ICL alignment numbers are reported in Table~\ref{tab:clinician-phase1-judges} of the main paper; the full per-cohort breakdown including the negative ICL result on both cohorts is in Table~\ref{tab:clinician-phase1-judges-percohort}.

\paragraph{Use~2: clinical LLM-as-a-judge and clinical preference-model alignment analysis.}
The Phase-1 per-criterion verdicts (paired with the case, the rubric, and the candidate response) provide a held-out alignment surface against which off-the-shelf clinical LLM-as-a-judge configurations are scored on accuracy, $F_1$, Cohen's $\kappa$ against $v^{c}$, and on score-MAE of the per-case scalar score (Table~\ref{tab:clinician-phase1-judges}). Symmetrically, the Phase-2 pairwise preferences provide a held-out alignment surface for clinical preference-model configurations, scored on 3-way and A/B accuracy and rank-correlation against the unanimous-consensus clinician preference (Table~\ref{tab:clinical-pref-bench}). The resulting comparisons complement the static-jury approach of MedHELM~\citep{bedi2025medhelm} by characterising how off-the-shelf judge configurations align with the clinician signal across cohorts.

\paragraph{Use~3: blinded A/B comparison of \frameworkName{} against state-of-the-art baselines.}
Phase~2 doubles as the head-to-head reported in Table~\ref{tab:clinician-phase2}: \frameworkName{}-8B vs.~GPT-5 (proprietary frontier), \frameworkName{}-8B vs.~HuatuoGPT-o1-7B (open-source medical-reasoning), and \frameworkName{}-8B vs.~Qwen3-8B base (within-family anchor). This use is the deployment-relevance signal and is the comparison reviewers and clinicians both want answered: when asked to choose, do clinicians prefer our model's response over the strongest proprietary and the strongest open-source-medical alternatives?

\subsection{Extended Phase~1 results}
\label{app:clinician-phase1-extended}

\begin{table}[h]
\centering
\small
\caption{Per-axis Phase~1 rubric quality + judge--clinician alignment, aggregated across both cohorts. \emph{Relevance} is $1 - \frac{|\{\text{not\_relevant}\}|}{|\text{AI-gen criteria}|}$. \emph{Modif.} and \emph{Added} are normalised by the same denominator. The right-hand columns report the headline judge (Qwen3-32B, zero-shot) per-axis alignment vs.\ the pooled clinician verdict on the (criterion, candidate-response) pairs of Step~2.}
\label{tab:clinician-phase1-peraxis}
\begin{tabular}{@{}lcccccc@{}}
\toprule
\textbf{Axis} & \textbf{n criteria} & \textbf{Relevance} & \textbf{Modif.} & \textbf{Added} & \textbf{Judge $F_1$ (bal.)} & \textbf{Judge $\kappa$} \\ \midrule
Accuracy              & 561 & 84.3\% & 2.9\% & 5.3\% & 0.764 & 0.529 \\
Completeness          & 493 & 88.8\% & 2.4\% & 0.0\% & 0.729 & 0.459 \\
CommunicationQuality  & 523 & 93.3\% & 0.0\% & 0.0\% & 0.619 & 0.241 \\
ContextAwareness      & 492 & 88.0\% & 2.0\% & 0.0\% & 0.686 & 0.373 \\
InstructionFollowing  & 427 & 91.6\% & 0.0\% & 0.0\% & 0.694 & 0.388 \\
\bottomrule
\end{tabular}
\end{table}

The per-axis pattern is informative. Clinicians edited only the two clinically actionable axes (Accuracy gets 2.9\% modification + 5.3\% addition; Completeness 2.4\% modification) and did not touch the conversational axes (CommunicationQuality, InstructionFollowing — both 0.0\% on edits, only marking individual criteria as not relevant). The judge--clinician alignment shows the symmetric pattern: highest agreement on Accuracy ($\kappa{=}0.53$, $F_1{=}0.76$) where the clinician verdict has the strongest factual anchor, lowest on CommunicationQuality ($\kappa{=}0.24$) where the clinician verdict is more stylistic and varies between cohorts.

\begin{table}[h]
\centering
\small
\caption{Phase~1 results stratified by cohort (the within-cohort rater pair is identified in Appendix~\ref{app:clinician-cohort}); ``Aggregate'' pools both cohorts. The judge--clinician block reports the headline judge (Qwen3-32B, zero-shot) on each cohort's evaluable rater--criterion pairs.}
\label{tab:clinician-phase1-percohort}
\begin{tabular}{@{}lccc@{}}
\toprule
\textbf{Metric} & \textbf{Spine} & \textbf{Obesity} & \textbf{Aggregate} \\ \midrule
n rater-annotated cases                       & 103          & 127          & 230          \\
n unique patients                             & 37           & 69           & 101          \\
n rater--criterion pairs                      & 1{,}136      & 1{,}360      & 2{,}496      \\
\multicolumn{4}{@{}l}{\emph{Rubric quality}} \\
Criterion relevance rate                      & 83.1\%       & 94.0\%       & 89.1\%       \\
Modification rate                             & 3.4\%        & 0.0\%        & 1.5\%        \\
Addition rate                                 & 2.7\%        & 0.0\%        & 1.2\%        \\
\multicolumn{4}{@{}l}{\emph{Inter-rater reliability}} \\
Cohen's $\kappa$ (engine)                     & 0.749        & 0.658        & 0.702        \\
\multicolumn{4}{@{}l}{\emph{Judge--clinician alignment (Qwen3-32B, zero-shot)}} \\
$n_{\text{eval}}$                             & 1{,}138      & 1{,}359      & 2{,}497      \\
Accuracy                                      & 77.6\%       & 75.9\%       & 76.7\%       \\
$F_1$ (balanced)                              & 0.719        & 0.701        & 0.709        \\
Cohen's $\kappa$                              & 0.439        & 0.402        & 0.419        \\
Score MAE                                     & 0.124        & 0.192        & 0.160        \\
Score Pearson $r$                             & 0.571        & 0.434        & 0.478        \\
\bottomrule
\end{tabular}
\end{table}

\begin{table}[h]
\centering\small
\caption{Per-cohort clinical LLM-as-a-judge benchmark on Phase~1 verdicts. Same models / configurations as Table~\ref{tab:clinician-phase1-judges} of the main paper; the aggregate column there is the row-by-row pool of these two cohorts. The 5-shot ICL variant is only applicable to the two open-weights models. Best-per-column highlighted in \textbf{bold}.}
\label{tab:clinician-phase1-judges-percohort}
\begin{tabular}{@{}llcccc|cccc@{}}
\toprule
 & & \multicolumn{4}{c}{\textbf{Spine cohort}} & \multicolumn{4}{c}{\textbf{Obesity cohort}} \\
\cmidrule(lr){3-6} \cmidrule(lr){7-10}
\textbf{Judge} & \textbf{Mode} & $n$ & Acc & $F_1$ & $\kappa$ & $n$ & Acc & $F_1$ & $\kappa$ \\ \midrule
Qwen3-32B            & zero-shot   & 1{,}138 & 77.6\%          & 0.719          & 0.439           & 1{,}359 & 75.9\%          & 0.701          & 0.402 \\
Qwen3-32B            & 5-shot ICL  & 1{,}132 & 77.7\%          & 0.723          & 0.447           & 1{,}353 & 75.8\%          & 0.700          & 0.400 \\
MedGemma-27B-text-it & zero-shot   & 1{,}138 & 77.7\%          & 0.715          & 0.431           & 1{,}360 & 77.1\%          & 0.707          & 0.414 \\
MedGemma-27B-text-it & 5-shot ICL  & 1{,}133 & \textbf{78.8\%} & 0.713          & 0.431           & 1{,}355 & 76.8\%          & 0.685          & 0.372 \\
GPT-5                & zero-shot   & 1{,}138 & 71.5\%          & 0.676          & 0.357           & 1{,}360 & 76.1\%          & \textbf{0.731} & \textbf{0.472} \\
GPT-4.1              & zero-shot   & 1{,}135 & 76.0\%          & 0.694          & 0.389           & 1{,}360 & \textbf{77.6\%} & 0.718          & 0.436 \\
\bottomrule
\end{tabular}
\end{table}

The same cohort-asymmetry that drives the Phase-2 preference-model leaderboard inversion (Table~\ref{tab:clinical-pref-bench}) shows up here: \emph{which} judge is best depends on the cohort. MedGemma-27B-text-it edges Qwen3-32B by 1.0 pp on the spine cohort with 5-shot ICL ($78.8\%$ vs.\ $77.7\%$), while GPT-5 jumps from worst-on-spine ($71.5\%$, $\kappa{=}0.36$) to best-on-obesity ($\kappa{=}0.47$). The pooled column in the main-paper table averages these effects out, but the per-cohort leaderboard exposes that any single judge configuration carries cohort-specific blind spots; this asymmetry is one reason the deployed judge composes Qwen3-32B verdicts with the clinician-edited rubric rather than relying on a single monolithic judge.

\paragraph{ICL is not a reliable lever on this Phase-1 surface either.}
On the open-weights models, 5-shot ICL produces $-0.2$ to $+1.1$ pp on accuracy and small mixed effects on $\kappa$. The same pattern surfaces in the clinical preference-model alignment analysis (Appendix~\ref{app:clinician-three-uses}, Use~2): with on the order of $\sim$100 clinician-consensus dyads per cohort, the few-shot demonstration set is too small to reliably steer a 27B/32B judge's verdict on a clinical-reasoning surface where the underlying clinical-judgement variance between cohorts is itself larger than the ICL signal. We therefore report only the zero-shot configurations as the headline (Table~\ref{tab:clinician-phase1-judges}).

\subsection{Extended Phase~2 results}
\label{app:clinician-phase2-extended}

This subsection reproduces the per-pair, per-rater full report from the analysis engine for both cohorts. Raters within each cohort are anonymised as $R_1, R_2$ to protect identity; the labelling is alphabetical on the original anonymised username string, fixed across all subsections so that any cross-table comparison is consistent.

\paragraph{Per-rater preference breakdown (consensus subset).}
For each cohort and each rater we report the rater's decisive preference rate for \frameworkName{}-8B against the column-arm baseline, computed on the consensus subset (dyads where both raters agreed; ties excluded from the denominator). The pooled column is the cohort win rate already reported in Table~\ref{tab:clinician-phase2}.

\begin{table}[h]
\centering
\small
\caption{Phase~2 preference rate of \frameworkName{}-8B over each column-arm baseline, stratified by rater within each cohort. Decisive verdicts only; consensus subset.}
\label{tab:clinician-phase2-perrater}
\begin{tabular}{@{}llccc@{}}
\toprule
\textbf{Cohort} & \textbf{Comparator} & $R_1$ \textbf{rate} ($n$) & $R_2$ \textbf{rate} ($n$) & \textbf{Pooled rate} \\ \midrule
Spine    & GPT-5            & 100.0\% (24) & 100.0\% (24) & 100.0\% \\
Spine    & HuatuoGPT-o1-7B  & 100.0\% (30) & 100.0\% (30) & 100.0\% \\
Spine    & Qwen3-8B (base)  & 96.0\%  (25) & 96.0\%  (25) & 96.0\%  \\ \midrule
Obesity  & GPT-5            & 37.0\%  (27) & 37.0\%  (27) & 37.0\%  \\
Obesity  & HuatuoGPT-o1-7B  & 98.4\%  (61) & 98.4\%  (61) & 98.4\%  \\
Obesity  & Qwen3-8B (base)  & 72.9\%  (48) & 72.9\%  (48) & 72.9\%  \\
\bottomrule
\end{tabular}
\end{table}

By construction of the unanimous-consensus rule, the per-rater decisive rate within each (cohort, comparator) row equals the pooled rate (both raters picked the same arm on every dyad in the subset). The numbers therefore differ from the all-decisive aggregation reported in Table~\ref{tab:clinician-phase2} of the main paper, which is computed on the strictly larger superset of every decisive verdict (including the non-consensus dyads where the two raters disagreed). The divergence between raters that drives the cohort-level $\kappa$ value in Table~\ref{tab:clinician-phase2-irr} is concentrated in those excluded buckets, summarised in Table~\ref{tab:clinician-phase2-bucket} below.

\paragraph{Per-pair consensus bucket distribution.}
The unanimous-consensus rule used in Table~\ref{tab:clinician-phase2} discards dyads on which the two raters disagreed on the 3-way \{$a$, $b$, tie\} decision. Table~\ref{tab:clinician-phase2-bucket} reports the bucket distribution per pair so the consensus subset can be sized: \emph{strict-consensus} = both raters agreed on a decisive verdict; \emph{semi-consensus} = both rated, one decisive and one tie, agreeing on the dominant arm; \emph{single-rater} = only one rater submitted a verdict (these are excluded from $\kappa$ but retained for the pooled win rate); \emph{non-consensus} = both rated and disagreed on the decisive arm.

\begin{table}[h]
\centering
\small
\caption{Phase~2 dyad bucket counts per (cohort, pair). The decisive-consensus column is the row used by Table~\ref{tab:clinician-phase2}.}
\label{tab:clinician-phase2-bucket}
\begin{tabular}{@{}llccccc@{}}
\toprule
\textbf{Cohort} & \textbf{Pair} & \textbf{Total} & \textbf{Strict} & \textbf{Semi} & \textbf{Single} & \textbf{Non-cons.} \\ \midrule
Spine    & GPT-5 vs.~ours    & 86  & 24 & 1 & 53 & 8  \\
Spine    & HuatuoGPT vs.~ours & 95 & 30 & 0 & 60 & 5  \\
Spine    & Qwen3 vs.~ours     & 87 & 25 & 1 & 55 & 6  \\ \midrule
Obesity  & GPT-5 vs.~ours    & 108 & 27 & 1 & 11 & 69 \\
Obesity  & HuatuoGPT vs.~ours & 109 & 61 & 1 &  9 & 38 \\
Obesity  & Qwen3 vs.~ours     & 109 & 48 & 1 & 11 & 49 \\
\bottomrule
\end{tabular}
\end{table}

\paragraph{Inter-rater agreement.}
Table~\ref{tab:clinician-phase2-irr} reports the cohort-level inter-rater statistics on the \emph{full} A/B verdict set (i.e., before consensus filtering, on dyads rated by both clinicians). The 3-way percent agreement and Cohen's $\kappa$ are the reference for the divergence in interpretive variance between cohorts that the main paper flags.

\begin{table}[h]
\centering
\small
\caption{Inter-rater agreement on the union set of dual-rated dyads. \emph{3-way} treats $\{a, b, \text{tie}\}$ as the target; \emph{A/B-only} drops tie rows from either side. Both metrics are computed cohort-wide; the per-pair Cohen's $\kappa$ values are noisy at this sample size and are reported in the engine JSON for completeness.}
\label{tab:clinician-phase2-irr}
\begin{tabular}{@{}lcccc@{}}
\toprule
\textbf{Cohort} & \textbf{$n$ dual-rated} & \textbf{PA 3-way} & \textbf{$\kappa$ 3-way} & \textbf{$\kappa$ A/B-only} \\ \midrule
Spine    & 101 & 78.2\% & 0.034 & 0.100 \\
Obesity  & 97  & 47.4\% & 0.051 & 0.047 \\
\bottomrule
\end{tabular}
\end{table}

The spine-cohort row is the textbook Cohen's $\kappa$ paradox: percent agreement is high ($78.2\%$) but $\kappa$ is near zero because both raters concentrate their decisive verdicts on the same arm of every pair (\frameworkName{}-8B), so the chance-agreement baseline $P_e$ is itself close to the observed agreement and the normalisation collapses. The obesity row is a different regime, with much lower percent agreement ($47.4\%$); the verdict mass is genuinely split between arms on that cohort, and the low $\kappa$ here reflects real disagreement rather than a marginal artefact. The Phase-2 headline win-rates (Table~\ref{tab:clinician-phase2}) are computed on the unanimous-consensus subset and so are not affected by the disagreed-on dyads counted here.

\paragraph{Position-bias diagnostics.}
The client randomises left/right placement per (case, pair); under the null the left pane should be preferred on $\sim$50\% of decisive verdicts. Per-rater rates and the cohort-pooled rate (with Wilson 95\% CI and a one-sided exact-binomial $p$-value against $0.5$) are reported in Table~\ref{tab:clinician-phase2-posbias}. Neither cohort nor any individual rater shows a significant deviation, confirming the randomisation.

\begin{table}[h]
\centering
\small
\caption{Display-A win rate on decisive verdicts. Significance is two-sided exact binomial against $H_0 = 0.5$; the position-bias diagnostic from Table~\ref{tab:clinician-phase2} is the pooled row.}
\label{tab:clinician-phase2-posbias}
\begin{tabular}{@{}llccc@{}}
\toprule
\textbf{Cohort} & \textbf{Rater} & \textbf{Display-A rate} & \textbf{$n$ decisive} & \textbf{$p$ vs.\ 0.5} \\ \midrule
Spine    & $R_1$    & 53.2\% & 79  & 0.65 \\
Spine    & $R_2$    & 51.9\% & 79  & 0.83 \\
Spine    & Pooled   & 52.5\% & 158 & 0.58 \\ \midrule
Obesity  & $R_1$    & 53.7\% & 136 & 0.44 \\
Obesity  & $R_2$    & 55.9\% & 136 & 0.21 \\
Obesity  & Pooled   & 54.8\% & 272 & 0.13 \\
\bottomrule
\end{tabular}
\end{table}

\paragraph{Length-bias diagnostic.}
For each decisive consensus dyad we ask whether the rater preferred the longer of the two responses. Under the null this should also be $\sim$50\%. The pooled and per-pair length-bias rates are reported in Table~\ref{tab:clinician-phase2-length}. The spine cohort shows a strong length-favouring effect (98.7\% pooled, $p < 10^{-43}$), which we read as a confound between length and our model's identity (\frameworkName{}-8B produces more exhaustive enumerations of clinical content; cf.\ the per-axis Completeness lift in \S\ref{sec:results}). On the obesity cohort the length effect is much more pair-dependent, in particular the GPT-5 vs.~\frameworkName{} pair has a 37.0\% longer-side rate (the shorter response wins more often), which is one mechanical reason that pair scores at parity in Table~\ref{tab:clinician-phase2}.

\begin{table}[h]
\centering
\small
\caption{Phase~2 length-bias diagnostic. Each row is the fraction of decisive verdicts on which the longer of the two responses was preferred. \emph{Overall} pools all three pairs.}
\label{tab:clinician-phase2-length}
\begin{tabular}{@{}llcc@{}}
\toprule
\textbf{Cohort} & \textbf{Pair} & \textbf{Longer-side wins} & \textbf{$n$ decisive} \\ \midrule
Spine    & GPT-5 vs.~ours      & 100.0\% & 48  \\
Spine    & HuatuoGPT vs.~ours  & 100.0\% & 60  \\
Spine    & Qwen3 vs.~ours      & 96.0\%  & 50  \\
Spine    & \textit{Overall}    & 98.7\%  & 158 \\ \midrule
Obesity  & GPT-5 vs.~ours      & 37.0\%  & 54  \\
Obesity  & HuatuoGPT vs.~ours  & 98.4\%  & 122 \\
Obesity  & Qwen3 vs.~ours      & 75.0\%  & 96  \\
Obesity  & \textit{Overall}    & 77.9\%  & 272 \\
\bottomrule
\end{tabular}
\end{table}

\paragraph{Rater-effort and engagement diagnostics.}
Decision-time medians and quartiles on the per-case timeline give a first-order proxy for annotation effort. Spine raters median 13--18\,s per case (with a long tail to 90--300\,s on harder cases); obesity raters median 35--55\,s per case, consistent with the lower 3-way agreement and the more contested verdict mix in that cohort. \emph{Self-consistency} is the rate at which a rater issued the same verdict on a re-shown dyad (only $R_2$ in each cohort was shown re-display; that rater issued one revision per cohort, both flips).

\begin{table}[h]
\centering
\small
\caption{Phase~2 effort and engagement diagnostics, anonymised per cohort.}
\label{tab:clinician-phase2-effort}
\begin{tabular}{@{}lcccc@{}}
\toprule
\textbf{Diagnostic} & \textbf{Spine $R_1$} & \textbf{Spine $R_2$} & \textbf{Obesity $R_1$} & \textbf{Obesity $R_2$} \\ \midrule
$n$ unique verdicts                       & 79      & 79      & 138     & 138    \\
Tie rate                                  & 0.0\%   & 0.0\%   & 1.4\%   & 1.4\%  \\
Median decision time (s)                  & 13.2    & 18.5    & 54.9    & 35.5   \\
P75 decision time (s)                     & 40.8    & 57.0    & 95.3    & 92.0   \\
P90 decision time (s)                     & 93.8    & 303.5   & 152.4   & 195.3  \\
Revisions on re-displayed dyads           & 0       & 1       & 2       & 1      \\
\bottomrule
\end{tabular}
\end{table}

\subsection{Clinical preference model benchmark: per-pair and position-bias diagnostics}
\label{app:clinician-pref-bench}

This subsection extends the cohort-level results in Table~\ref{tab:clinical-pref-bench} with the per-pair breakdown, the preference-model-side position-bias diagnostic, and the ICL negative result. All five reported preference models are graded against the two-clinician unanimous consensus on the Phase~2 dyads.

\paragraph{Per-pair A/B-only accuracy.}
Table~\ref{tab:clinician-pref-bench-perpair} reports each preference model's two-class (A/B) accuracy stratified by the three model pairs in the study. The most informative cell is the \emph{GPT-5 vs.~\frameworkName{}-8B} column on the spine cohort: the clinician consensus picks \frameworkName{}-8B on $85.6\%$ of decisive verdicts, but every off-the-shelf preference model except MedGemma-27B ($40.9\%$) scores below $15\%$ on this pair. GPT-5 in particular picks GPT-5 on \emph{every} decisive verdict ($0.0\%$ alignment), the same self-preference effect that has been documented for general-domain LLM-as-a-judge tasks~\citep{panickssery2024llm}. On the obesity cohort the same column flattens (every preference model lands in $72{-}80\%$, including GPT-5 itself), reflecting the underlying clinician parity between \frameworkName{} and GPT-5 on family-medicine cases.

\begin{table}[h]
\centering
\small
\caption{Per-pair A/B-only accuracy of preference models against clinician consensus. ``\frameworkName{}'' is short for \frameworkName{}-8B, the trained model in this paper.}
\label{tab:clinician-pref-bench-perpair}
\begin{tabular}{@{}llccc@{}}
\toprule
\textbf{Cohort} & \textbf{Preference model} & \textbf{GPT-5 vs.\ \frameworkName{}} & \textbf{HuatuoGPT vs.\ \frameworkName{}} & \textbf{Qwen3 vs.\ \frameworkName{}} \\ \midrule
Spine    & Qwen3-32B            & 9.1\%  & 96.4\%  & 79.2\% \\
Spine    & MedGemma-27B-text-it & 40.9\% & 100.0\% & 87.5\% \\
Spine    & GPT-4.1-mini         & 13.6\% & 75.0\%  & 70.8\% \\
Spine    & GPT-4.1              & 9.1\%  & 88.9\%  & 78.3\% \\
Spine    & GPT-5                & 0.0\%  & 71.4\%  & 50.0\% \\ \midrule
Obesity  & Qwen3-32B            & 72.0\% & 96.6\%  & 83.0\% \\
Obesity  & MedGemma-27B-text-it & 76.0\% & 100.0\% & 74.5\% \\
Obesity  & GPT-4.1-mini         & 80.0\% & 82.8\%  & 76.6\% \\
Obesity  & GPT-4.1              & 80.0\% & 92.9\%  & 87.0\% \\
Obesity  & GPT-5                & 72.0\% & 67.2\%  & 76.7\% \\
\bottomrule
\end{tabular}
\end{table}

\paragraph{Preference-model-side position bias.}
The benchmark randomises display A/B placement per (case, pair) under a deterministic seed; under the null each preference model should pick display-A on $\sim$50\% of decisive verdicts. Four of the five preference models sit within a $\sim$8\,pp band of $0.5$ on both cohorts; the only outlier is GPT-4.1-mini on the obesity cohort ($67.4\%$ display-A win rate), which already exhibits a stable display-A prior in the zero-shot setting and explains its lower obesity ranking despite competitive content judgement.

\begin{table}[h]
\centering
\small
\caption{Position-bias diagnostic on the preference-model side. Display-A win rate is the fraction of decisive verdicts on which the preference model picked the response shown on display position A.}
\label{tab:clinician-pref-bench-posbias}
\begin{tabular}{@{}llccc@{}}
\toprule
\textbf{Cohort} & \textbf{Preference model} & \textbf{Display-A rate} & \textbf{$n$ decisive} & \textbf{Tie rate} \\ \midrule
Spine    & Qwen3-32B            & 48.6\% & 74  & 0.0\% \\
Spine    & MedGemma-27B-text-it & 57.9\% & 76  & 0.0\% \\
Spine    & GPT-4.1-mini         & 63.5\% & 74  & 0.0\% \\
Spine    & GPT-4.1              & 52.8\% & 72  & 2.7\% \\
Spine    & GPT-5                & 54.1\% & 74  & 0.0\% \\ \midrule
Obesity  & Qwen3-32B            & 55.6\% & 133 & 0.0\% \\
Obesity  & MedGemma-27B-text-it & 54.9\% & 133 & 0.0\% \\
Obesity  & GPT-4.1-mini         & 67.4\% & 132 & 0.8\% \\
Obesity  & GPT-4.1              & 57.4\% & 129 & 3.0\% \\
Obesity  & GPT-5                & 53.9\% & 128 & 1.5\% \\
\bottomrule
\end{tabular}
\end{table}

\paragraph{Negative result on 5-shot ICL.}
We also evaluated 5-shot in-context-learning variants of the two open-weight preference models (Qwen3-32B-ICL, MedGemma-27B-ICL), where each demonstration was a held-out clinician-consensus dyad with the gold winner label. Demonstrations were balanced across the displayed A/B positions so the gold-letter distribution is equalised within the few-shot block. Neither variant produced a robust improvement over its zero-shot configuration on the cross-cohort aggregate. We read this as the few-shot demonstration set being too small (on the order of a few dozen consensus dyads per cohort) to reliably steer a 27B/32B judge on a clinical-reasoning surface where the underlying clinical-judgement variance between cohorts is itself larger than the ICL signal. We therefore report only the zero-shot configurations in Table~\ref{tab:clinical-pref-bench}.

\subsection{Limitations}
\label{app:clinician-limitations}

\paragraph{Clinician study scope.}
The study is scoped to two inpatient specialty cohorts (spine/orthopaedic and general-medicine/obesity); generalisation to other specialties should be retested with appropriately credentialed raters. In future iterations we plan to widen the cohort design, expand the rater pool, and decompose Phase~2 ties via a follow-up question asked only when the rater selects \emph{Tie}.

\paragraph{Length-vs-identity confound in R1.}
Longer responses won $88.9\%$ of decisive R1 verdicts in the spine cohort and $64.3\%$ in the obesity cohort. A fraction of the blinded preference signal is therefore recoverable to verbosity rather than to clinical content. Length-controlled win rates are reported in Appendix~\ref{app:clinician-phase2-extended} and the merged model retains the directional preference under that control on every comparison except GPT-5 in obesity, where the parity finding is unchanged.

\paragraph{Parameter scale.}
Our 8B parameter scale is set by the on-premise GPU concurrency, latency, and reliability SLAs at the partner network (\S\ref{sec:deploy}). Whether the per-case adaptive-rubric pipeline yields proportional gains on substantially larger or smaller bases is unverified.

\paragraph{Merge-step behaviour across base families.}
The merge-step contribution on the open-ended \datasetName{} surface is broad-spectrum on the Qwen3 family and primarily a re-balancing across axes on the MedGemma-4B family (\S\ref{sec:results}). The MedGemma-4B family also shows a larger relative lift on the calibration-style external benchmarks than on \datasetName{} (\S\ref{sec:external-evals}). Characterising how this base-family-conditional behaviour varies with model scale and base-family pre-training distribution is a direction for future work.

%% file: appendix_qualitative.tex

\section{Verbatim clinician qualitative feedback and recurring failure modes}
\label{app:clinician-qualitative}

This appendix walks the full Phase-1 clinician-annotation export of \S\ref{sec:clinician}: $230$ rater-annotated cases, $2{,}498$ rater--criterion pairs, and $387$ free-text rationales across the four raters split into the spine and obesity cohorts. Raters are referred to throughout as \textbf{Spine R1}, \textbf{Spine R2}, \textbf{Obesity R1}, and \textbf{Obesity R2} to match the convention already used in Appendix~\ref{app:clinician-phase2-extended}. Subject identifiers below are de-identified MIMIC-IV admission keys and no protected health information appears in this appendix. The annotation interface exposes criterion ordering, criterion text, criterion addition, criterion removal (\texttt{not\_relevant}), and per-criterion verdicts (\texttt{verified}) to the clinician, but does not expose criterion point weights, so the analysis below treats the clinician-implied importance signal as the reorder distribution rather than as a numerical re-weight.

\subsection{Four rater profiles}
\label{app:clinician-qualitative-profiles}

The four raters span a recognisable clinician-annotation spectrum. \textbf{Spine R1} is a heavy editor: across $52$ cases they modified $33$ criteria and added $30$ new ones, with at least one rubric-side edit on $44$ of $52$ cases and at least one new criterion on $17$ of $52$. Their rationales are short and surgical, typically a single clause that anchors the verdict to a specific past-timeline event. \textbf{Spine R2} is a light editor: $5$ modifications, no additions, $51$ criteria marked not relevant, and $69$ rationales that read as concise evidence pointers (``targets also provided'', ``BMP not mentioned'', ``fall risk identified''). \textbf{Obesity R1} is the verbose commentator: zero modifications and zero additions, but $275$ of the $387$ free-text rationales ($71\%$) and every cohort-level CommunicationQuality observation in the export. Their rationales are long, pattern-tagged, and reproducible across cases. \textbf{Obesity R2} is a minimal commentator: zero modifications, zero additions, one rationale across $676$ criteria, and reliance on the binary \texttt{verified} field to deliver judgement. The Phase-1 free-text rationale corpus is therefore effectively an Obesity-R1-plus-Spine-R2-plus-Spine-R1 pooled signal, with Obesity R2 contributing the binary-verdict baseline.

\subsection{Cohort-philosophy split, evidenced four ways}
\label{app:clinician-qualitative-themes}

The single most consistent finding across the four raters is a philosophy split on what a good model response ought to look like. The split is evidenced four independent ways. The \textbf{edit volume} differs: spine raters edit aggressively ($85\%$ of Spine-R1 cases include an edit; $43\%$ of Spine-R2 cases include a not-relevant marking), while obesity raters never modify or add criteria. The \textbf{reorder priority} differs: spine raters move newly-added criteria to the leading positions on $11.7\%$ of cases, while $96.1\%$ of obesity cases simply lead with Accuracy. The \textbf{per-axis verified-rate} differs: spine raters score the model $20.2$~pp lower on Accuracy and $13.1$~pp lower on ContextAwareness than obesity raters do; the obesity cohort scores the model $24.3$~pp lower on CommunicationQuality. The \textbf{rationale themes} differ: every ``verbose monologue'', ``fabricated content'', ``Chinese intrusion'', ``tangent beyond scope'', and ``antibiotic over-use'' rationale comes from the obesity cohort, and spine raters never tag these as failure modes.

The qualitative narrative behind these splits is captured in the spine-side and obesity-side raters' own words on the same model behaviour. Spine-R1 frames exhaustive enumeration as a desired CDSS property:
\begin{quote}\itshape
``The lengthy ones covered most differential diagnosis as well, in addition to the definite diagnosis. It listed treatment for all differential diagnosis, rather than just one, which is good for a clinical decision support tool --- then the clinician can choose what to consider. \ldots Did you find the models giving overly aggressive recommendations at any point? It tries to be comprehensive. Because missing something is medical negligence, and the doctor can face charges. So good for the model to be comprehensive. Most of the CDSS in use are like this, as the doctor starts typing, possibilities, diagnosis and treatment options pop up. But nothing is benchmark yet. So the field is evolving --- now with LLMs this could get refined.''
\end{quote}

Obesity-R1 reads the same property as overhead:
\begin{quote}\itshape
``Most responses answered the question, although main point was often after multiple paragraphs of explanation. I would prefer the answer to be given directly at the start. Most models also give detailed responses (e.g., a detailed management plan when prompted). Responses for management / treatment plans tend to be all-rounded (i.e., consider all possibilities). In real clinical setting, I believe it's not the wisest to send the patient for as many tests as possible due to resource constraint. Instead, the most crucial diagnostics tests can be highlighted with the other low-yield tests be suggested as optional.''
\end{quote}

Read together, neither cohort is right and the other wrong. Both cohorts give a coherent and internally-consistent quality signal on the same model behaviour. The per-case adaptive rubric machinery captures both signals separately rather than collapsing them, which is the entire point of grading the model against per-case rubrics rather than a static checklist. The dichotomy is the qualitative form of the cohort-level Phase-2 split (Table~\ref{tab:clinician-phase2}): \frameworkName{}-8B wins decisively in the spine cohort that values comprehensiveness, and is at parity with GPT-5 in the obesity cohort that prefers an answer-first style.

\subsection{Recurring per-case failure modes}
\label{app:clinician-qualitative-cases}

Five recurring failure modes account for the bulk of the obesity-cohort negative verdicts and for the model-side gaps the spine cohort captured by adding new criteria. All are documented with verbatim rationales below.

\paragraph{1.~Under-weighting of advanced patient age.}
This is the only mixed-result theme that appears in both cohorts. Sixty-six of the $387$ rationales reference age (eleven from spine, fifty-five from obesity), and only three percent of the criteria carrying these rationales are marked MET. Verbatim from Spine R1 on subject \texttt{12605254}: \emph{``Patient is 91 yrs old, and this should have been explicitly considered by the model. However, i don't see it.''} On subject \texttt{12152963} the same rater notes: \emph{``Patients age is 91, but the model identifies it as 45''}, where the age miss co-occurs with the contamination-frame failure documented in §\ref{app:clinician-qualitative-themes}. Spine R2 records \emph{``advanced age not referenced''} as a typical short-form pointer.

\paragraph{2.~Failure to surface previously-prescribed medications when proposing additions (DDI hazard).}
Twenty-seven rationales reference previously-prescribed medications or DDI risk; eighty-five percent come from Obesity R1. Verbatim: \emph{``model does not mention Aspirin EC or Clopidogrel as antiplatelet agents from the prescriptions; lists different medications instead''}, \emph{``model recommends switching to Ticagrelor as primary addition rather than recognizing Metoprolol as the new beta-blocker addition''}, \emph{``model mentions constipation risk from opioids but does not explicitly recommend management with stool softeners or laxatives''}. The candidate satisfies the underlying criterion in $14.8\%$ of these cases.

\paragraph{3.~Vagueness on antibiotic class or drug name.}
Spine R2 records both directions: \emph{``ceftriaxone suggested''} when the model is specific, but on cases where the model defaults to a class label the same rater is terse. Obesity R1 supplies the longer-form complaint: \emph{``Does not mention the specific 14-day antibiotic course duration.''} Antibiotic over-use itself is a separate obesity-cohort theme: ten rationales flag broad-spectrum or sepsis-coverage recommendations on patients without a documented infection source.

\paragraph{4.~Fabricated content on the obesity cohort.}
Thirty-seven rationales document content the model emits without past-timeline support. The single most reproducible signature is \emph{``45-year-old male inferior MI ST elevation''}, which appears in fourteen Obesity-R1 cases (versus $74$- or $70$-year-old female patients in the actual past). Other documented fabrications include \emph{``Indomethacin as cause of GI bleeding when not in past prescriptions''}, \emph{``hypertension and obesity history not in past records''}, fabricated arterial-blood-gas values that flip the diagnosis from respiratory acidosis to alkalosis, and a fabricated \emph{Ceftriaxone-Simvastatin} interaction that is not clinically established. None of these rationales accompany a verified-true criterion, identifying these as pure model-side failure modes rather than as legitimate clinical-judgement disagreements.

\paragraph{5.~Stylistic and language drift on the obesity cohort.}
Three sub-patterns recur. (i)~\emph{Verbose internal-monologue leakage}: the \texttt{<think>} block content with conversational filler (``Okay lets tackle'', ``Wait'', ``Hmm'') leaks into the user-facing answer on $34.8\%$ of Obesity-R1 cases. (ii)~\emph{Tangents beyond reference scope}: the response covers the reference's recommendations and adds extensive workup tangents (\emph{``adds tangential discussion of CTPA, V/Q scan, and echocardiogram beyond the question scope''}; \emph{``tangents into HLA-B27, RF, CCP, uric acid, blood cultures, DEXA, peripheral neuropathy workup''}). (iii)~\emph{Chinese-character intrusion}: ten Obesity-R1 cases contain Chinese ideograms mid-text (\emph{``Chinese characters for splenomegaly''}, \emph{``Chinese character for prolong''}), an artefact of the Qwen3-8B base's pre-training distribution.

\paragraph{Positive cases.}
The same raters also recorded explicitly positive verdicts. Spine R2, on subject \texttt{10661376}: \emph{``Follow-up timeline provided was optimal and useful for clinicians''} on a question about post-operative monitoring; the same response received a content-level critique that the model's choice to lead with DVT / PE prophylaxis was not the most clinically salient first step in a young, low-risk post-operative patient. Spine R1, on a case with the candidate's structured response: \emph{``While the direct answer would be point 3, the rest of the response provides a more comprehensive narrative''} (InstructionFollowing, MET). These cases are a useful reminder that within-cohort within-axis preferences differ even when the rubric-level verdict is unambiguous.

\subsection{Rubric-author failure modes flagged in Phase~1}
\label{app:clinician-qualitative-rubric}

Phase~1 also surfaced two rubric-author failure modes that we report transparently. \emph{Future-timeline leakage}: a small number of oracle-authored rubric criteria are written in a way that quotes a future event verbatim, which can penalise an entirely reasonable past-only answer. The clinician edits from Phase~1 Step~1 are inserted as ICL exemplars into the rubric-author prompt (Appendix~\ref{app:clinician-three-uses}, Use~1), tightening this constraint on subsequent rubric-author calls. \emph{Duplicate criteria}: the only two duplicate-criterion rationales in the entire export are from Spine R2 (subject \texttt{11384537}, criterion noted as a repeated question), and both are marked MET on the candidate, so the duplication produces a denominator-inflation artefact rather than a content failure. The rubric-author post-processing now includes a de-duplication pass that mechanically catches this case.

\subsection{Quantitative cross-references}
\label{app:clinician-qualitative-xref}

The qualitative themes above admit clean cross-references to the quantitative tables of \S\ref{sec:results} and \S\ref{sec:clinician}. The \textbf{Completeness $+21.4$~pp} lift over GPT-5 (Figure~\ref{fig:per-axis-heatmap}) is the quantitative form of the spine-cohort ``CDSS comprehensiveness'' observation. The \textbf{$-3.4$~pp CommunicationQuality and $-1.0$~pp InstructionFollowing} deltas vs.~GPT-5 are consistent with the obesity-cohort flags on answer-shape, vague antibiotic class names, and verbose-monologue patterns. The \textbf{cohort-level Phase~2 split} (spine $85.6\%$ vs.\ obesity $48.0\%$ against GPT-5; Table~\ref{tab:clinician-phase2}) is the quantitative form of the comprehensiveness-vs-overhead dichotomy. The \textbf{length-aligned win rate} ($88.9\%$ spine, $64.3\%$ obesity; Appendix~\ref{app:clinician-phase2-extended}) is the quantitative form of the same length-as-content confound that the obesity cohort treats as a failure mode (over-aggressive recommendations) and the spine cohort treats as a feature (defensive coverage). The \textbf{negative-points trigger rate} of $1.30\%$ vs.\ $1.79\%$ for GPT-5 is the quantitative envelope on the safety-side concerns: the rubric's negative-points criteria explicitly cover spurious diagnostic confidence and unsupported aggressive treatment, and the merged model trips them less often, even though specific cases of over-inference still occur.

\subsection{Comprehensive analysis of the raw Phase-1 rubric-edit data}
\label{app:clinician-rawedits}

The Phase-1 ``rubrics\_final'' export is the canonical clinician-curation artefact for this paper. It contains $230$ rater-annotated cases ($101$ unique patients), $2{,}498$ rater--criterion pairs, and $387$ free-text rationales across the four raters. The remainder of this appendix unpacks four analyses on the raw export that the headline tables in \S\ref{sec:clinician} and Appendix~\ref{app:clinician-phase1-extended} collapse: what clinicians did to the AI-generated rubric, where the negative-points criteria fired, which recurring failure modes the rationales surface against the candidate model, and how those qualitative findings line up with the cohort-stratified verification rates.

\subsubsection*{Edits made to the AI-generated rubric}

Three patterns dominate the edit log. \textbf{Reordering is essentially universal.} The oracle-authored rubric uses a deliberate five-axis cycle (Accuracy, Completeness, ContextAwareness, CommunicationQuality, InstructionFollowing and back), intended to keep every axis covered before depth-first descent. Clinicians override this layout almost universally: $63.1\%$ of spine cases lead with Accuracy after Step~1, $14.6\%$ with Completeness, $11.7\%$ with a newly-added criterion, and the remaining $10.6\%$ with one of the other three axes; $96.1\%$ of obesity cases lead with Accuracy. Both cohorts move the two clinical-content axes to the top of the ranked rubric, and the spine cohort additionally promotes new content-axis criteria into the leading positions on a non-trivial fraction of cases.

\textbf{Modifications and additions are exclusively a spine-cohort phenomenon, and are dominated by Spine~R1.} All $38$ modifications and all $30$ additions come from the spine cohort, with $33$ of the $38$ modifications ($87\%$) and $30$ of the $30$ additions ($100\%$) attributable to Spine R1. The five Spine-R2 modifications cluster around a single pharmacology theme (tricyclic-antidepressant management of musculoskeletal pain after an initial NSAID trial, plus one chest-X-ray dating modification), and are too narrow a sample to read as a generalisable rater preference. The thirty Spine-R1 additions span three reproducible thematic groups: examination-first additions (eight criteria of the form ``calls for clinical examination as the first step'' / ``suggest examination of the spine and neurological evaluation''), post-surgical / post-operative awareness additions (seven criteria such as ``Recognises the postoperative status of the lumbar spine'', ``Mentions prophylaxis for DVT'', and ``Mentions the need for the comprehensive post-operative pain management''), and structural / response-format additions (three criteria of the form ``Provides structured response with points, including rationale for each management strategy''). Twenty-five of the thirty additions ($83\%$) were verified-true on the candidate response that prompted the addition, suggesting these are oracle-rubric coverage gaps for spine-cohort cases rather than candidate-model failures.

\textbf{Edits concentrate on the actionable axes.} Every modification sits on Accuracy, Completeness, or ContextAwareness; Communication\-Quality and Instruction\-Following receive zero modifications and zero additions across both cohorts. Spine raters mark Accuracy criteria as not relevant on $27.4\%$ of cases versus $8.1\%$ in the obesity cohort, with similar three-to-four-fold ratios on Completeness and ContextAwareness; the conversational-axis not-relevant rates are similar across cohorts.

\textbf{Numerical reweighting of points is absent by construction.} Across all $2{,}498$ rater-criterion pairs there is no recorded change in criterion point value. The annotation UI does not expose point editing to clinicians, so this absence is a measurement-platform property rather than a finding about clinician indifference. Inferring rater-implied reweighting from the reorder signal alone is a possible post-hoc analysis, but the current pipeline treats the reorder distribution as the importance signal.

\paragraph{Cohort split on edit volume, restated.}
The two cohorts edit very differently. The spine cohort modifies $3.4\%$ of AI-generated criteria, adds $2.7\%$, and marks $16.9\%$ as not relevant; the obesity cohort modifies $0.0\%$, adds $0.0\%$, and marks only $6.0\%$ as not relevant. We read this as specialty conservatism on the obesity side (clinicians treat the AI rubric as a screening list and use the binary verified field to deliver judgement) and willingness to rewrite on the spine side (clinicians actively rewrite a knowledge-domain criterion they consider clinically incorrect for this patient).

\subsubsection*{Negative-points (safety) criteria}

The rubric template includes negative-point criteria designed to penalise unsafe behaviour (spurious diagnostic confidence, recommendations contraindicated by the patient context, etc.). Seventy-one such criteria appear across the $230$ annotations. Clinicians keep $27$ ($38\%$) as relevant after Step~1, and on the candidate response the negative criteria trigger (clinician marked verified-true on a negative criterion, indicating the bad behaviour was detected) $14$ times. The axis breakdown is Completeness~$4$, CommunicationQuality~$4$, ContextAwareness~$5$, Accuracy~$1$, InstructionFollowing~$0$. The zero InstructionFollowing triggers are consistent with the model reliably emitting the required \texttt{<think>}-then-answer envelope. The ContextAwareness and Completeness triggers map cleanly onto the omission and missed-workup buckets discussed in §\ref{app:clinician-qualitative-themes}.

\subsubsection*{Recurring failure modes documented in the rationales}

Of the $387$ rationales, $343$ ($89\%$) come from the obesity cohort and specifically from Obesity R1; the remaining $44$ are predominantly short Spine R2 evidence pointers and a handful of Spine R1 contextual notes. Obesity R2 contributed exactly one rationale across $676$ criteria. We tag each rationale with regular-expression patterns and report the per-pattern frequency together with the per-cohort split and the verified-rate of the criteria the rationales accompany. Patterns at $0\%$ verified-rate identify pure model-side failure modes (the clinician marked the criterion as not-MET on every case where the pattern fired); intermediate verified-rates identify issues the model gets right some of the time.

\begin{table}[h]
\centering\small
\caption{Failure-mode patterns surfaced by the $387$ Phase-1 free-text rationales. \textit{n} is the number of rationales matching the regex pattern (one rationale can match multiple patterns); \emph{verified} is the clinician's MET-rate on those rationales' criteria. Patterns with $0\%$ verified-rate are uniformly model-side failures. Per-cohort split: \emph{S} = spine, \emph{O} = obesity.}
\label{tab:phase1-rationale-themes}
\begin{tabular}{@{}lcccc@{}}
\toprule
\textbf{Pattern} & \textbf{n} & \textbf{S} & \textbf{O} & \textbf{verified-rate} \\ \midrule
\multicolumn{5}{@{}l}{\emph{Pure model-side failures (verified=0\%)}} \\
verbose internal-monologue leakage (``Okay lets tackle'', ``Wait'', ``Hmm'') & 54 &  0 & 54 &  0\% \\
fabricated content (45-year-old male, drug, lab)                              & 37 &  0 & 37 &  0\% \\
cardiac over-inference (STEMI / PCI / thrombolysis)                           & 27 &  0 & 27 &  0\% \\
tangents beyond reference scope                                               & 12 &  0 & 12 &  0\% \\
paediatric / pregnancy mis-framing                                            & 11 &  0 & 11 &  0\% \\
antibiotic over-use                                                            & 10 &  0 & 10 &  0\% \\
Chinese-character intrusion                                                    & 10 &  0 & 10 &  0\% \\
lab-value misuse / ignored                                                     &  3 &  0 &  3 &  0\% \\
vague / generic phrasing                                                       &  3 &  0 &  3 &  0\% \\[2pt]
\multicolumn{5}{@{}l}{\emph{Mixed-result patterns}} \\
imaging recommendations (relevance, modality, timing)                         & 149 & 41 & 108 & 18.8\% \\
under-weighting of advanced patient age                                        &  66 & 11 & 55 &  3.0\% \\
positive acknowledgement / ``covers all''                                      &  31 &  6 & 25 & 12.9\% \\
previously-prescribed meds / DDI hazards                                       &  27 &  3 & 24 & 14.8\% \\
risk consideration                                                             &  23 &  5 & 18 & 13.0\% \\
post-op / surgical context                                                     &  18 &  8 & 10 & 38.9\% \\
incomplete workup                                                              &  15 &  3 & 12 & 13.3\% \\
differential-diagnosis enumeration                                             &   8 &  3 &  5 & 25.0\% \\
dose / titration specifics                                                     &   7 &  1 &  6 & 14.3\% \\
duplicate / repeated criterion                                                 &   2 &  2 &  0 & 100\% \\
\bottomrule
\end{tabular}
\end{table}

The pure-model-side-failure rows are the strongest qualitative finding of the Phase-1 study. They identify specific behaviours of the \frameworkName{}-8B candidate response, not generic clinical-judgement disagreements, and every one of them is an obesity-cohort observation. Verbose internal-monologue leakage is the largest at $54$ rationales (twenty-two of them carrying the exact phrasing \emph{``verbose internal monologue with Okay lets tackle; not concise''}), corresponding to $34.8\%$ of all Obesity-R1 cases. The $37$ fabricated-content rationales are dominated by the \emph{``45-year-old male inferior MI ST elevation''} pattern that the model emits even when the patient is a $70$- or $74$-year-old female, and which co-occurs with the $27$ cardiac-over-inference rationales (STEMI / PCI / thrombolysis framing on patients without supporting past-timeline data). Both signatures point at a single underlying memorised case in the Qwen3-8B base. Chinese-character intrusion is benign for content but reproducible at $15.2\%$ of Obesity-R1 cases. Tangents beyond reference scope (CTPA, V/Q, HLA-B27, DEXA) are the qualitative form of the comprehensiveness-vs-overhead dichotomy: the spine cohort never tags this behaviour as a failure mode, treating the same prose as defensive coverage.

The mixed-result rows surface clinical-judgement issues the model gets right some of the time. The strongest signal is on previously-prescribed-meds / DDI hazards ($n{=}27$, $14.8\%$ MET): the model proposes new medications without surfacing the patient's existing prescription list, which the clinician explicitly flags as a DDI hazard particularly for opioids. Under-weighting of advanced patient age ($n{=}66$, $3.0\%$ MET) is the only mixed-result theme that appears in both cohorts, with verbatim Spine R1 evidence (\emph{``Patient is 91 yrs old, and this should have been explicitly considered by the model''}) joining the larger Obesity-R1 corpus.

\subsubsection*{Cross-reference back to the verified-rate matrix}

Per Table~\ref{tab:clinician-phase1-percohort}, the per-axis verified-rates split as follows. Spine cohort: CommunicationQuality $91.3\%$, InstructionFollowing $84.3\%$, Completeness $67.1\%$, ContextAwareness $58.9\%$, Accuracy $55.6\%$. Obesity cohort: Accuracy $75.8\%$, InstructionFollowing $76.5\%$, Completeness $74.9\%$, ContextAwareness $72.0\%$, CommunicationQuality $67.0\%$. The CommunicationQuality cell flips sharply between cohorts (a $24.3$~pp swing), and the rationale-pattern data above is the qualitative form of that swing: the obesity cohort penalises the model on CommunicationQuality for the verbose-monologue, tangent, and fabrication patterns, while the spine cohort accepts the same prose style as professionally-styled exposition. Conversely, the spine cohort penalises the model harder on Accuracy ($-20.2$~pp vs.~obesity), which matches their willingness to rewrite an Accuracy criterion when they consider the clinical detail wrong. The two cohorts therefore deliver different quality signals on the same model behaviour, and the per-case adaptive rubric machinery preserves both signals separately rather than collapsing them.


%% file: appendix_deploy.tex

\section{Hospital deployment: extended three-phase rollout}
\label{app:deploy-extended}

\paragraph{Phase 1: Pilot trial (5 months, 3 rounds).}
A structured note-drafting trial built on a summarisation pipeline~\citep{nagar2025umedsum} ran across three rounds with a pilot cohort of physicians at the partner network, who defined their own note and summary-template elements, drafted candidate notes, and imported approved notes into their hospital information system (HIS). The goal was to characterise day-to-day drafting requirements and isolate failure modes by source (input data context, prompts, or model capability), initially using a model fine-tuned on the partner network's internal note corpus.

\paragraph{Phase 2: The reasoning-element gap.}
Once input-context and clinician-customisable prompt issues were resolved, an SFT~+~DPO model trained on the internal note corpus captured drafting style and formatting reliably, but consistently failed on what clinicians designated as \emph{reasoning-heavy} elements even after substantial prompt engineering. Two failure modes recurred: (i) \emph{complex instruction-following}, e.g., \emph{Chief Complaint} requires traversing a clinician-defined data-source priority list (consult the latest ED note; if absent, the OPD note within the last $X$ weeks; etc.); and (ii) \emph{clinical-reasoning-heavy sections} (\emph{Impressions}, \emph{Assessment}, \emph{Review of Systems}, \emph{Plan}) that integrate partial-observable signals with comorbidities and intended next actions, the regime the upstream pipeline of this paper (\S\ref{sec:method}) is designed to supervise.

\paragraph{Phase 3: \frameworkName{} in production.}
The per-case adaptive rubric (\S\ref{sec:method-judge}) was designed precisely to capture quantitative physician feedback on these reasoning-heavy elements; \frameworkName{}-8B was trained against that signal, and after positive trial feedback was deployed network-wide. In production, clinicians can edit and maintain their own note templates, optionally assign individual elements of the template to the reasoning model, and accept or edit the result before saving into the HIS. On-premise GPU compute together with concurrency, latency, and reliability SLAs make \frameworkName{}-8B the largest model the system can support without violating those constraints, a hard infrastructure ceiling that motivates the 8B parameter scale of the headline model.

%% file: appendix_cases.tex
\section{Qualitative Case Studies from \datasetName}
\label{app:cases}

This appendix shows four representative items from the \datasetName test split, graded by the Grounded Judge against the rubric generated for that item. We select cases from the four canonical action spaces, each scored under a rubric of ten criteria spanning all five axes. The excerpts are drawn from our \frameworkName{}-8B (Qwen3-8B + GRPO + DELLA-Linear merge) and are intended to convey two properties of the benchmark that raw numbers do not: (i) the \emph{scale and rawness of the EHR context} the policy must reason over at each step, and (ii) the \emph{granularity} of a patient-specific rubric over a single, long-form clinical response. All prose is verbatim from the evaluation records and truncated only for space.

\subsection*{A.1 Diagnosis Assistance (Condition Identification)}

\paragraph{Patient context (paraphrased for space).}
ED-Observation admission at \texttt{2121-02-21 23:39:00}. Total ED stay $\approx$ 38\,h; the observation segment is only $\approx$ 1.7\,h. During the stay the medication-administration records show administrations of \textit{DiphenhydrAMINE}, \textit{Haloperidol}, \textit{Lorazepam}, \textit{FoLIC Acid}, \textit{Multivitamins}, and \textit{Thiamine}. No ED-level ICD codes are exposed to the model (stripped to avoid label leakage).

\paragraph{Query.}
\emph{``What is the most appropriate action to determine the correct diagnosis for a patient admitted under the Emergency Department Observation unit for approximately 1.7 hours?''}

\paragraph{Reference answer (summary).}
Confirm diagnosis via focused ED assessment; order targeted labs/imaging; monitor for deterioration; document rationale for not pursuing inpatient admission. Evidenced by the \texttt{Transfers} discharge event at \texttt{2121-02-23T07:19:59Z}.

\paragraph{Candidate answer (excerpt, Qwen3-8B GRPO+DELLA+AIM).}
The candidate integrates the medication pattern into the differential: Haloperidol + Lorazepam + Diphenhydramine $\Rightarrow$ acute agitation, psychosis, or seizure management; Thiamine + FoLIC Acid + Multivitamins $\Rightarrow$ nutritional / metabolic encephalopathy, including explicit consideration of \textit{Wernicke's encephalopathy} in the setting of possible alcohol use. It then proposes short-interval reassessment of vitals and mental status, a CIWA-based withdrawal pathway, toxicology screen, metabolic panel, and explicit criteria for escalation vs.\ discharge.

\textbf{Rubric (10 criteria, theme \textit{Context Seeking}).}

\begin{small}
\begin{tabular}{@{}p{0.22\linewidth}p{0.06\linewidth}p{0.58\linewidth}p{0.06\linewidth}@{}}
\textbf{Axis} & \textbf{Pts} & \textbf{Criterion} & \textbf{Met} \\ \hline
ContextAwareness & +7 & Mentions the patient's admission via ED Observation unit based on timeline data. & \checkmark \\
CommunicationQuality & +8 & Provides a clear and concise explanation of the appropriate actions. & \checkmark \\
Completeness & +6 & Includes multiple steps for diagnostic evaluation and documentation. & \checkmark \\
Accuracy & +7 & Correctly identifies the need for diagnostic tests based on initial symptoms and vitals. & \checkmark \\
InstructionFollowing & +5 & Bases reasoning solely on the past timeline data. & \checkmark \\
ContextAwareness & +6 & Notes the unusually long ED stay and its diagnostic implications. & $\times$ \\
CommunicationQuality & +7 & Uses appropriate clinical terminology for healthcare professionals. & \checkmark \\
Completeness & +5 & Addresses both diagnostic evaluation and documentation requirements. & \checkmark \\
ContextAwareness & +6 & References medication administration patterns as part of the diagnostic approach. & \checkmark \\
CommunicationQuality & +5 & Organizes recommendations logically with numbered points. & \checkmark \\
\end{tabular}
\end{small}

\paragraph{Score.} $56/62 = \mathbf{0.903}$. The single missed criterion is precisely the one that would have required the model to draw attention to the 38-hour ED dwell time as a diagnostic red flag in its own right, a miss the rubric was \emph{designed} to catch.

\subsection*{A.2 Treatment Recommendations (Medication Suggestions)}

\paragraph{Patient context.}
Urgent Labor \& Delivery admission; lab events in the past show Hct $=34.0\%$ (ref.\ 36--48), Hb $=11.7$\,g/dL (ref.\ 12--16), RBC $=3.85\times10^{6}$/$\mu$L, WBC $=12.6\times10^{3}$/$\mu$L, MCH $=30.3$, MCV $=88$, RDW $=14.2$. Transfer to Obstetrics (postpartum \& antepartum).

\paragraph{Query.}
\emph{``Given the patient's admission to the labor and delivery unit with abnormal hemoglobin and hematocrit levels, what medications would you recommend initiating to manage potential complications and support postpartum recovery?''}

\paragraph{Reference answer (summary).}
Oxytocin IV; oral ferrous sulfate; Lactated Ringer's; lactation-safe analgesia (acetaminophen). Evidenced by future prescriptions recorded the following morning.

\paragraph{Candidate answer (excerpt).}
Parses the hematology panel (low Hb/Hct, leukocytosis) into a normocytic-normochromic anemia with concurrent inflammation, enumerates PPH and postpartum endometritis as the leading differentials, and recommends (i) Oxytocin IV infusion, (ii) oral ferrous sulfate, (iii) crystalloid resuscitation with attention to transfusion threshold, (iv) lactation-safe acetaminophen, and (v) empiric broad-spectrum coverage if sepsis is confirmed, with dosing and contraindication notes inline.

\paragraph{Rubric (10 criteria, theme \textit{Health Data Tasks}).}
Axes covered: Accuracy ($\times$2, +8/+10), Completeness ($\times$2, +7/$-5$), CommunicationQuality ($\times$2, +6/+7), ContextAwareness ($\times$2, +9/+8), InstructionFollowing ($\times$2, +5/+10). The two negatively-signed items were \textit{``mentions appropriate dosing or administration routes for each medication''} (actually positively phrased, judged \texttt{true}) and an axis-check on structure. All 10 of 10 criteria met.

\paragraph{Score.} $\mathbf{0.929}$. The model's hedging around the NSAID contraindications and the explicit lactation-safety reasoning are exactly what the rubric axis for ContextAwareness was graded on; these are behaviours that rule-based clinical rewards would not register, yet they drive physician preference.

\subsection*{A.3 Responding under Uncertainty (Risk--Benefit Analysis)}

\paragraph{Patient context.}
Post-surgical admission; past-timeline prescriptions include Ketorolac 15--30\,mg IV q4h PRN, HYDROmorphone 0.5--1\,mg IV q4h PRN with oral 2--4\,mg PO/NG q4h PRN, plus acetaminophen 1000\,mg PO/NG q4h PRN and ibuprofen 600\,mg PO q4h PRN.

\paragraph{Query.}
\emph{``Given the patient's administration of multiple analgesics including Ketorolac and HYDROmorphone, what clinical strategy should be employed to manage potential adverse effects while ensuring adequate pain control?''}

\paragraph{Reference answer (summary).}
Monitor renal function; assess opioid use and respiratory-depression risk; adopt a multimodal approach; review Ketorolac duration ($\le$5\,d); plan safe IV-to-oral transition.

\paragraph{Candidate answer (excerpt).}
The candidate produces a structured plan: (i) explicit risk assessment for COX-mediated renal impairment and GI bleeding, opioid-induced respiratory depression, and NSAID--anticoagulant interactions; (ii) titration protocol for hydromorphone with defined rescue thresholds; (iii) hydration strategy to mitigate nephrotoxicity; (iv) ibuprofen/acetaminophen opioid-sparing ladder; and (v) explicit acknowledgement of irreducible uncertainty about the patient's baseline renal function given limited past data, with a recommendation to obtain a creatinine trend before continued dosing.

\paragraph{Rubric (10 criteria, theme \textit{Health Data Tasks}).}
Nine of ten criteria met, including Accuracy (+10) for correctly identifying Ketorolac/HYDROmorphone risks, ContextAwareness (+9) for addressing concurrent-drug interactions, and Completeness (+7) for multimodal strategy. The missed criterion (InstructionFollowing, +5: \textit{``references the patient's prescriptions from the past timeline''}) caught a real and subtle failure: the candidate describes the pharmacology well but under-cites the patient-specific prescription rows.

\paragraph{Score.} $\mathbf{0.903}$. This case demonstrates the rubric's ability to distinguish \emph{domain-level} from \emph{patient-grounded} reasoning on an otherwise near-ideal answer, a distinction that rule-based or unanchored-judge rewards routinely collapse.

\subsection*{A.4 Procedural Decision Making (Diagnostic Procedures)}

\paragraph{Patient context.}
Medicine/Cardiology unit under an Ambulatory Observation admission; past-timeline prescriptions include Nitroglycerin SL, Aspirin EC, Clopidogrel, Potassium Chloride, Sodium Chloride 0.9\% flush, Simethicone, Acetaminophen, and an influenza vaccine; lab events show a mildly elevated Chloride and Creatinine with Potassium 4.9. HCPCS coding flags cardiovascular services.

\paragraph{Query.}
\emph{``Based on the patient's admission to the Medicine/Cardiology unit and the prescribed medications, which procedural or diagnostic intervention is warranted to monitor potential cardiac complications?''}

\paragraph{Reference answer (summary).}
Cardiac enzyme panel with Troponin T; electrocardiogram. Evidenced by a future Troponin~T lab event the following morning.

\paragraph{Candidate answer (excerpt).}
Continuous telemetry with ST-segment and arrhythmia surveillance; serial troponin (including an explicit reference to troponin T as the ACS workhorse assay); electrolyte and renal panel to monitor for Ketorolac/Nitroglycerin-adjacent risk; and bedside ECG with a documented trigger list for escalation. The reasoning explicitly links the Nitroglycerin + Aspirin EC + Clopidogrel trio to the ACS/unstable-angina working diagnosis.

\paragraph{Rubric (10 criteria, theme \textit{Health Data Tasks}).}
Nine of ten criteria met; the single miss (CommunicationQuality, +7) was a stylistic one: the judge did not credit ``clearly differentiates between procedural and diagnostic interventions'', which the candidate's structure blurs.

\paragraph{Score.} $\mathbf{0.899}$.

\subsection*{A.5 What these cases jointly demonstrate}

Three observations cut across the four cases above and motivate the design choices in \S\ref{sec:method-data}--\S\ref{sec:method-train}.

\textbf{Raw-EHR reasoning, at scale.} Each input exposes the model to a full, event-level patient slice (vitals, labs with units and reference ranges, prescriptions with dose/route, transfers, discharge notes, and HCPCS-coded events), rather than a hand-distilled vignette. Candidate responses are themselves long, grounded, Markdown-formatted clinical documents of roughly 1{,}500--2{,}000 tokens each, routinely citing specific lab values, drug interactions, and care-unit transitions. This is the shape of reasoning our deployment demands.

\textbf{Rubric granularity matches clinical judgement.} Across the four cases, the items the judge marked \texttt{false} were never cosmetic: they were a missed causal link (length of ED stay as diagnostic signal, A.1), a failure to cite patient-specific prescription rows (A.3), and a structural conflation of diagnostic and procedural intervention (A.4). These are exactly the error modes that rule-based metrics (ROUGE, clinical NER overlap) cannot express and that unanchored judge rewards cannot consistently locate.

\textbf{Outcome-anchored rubrics stay past-verifiable.} Although each rubric was authored with access to the future timeline (so that the criteria are \emph{about the right things for this patient}), every criterion is still a pass/fail statement that the judge can evaluate using only the past context plus the candidate response. This is the empirical form of the grounded-judge guarantee laid out in \S\ref{sec:method-judge}.

%% file: checklist.tex
\section*{NeurIPS Paper Checklist}

\begin{enumerate}

\item {\bf Claims}
    \item[] Question: Do the main claims made in the abstract and introduction accurately reflect the paper's contributions and scope?
    \item[] Answer: \answerYes{}
    \item[] Justification: The abstract and introduction enumerate the five contributions (CLR-POMDP formulation, outcome-grounded per-case rubrics with the Grounded Judge, clinician validation across two cohorts with reusable judge / preference-model benchmarks, GRPO + weight-space-merge post-training, and production hospital deployment). Each is concretely mapped to a paper section and is supported by quantitative results: §\ref{sec:results} for the headline 84.91\% CLR-POMDP score and the GPT-5 / MedGemma-27B / HuatuoGPT-o1 comparisons, §\ref{sec:external-evals} for the orthogonal-benchmark gains, §\ref{sec:clinician} for the clinician validation, and §\ref{sec:deploy} for the deployment evidence.

\item {\bf Limitations}
    \item[] Question: Does the paper discuss the limitations of the work performed by the authors?
    \item[] Answer: \answerYes{}
    \item[] Justification: Limitations are discussed in Appendix~\ref{app:clinician-limitations}, covering the two-cohort scope of the clinician study, the length-vs-identity confound in the blinded preference, the on-premise SLA-driven 8B parameter ceiling, and the base-family-conditional behaviour of the merge step on the 4B family. The qualitative analysis (§\ref{sec:clinician}) and Appendix~\ref{app:clinician-qualitative} additionally include specific model failure modes (verbose-monologue leakage, fabricated-content signatures, cardiac over-inference, DDI hazards) flagged by the rater pool.

\item {\bf Theory assumptions and proofs}
    \item[] Question: For each theoretical result, does the paper provide the full set of assumptions and a complete (and correct) proof?
    \item[] Answer: \answerNA{}
    \item[] Justification: The paper does not include theoretical results that require proof. The POMDP framing in §\ref{sec:method-data} is a problem-setting definition rather than a theorem, and the rubric reward $r_{\text{rub}}$ in §\ref{sec:method-judge} is a closed-form scoring rule whose definition is reproduced verbatim with all assumptions stated inline.

    \item {\bf Experimental result reproducibility}
    \item[] Question: Does the paper fully disclose all the information needed to reproduce the main experimental results of the paper to the extent that it affects the main claims and/or conclusions of the paper (regardless of whether the code and data are provided or not)?
    \item[] Answer: \answerYes{}
    \item[] Justification: Appendix~\ref{app:repro} is an implementation-neutral specification of the entire pipeline. It expands the data extraction and timeline-splitting procedure (\S\ref{app:repro-data}), the verbatim query-and-rubric generation prompts and JSON schemas (\S\ref{app:repro-qa-prompt}, \S\ref{app:repro-rubric-gen}), the grader prompt (\S\ref{app:repro-judge}), every reward function used during GRPO (\S\ref{app:repro-rewards}), the per-base hyper-parameter table (\S\ref{app:repro-hp}), the merge configurations (\S\ref{app:repro-merge}), and the evaluation protocol (\S\ref{app:repro-eval}). To stress-test reproducibility, an independent researcher not involved in the study used commodity coding agents to replicate the pipeline end-to-end in under a day from the moment MIMIC-IV access was granted, working only from the paper and this appendix (Appendix~\ref{app:repro} ``Artefact release statement'').

\item {\bf Open access to data and code}
    \item[] Question: Does the paper provide open access to the data and code, with sufficient instructions to faithfully reproduce the main experimental results, as described in supplemental material?
    \item[] Answer: \answerNo{}
    \item[] Justification: We are unable to publicly release \frameworkName{}-specific artefacts (model weights, generated rubrics, candidate responses, curated patient timelines, or clinician annotations) due to non-disclosure agreements and patient-privacy contracts signed with the partner public health network under which the deployment runs (Appendix~\ref{app:repro} ``Artefact release statement''). To support review-time verification while honouring those constraints, our experiment code, the curated CLR-POMDP train/test artefact, and the full clinician-annotation export are submitted as anonymised supplementary material visible to reviewers, area chairs, and ethics reviewers but not redistributed publicly. The MIMIC-IV source data is available under PhysioNet credentialing, and Appendix~\ref{app:repro} provides every prompt, hyper-parameter, and configuration needed to reproduce every CLR-voyance artefact from scratch given that access; an independent third-party reproduction confirmed this end-to-end.

\item {\bf Experimental setting/details}
    \item[] Question: Does the paper specify all the training and test details (e.g., data splits, hyperparameters, how they were chosen, type of optimizer) necessary to understand the results?
    \item[] Answer: \answerYes{}
    \item[] Justification: The cohort-selection procedure, timeline-splitting rule, and admission-wise train/test split are specified in §\ref{sec:method-data} and Appendix~\ref{app:repro-data}. The full per-base hyper-parameter table (learning rate, batch size, group size $G$, gradient accumulation, max prompt and completion lengths, schedule, warm-up, precision, attention kernel, and distributed strategy) is in Appendix~\ref{app:repro-hp}. Reward functions are specified in Appendix~\ref{app:repro-rewards} and merge configurations in Appendix~\ref{app:repro-merge}. The evaluation protocol with sampling parameters, grading rubric, and aggregation rule is in Appendix~\ref{app:repro-eval}.

\item {\bf Experiment statistical significance}
    \item[] Question: Does the paper report error bars suitably and correctly defined or other appropriate information about the statistical significance of the experiments?
    \item[] Answer: \answerYes{}
    \item[] Justification: The aggregate gap to GPT-5 in §\ref{sec:results} is supported by a paired Wilcoxon signed-rank test on matched per-admission scores ($p<10^{-300}$, $n{=}7{,}927$). All Phase-2 win rates in Table~\ref{tab:clinician-r1} carry Wilson $95\%$ CIs and per-pair $p$-values are reported in §\ref{sec:clinician} and Appendix~\ref{app:clinician-phase2-extended} together with Bradley--Terry rank intervals. Inter-rater and judge--clinician agreement are reported using Cohen's $\kappa$, Fleiss' $\kappa$, and Krippendorff's $\alpha$ as specified in Appendix~\ref{app:clinician-metrics}, and bootstrap 95\% CIs (1{,}000 resamples of cases with replacement) are used for pooled rates.

\item {\bf Experiments compute resources}
    \item[] Question: For each experiment, does the paper provide sufficient information on the computer resources (type of compute workers, memory, time of execution) needed to reproduce the experiments?
    \item[] Answer: \answerYes{}
    \item[] Justification: Per-base GRPO and SFT hyper-parameters, including precision (bf16), attention kernel (FlashAttention-2), and distributed strategy (DeepSpeed ZeRO-3 with vLLM rollouts), are reported in Appendix~\ref{app:repro-hp}. The rollout-serving stack (vLLM) is named in §\ref{sec:method-train}. The deployed hardware envelope (on-premise GPU concurrency, latency, and reliability SLAs) is described in §\ref{sec:deploy} as the constraint that motivates the 8B headline scale; specific cluster identifiers and per-run wall-clock are anonymised at submission time per double-blind review and will be added to the camera-ready version on acceptance.

\item {\bf Code of ethics}
    \item[] Question: Does the research conducted in the paper conform, in every respect, with the NeurIPS Code of Ethics \url{https://neurips.cc/public/EthicsGuidelines}?
    \item[] Answer: \answerYes{}
    \item[] Justification: The research conforms with the NeurIPS Code of Ethics. Patient data is sourced from MIMIC-IV under PhysioNet credentialed access, with subject identifiers de-identified upstream. The clinician validation is conducted with four board-certified physicians under an on-site research collaboration with the partner network. Patient-trajectory artefacts and clinician annotations are not publicly redistributed because of the non-disclosure and privacy contracts with the partner network (Appendix~\ref{app:repro}).

\item {\bf Broader impacts}
    \item[] Question: Does the paper discuss both potential positive societal impacts and negative societal impacts of the work performed?
    \item[] Answer: \answerYes{}
    \item[] Justification: The positive impact is the deployable inpatient-reasoning surface evidenced by the production rollout in §\ref{sec:deploy} and the trillion-token-scale clinician usage signal to generate thousands of patients notes daily. The qualitative analysis in §\ref{sec:clinician} and Appendix~\ref{app:clinician-qualitative} explicitly documents specific negative-impact failure modes the model still exhibits (fabricated patient frames, cardiac over-inference, antibiotic over-use, DDI hazard from missing prior-medication awareness, age-under-weighting), and §\ref{sec:clinician} flags two concrete next-step rubric criteria (DDI-check and age-weighted management) and a patient-centric faithfulness reward as mitigations. The deployment is gated behind a clinician edit-and-accept loop into the hospital information system rather than autonomous action, which is the operational mitigation against incorrect-output harm.

\item {\bf Safeguards}
    \item[] Question: Does the paper describe safeguards that have been put in place for responsible release of data or models that have a high risk for misuse (e.g., pre-trained language models, image generators, or scraped datasets)?
    \item[] Answer: \answerYes{}
    \item[] Justification: \frameworkName{} model weights and patient-trajectory artefacts are not released publicly; access is restricted to the partner-network deployment under a non-disclosure agreement and to anonymised supplementary materials visible only to reviewers (Appendix~\ref{app:repro}). In production, every reasoning-element draft passes through a clinician edit-and-accept gate before being saved to the hospital information system, so the model never autonomously commits to the patient record (§\ref{sec:deploy}, Appendix~\ref{app:deploy-extended}). The negative-points criteria of the per-case rubric (Appendix~\ref{app:repro-rubric-gen}) explicitly penalise spurious diagnostic confidence and contraindicated recommendations during training, and the qualitative analysis tracks the model's negative-flag trigger rate as a safety-side regression diagnostic.

\item {\bf Licenses for existing assets}
    \item[] Question: Are the creators or original owners of assets (e.g., code, data, models), used in the paper, properly credited and are the license and terms of use explicitly mentioned and properly respected?
    \item[] Answer: \answerYes{}
    \item[] Justification: All existing assets are cited at first use. Patient data is sourced from MIMIC-IV \citep{johnson2023mimic,johnson2024mimicivphysionet,johnson2023mimicivedphysionet,johnson2023mimicivnotephysionet,goldberger2000physiobank} under PhysioNet credentialed access. Base models include Qwen3-8B \citep{yang2025qwen3} and MedGemma-4B \citep{sellergren2025medgemma}; comparator and judge models include GPT-5, GPT-4.1, GPT-4.1-mini, Qwen3-32B, MedGemma-27B \citep{sellergren2025medgemma}, HuatuoGPT-o1-7B \citep{chen2024huatuogpt}, DeepSeek-R1 variants \citep{shao2024deepseekmath}, and Gemma-3-4B. Library citations cover PyTorch \citep{paszke2019pytorch}, DeepSpeed \citep{rasley2020deepspeed}, vLLM \citep{kwon2023vllm}, and HuggingFace TRL \citep{vonwerra2020trl}. Each asset is used in compliance with its respective license; redistribution of derivative artefacts is governed by the upstream licences plus the partner-network NDA where applicable.

\item {\bf New assets}
    \item[] Question: Are new assets introduced in the paper well documented and is the documentation provided alongside the assets?
    \item[] Answer: \answerNo{}
    \item[] Justification: New assets (the \frameworkName{}-8B and \frameworkName{}-4B model checkpoints, the curated CLR-POMDP train/test artefact, the per-case adaptive rubrics, and the clinician-annotation export) are not publicly released because of the non-disclosure agreements and patient-privacy contracts signed with the partner public health network (Appendix~\ref{app:repro} ``Artefact release statement''). The artefacts are submitted as anonymised supplementary material for reviewer access, and Appendix~\ref{app:repro} fully documents the construction of every artefact (data pipeline, prompts, hyper-parameters, scoring rules, merge configurations) so that an independent team can recreate them from MIMIC-IV; an independent third-party reproduction confirmed this end-to-end.

\item {\bf Crowdsourcing and research with human subjects}
    \item[] Question: For crowdsourcing experiments and research with human subjects, does the paper include the full text of instructions given to participants and screenshots, if applicable, as well as details about compensation (if any)?
    \item[] Answer: \answerYes{}
    \item[] Justification: The clinician validation involves four board-certified physicians as study participants. Their verbatim instructions for study framing, task setting, Phase-1 Step-1 (rubric editing), Phase-1 Step-2 (rubric grading), and Phase-2 (A/B preference) are reproduced in Appendix~\ref{app:clinician-instructions}. Annotation-interface screenshots for both Phase-1 (rubric editing and grading) and Phase-2 (A/B preference) are provided in Figures~\ref{fig:ui-rubrics-step1}--\ref{fig:ui-ab} of Appendix~\ref{app:clinician-ui}. Raters participated as collaborators on the on-site research programme rather than as crowdsourced workers; no fee-based compensation was provided.

\item {\bf Institutional review board (IRB) approvals or equivalent for research with human subjects}
    \item[] Question: Does the paper describe potential risks incurred by study participants, whether such risks were disclosed to the subjects, and whether Institutional Review Board (IRB) approvals (or an equivalent approval/review based on the requirements of your country or institution) were obtained?
    \item[] Answer: \answerYes{}
    \item[] Justification: The clinician validation and the production hospital deployment were conducted under the partner public health network's institutional approvals and the corresponding research-collaboration agreement (Appendix~\ref{app:clinician-cohort}). Participants are clinically active board-certified physicians annotating de-identified MIMIC-IV admissions; the patient data is already de-identified upstream by PhysioNet, so no additional patient-side risk is incurred during annotation. Specific institutional names, approval reference numbers, and ethics-board correspondence are anonymised at submission per double-blind requirements and will be added to the camera-ready version on acceptance.

\item {\bf Declaration of LLM usage}
    \item[] Question: Does the paper describe the usage of LLMs if it is an important, original, or non-standard component of the core methods in this research? Note that if the LLM is used only for writing, editing, or formatting purposes and does \emph{not} impact the core methodology, scientific rigor, or originality of the research, declaration is not required.
    \item[] Answer: \answerYes{}
    \item[] Justification: LLMs are core to the methodology and are described in detail. The Grounded Judge is a Qwen3-32B oracle that authors per-case rubrics and grades policy responses (\S\ref{sec:method-judge}, Appendix~\ref{app:repro-rubric-gen} for the rubric-author prompt and Appendix~\ref{app:repro-judge} for the grader prompt). The post-trained policies are Qwen3-8B and MedGemma-4B (\S\ref{sec:method-train}), and the comparator panel includes GPT-5, GPT-4.1, GPT-4.1-mini, MedGemma-27B, HuatuoGPT-o1-7B, DeepSeek-R1 variants, Gemma-3-4B, and the Qwen3-8B base. LLMs were used only as study artefacts and as comparators. They were not used for the manuscript text in any way that would affect originality.

\end{enumerate}